\documentclass[lettersize,journal]{IEEEtran}

\usepackage{tikz}
\usetikzlibrary{arrows.meta, positioning}

\usepackage{amsmath,amsfonts}
\usepackage{algorithmic}
\usepackage{algorithm}
\usepackage{array}
\usepackage{textcomp}
\usepackage{stfloats}
\usepackage{url}
\usepackage{verbatim}
\usepackage{graphicx}
\usepackage{booktabs}
\usepackage{subcaption}
\usepackage{cite}
\usepackage{dsfont}
\usepackage{bm}
\usepackage{multirow}

\usepackage{siunitx}
\usepackage{hyperref}
\usepackage{listings}
\usepackage{multirow}

\newcommand{\PARFOR}[1]{\STATE \textbf{parfor} #1 \textbf{do}\begin{ALC@g}}
\newcommand{\ENDPARFOR}{\end{ALC@g}\STATE \textbf{end parfor}}
\definecolor{mygreen}{rgb}{0,0.6,0}
\definecolor{mygray}{rgb}{0.5,0.5,0.5}
\definecolor{mymauve}{rgb}{0.58,0,0.82}
\definecolor{myblue}{rgb}{0,0,1}  % Custom color for specific keywords

\lstdefinestyle{mystyle}{
    backgroundcolor=\color{white},      % choose the background color
    commentstyle=\color{mygreen},       % comment style
    keywordstyle=\color{myblue},        % python keyword style
    identifierstyle=\color{black},      % identifier style (including jnp and functions)
    numberstyle=\tiny\color{mygray},    % the style that is used for the line-numbers
    stringstyle=\color{mymauve},        % string literal style
    basicstyle=\ttfamily\footnotesize,  % the basic font style
    breakatwhitespace=false,            % sets if automatic breaks should only happen at whitespace
    breaklines=true,                    % sets automatic line breaking
    captionpos=b,                       % sets the caption-position to bottom
    keepspaces=true,                    % keeps spaces in text, useful for keeping indentation of code (possibly needs columns=flexible)
    numbers=left,                       % where to put the line-numbers; possible values are (none, left, right)
    numbersep=2pt,                    % how far the line-numbers are from the code
    showspaces=false,                   % show spaces everywhere adding particular underscores; it overrides 'showstringspaces'
    showstringspaces=false,             % underline spaces within strings only
    showtabs=false,                     % show tabs within strings adding particular underscores
    tabsize=2,                          % sets default tabsize to 2 spaces
    language=Python,
    literate={all}{{\textcolor{black}{all}}}3
             {any}{{\textcolor{black}{any}}}3
             {sum}{{\textcolor{black}{sum}}}3,
    xleftmargin=0pt,                    % Adjust left margin
}

\begin{document}

\title{Bridging Evolutionary Multiobjective Optimization and GPU Acceleration via Tensorization}
%\author{\IEEEauthorblockN{Anonymous Authors}}
\author{Zhenyu Liang,
        Hao Li,
        Naiwei Yu,
        Kebin Sun,
        and Ran Cheng
        \thanks{Zhenyu Liang, Hao Li, Naiwei Yu, and Kebin Sun are with the Department of Computer Science and Engineering, Southern University of Science and Technology, Shenzhen 518055, China (e-mails: zhenyuliang97@gmail.com; li7526a@gmail.com; yunaiweiyn@gmail.com; sunkebin.cn@gmail.com).}
        \thanks{
        Ran Cheng is with the Department of Data Science and Artificial Intelligence and the Department of Computing, The Hong Kong Polytechnic University, Hong Kong SAR, China. e-mail: ranchengcn@gmail.com. (\emph{Corresponding author: Ran Cheng})}
}

\markboth{IEEE Transactions on Evolutionary Computation,~2025}%
{Liang \MakeLowercase{\textit{et al.}}: Bridging Evolutionary Multiobjective Optimization and GPU Acceleration}



\maketitle

\begin{abstract}
Evolutionary multiobjective optimization (EMO) has made significant strides over the past two decades. However, as problem scales and complexities increase, traditional EMO algorithms face substantial performance limitations due to insufficient parallelism and scalability. 
While most work has focused on algorithm design to address these challenges, little attention has been given to hardware acceleration, thereby leaving a clear gap between EMO algorithms and advanced computing devices, such as GPUs.
To bridge the gap, we propose to parallelize EMO algorithms on GPUs via the \emph{tensorization} methodology. 
By employing tensorization, the data structures and operations of EMO algorithms are transformed into concise tensor representations, which seamlessly enables automatic utilization of GPU computing. 
We demonstrate the effectiveness of our approach by applying it to three representative EMO algorithms: NSGA-III, MOEA/D, and HypE.
To comprehensively assess our methodology, we introduce a multiobjective robot control benchmark using a GPU-accelerated physics engine. Our experiments show that the tensorized EMO algorithms achieve speedups of up to 1113\(\times\) compared to their CPU-based counterparts, while maintaining solution quality and effectively scaling population sizes to hundreds of thousands. 
Furthermore, the tensorized EMO algorithms efficiently tackle complex multiobjective robot control tasks, producing high-quality solutions with diverse behaviors. 
Source codes are available at \url{https://github.com/EMI-Group/evomo}.
\end{abstract}


\section{Introduction}

In many real-world optimization problems (e.g., material design~\cite{MaterialApp1, MaterialApp2}, energy management~\cite{EnergyApp1, EnergyApp2}, network optimization~\cite{NetworkApp}, and portfolio optimization~\cite{PorfolioApp}), decision-makers must consider multiple (and often conflicting) objectives simultaneously. 
Without loss of generality, such multiobjective optimization problems (MOPs) can be defined as:
\begin{equation}
    \underset{\mathbf{x}}{\text{minimize}} \quad \mathbf{f}(\mathbf{x}) = (f_1(\mathbf{x}), f_2(\mathbf{x}), \ldots, f_m(\mathbf{x})),
\end{equation}
where \(\mathbf{x} = (x_1, x_2, \ldots, x_d) \in \mathcal{X} \subseteq \mathbb{R}^d\) is the decision vector and \(d\) is the number of decision variables. \(\mathbf{f}: \mathcal{X} \rightarrow \mathbb{R}^m\) maps the decision vector to an $m$-dimensional objective space. 
Each \(f_i: \mathcal{X} \rightarrow \mathbb{R}\) for \(i = 1, \ldots, m\) represents an objective function that needs to be minimized (or maximized). 
The key challenge in solving MOPs is to identify a set of trade-off solutions known as the Pareto set (PS), where no single solution can optimize all objectives simultaneously. 
This set contains all Pareto-optimal or nondominated solutions, and their corresponding points in the objective space collectively form the Pareto front (PF).

Over the past two decades, the field of evolutionary multiobjective optimization (EMO)~\cite{MOOBook,survey2} has seen rapid advancements, resulting in the development of various effective algorithms for solving MOPs. Broadly, these algorithms can be categorized into three main methods: 1) dominance-based; 2) decomposition-based; and 3) indicator-based. Dominance-based algorithms, such as NSGA-II~\cite{nsga2} and NSGA-III~\cite{NSGA3}, select solutions based on dominance relations between individuals. Decomposition-based algorithms, like MOEA/D~\cite{moead}, break down an MOP into multiple simpler subproblems, which are optimized collaboratively. Indicator-based algorithms, such as HypE~\cite{hype}, focus on optimizing specific performance indicators, like hypervolume (HV)~\cite{hv}.

While EMO algorithms have proven effective in solving various MOPs, their performance is significantly constrained by limitations in \emph{computing power}. First, since the majority of existing EMO algorithms still rely on CPUs for execution, their computational efficiency is inherently limited, particularly when addressing large-scale MOPs (LSMOPs)~\cite{lsmop}. Second, the inconsistent implementation of EMO algorithms across different methods has led to fragmentation, making it difficult to standardize solutions and apply them across diverse domains. Without a unified framework, efforts to generalize these algorithms and enhance computational efficiency are impeded. Third, much of the current research remains focused on relatively simpler numerical optimization tasks, often neglecting computationally intensive real-world applications. This narrow focus further limits the practical use of EMO algorithms in scenarios where real-time performance and scalability are crucial.

To address these limitations, one promising method is to incorporate modern computational accelerators such as GPUs. With their powerful parallel processing capabilities, GPUs have demonstrated significant performance improvements in fields like deep learning~\cite{DL}. However, to fully leverage the potential of GPUs for EMO algorithms, a systematic method of parallelization is necessary, yet little effort has been made in this direction so far.

Given the high concurrency enabled by the large number of \emph{tensor cores}~\cite{tensorcore}, GPUs are particularly well-suited for efficient handling and acceleration of large-scale data processing. 
Correspondingly, one promising method for parallelization on GPUs is \emph{tensorization}, i.e., representing data structures and operations as tensors.
Building upon this concept, we introduce a concise and general tensorization methodology for accelerating EMO algorithms on GPUs. 
By leveraging tensor operations and the inherent parallelism of GPUs, this methodology systematically explains how to transform EMO algorithms into concise tensor representations. 
Using this methodology, we implement tensorized versions of three representative EMO algorithms from each category: 1) the dominance-based NSGA-III~\cite{NSGA3}; 2) the decomposition-based MOEA/D~\cite{moead}, and 3) the indicator-based HypE~\cite{hype}.
Moreover, to evaluate the performance of tensorized EMO algorithms in GPU computing environments, we develop a multiobjective robot control benchmark using Brax~\cite{brax}, a GPU-accelerated physics engine.
The main contributions of this research are as follows:

\begin{enumerate}
    \item 
    We introduce a general tensorization methodology for EMO algorithms that transforms key data structures (i.e., candidate solutions and objective values) and operations (i.e., crossover, mutation, and selection) into tensor representations. 
    This approach establishes concise yet versatile mathematical models for enabling efficient GPU acceleration of EMO algorithms.
    
    \item 
    We apply the proposed tensorization methodology to three representative EMO algorithms: NSGA-III, MOEA/D, and HypE. The tensorized algorithms achieve up to {1113\(\times\) speedup} compared to their CPU-based counterparts, while maintaining solution quality and effectively scaling population sizes to hundreds of thousands.

    \item 
    We develop a multiobjective robot control benchmark called {MoRobtrol}. 
    This benchmark represents a computationally intensive scenario with complex black-box properties.
    It demonstrates the ability of the tensorized EMO algorithms to efficiently generate high-quality solutions with diverse behaviors in such computationally expensive environments.

\end{enumerate}

The structure of this article is as follows: Section~\ref{sec:background} reviews the background and related work. Section~\ref{sec:methodology1} introduces the tensorization methodology for GPU acceleration. Section~\ref{sec:methodology2} details the implementations of core operations in three representative EMO algorithms. Section~\ref{sec:mrcb} introduces the multiobjective robot control benchmark. 
Section~\ref{sec:experiments} outlines the experimental setup and results. 
Section~\ref{sec:conclusion} summarizes the findings and discusses future work.

\section{Background}
\label{sec:background}
\subsection{Taxonomy of EMO Algorithms}
Traditional EMO algorithms can generally be classified into three main categories based on their selection mechanisms: dominance-based, decomposition-based, and indicator-based~\cite{MOOBook}.

Dominance-based EMO algorithms are pivotal in addressing complex optimization tasks through Pareto dominance. As the pioneering algorithm in this category, NSGA-II~\cite{nsga2} introduced a fast nondominated sorting approach, which has since become a foundation for many subsequent algorithms. SPEA2~\cite{SPEA2} introduces a fine-grained fitness assignment strategy, density estimation, and enhanced archive truncation to improve performance.
GrEA~\cite{grEA} enhances convergence and diversity balance by using grid dominance. NSGA-III~\cite{NSGA3} further advances diversity management in higher-dimensional spaces with reference points. Recent algorithms build on these foundations with innovative strategies. BiGE~\cite{bige} focuses on proximity and diversity through bi-goal optimization, while VaEA~\cite{vaea} balances these using vector-angle-based principles. RSEA~\cite{rsea} improves performance by projecting solutions into a radial space, and NSGA-II/SDR~\cite{nsga2-sdr} introduces a novel dominance relation with adaptive niching techniques. MSEA~\cite{msea} divides the optimization process into stages to enhance diversity preservation. PMEA~\cite{PMEA} eliminates dominance resistance solutions using an interquartile range method.

Decomposition-based EMO algorithms address MOPs by decomposing them into simpler subproblems~\cite{DecomSurvey}. These algorithms can be further categorized into two types: 1) weighted aggregation-based and 2) reference set-based methods. MOEA/D~\cite{moead} is the most representative weighted aggregation-based method, which aggregates objectives using weight vectors and has inspired variants like MOEA/D-DRA~\cite{moeaddra} and EAG-MOEA/D~\cite{eagmoead}.
Recent developments include MOEA/D-AAWNs~\cite{MOEAD-AAWNs}, which adapts weight vectors and neighborhoods to enhance diversity. MOEA/D-GLCM~\cite{MOEAD-GLCM} employs bidirectional global search and adaptive neighborhood strategies to improve population distribution.
Reference set-based methods, on the other hand, divide the objective space using reference points or vectors, guiding the search toward underexplored regions while maintaining diversity. Representative algorithms of this type include MOEA/D-M2M~\cite{moeadm2m}, RVEA~\cite{rvea}, and $\theta$-DEA~\cite{tdea}. 
A recent work, ECRA-DEA~\cite{ECRA-DEA}, adaptively allocates resources across subspaces using fitness contribution and improvement rates.

Indicator-based EMO algorithms rely on performance indicators for selection~\cite{IBEASurvey}. 
As a notable example, IBEA~\cite{ibea} uses binary $\epsilon^+$ indicators for decision-making. 
Following this, SMS-EMOA~\cite{smsemoa} and HypE~\cite{hype} are representative HV-based algorithms, with HypE using Monte Carlo sampling to approximate HV.
BCE-IBEA~\cite{bceibea} integrates bi-criterion evolution with the IBEA framework, while SRA~\cite{sra} combines multiple indicators with stochastic ranking based environmental selection. 
MOMBI-II~\cite{mombi2} uses the R2 metric, and AR-MOEA~\cite{armoea} is guided by the enhanced inverted generational distance (IGD) metric.
MaOEA/IGD~\cite{maoeaigd} employs the IGD metric. 
Recently, R2HCA-EMOA~\cite{R2HCA-EMOA}, HVCTR~\cite{HVCTR}, and IMOEA-ARP~\cite{IMOEA-ARP} extended the SMS-EMOA framework with innovations in HV approximation, reference point management, and diversity handling, respectively.

\subsection{GPU Acceleration in EMO Algorithms}
Most previous efforts in GPU acceleration for EMO have focused on specific algorithms and implementations of certain algorithmic components. 
An early contribution by Wong~\cite{gpu-nsga2-2009} introduced GPU-accelerated nondominated sorting in NSGA-II.
Building on this, Sharma and Collet~\cite{G-ASREA} proposed G-ASREA, a GPU-accelerated variant of NSGA-II that incorporates an external archive to sort nondominated solutions on the GPU.
Arca \textit{et al.}~\cite{gpu-nsga2-on-fuel} applied GPU acceleration to the evaluation phase of NSGA-II, specifically within the context of fuel treatment optimization. 
Aguilar-Rivera~\cite{gpu-nsga2-2020} further developed a fully vectorized NSGA-II, employing stochastic non-domination sorting and grid-crowding techniques.

In addition to dominance-based algorithms, Souza and Pozo~\cite{MOEAD-ACO} proposed the GPU-accelerated MOEA/D-ACO algorithm, where solution construction and pheromone matrix updates benefit from data-parallel processing. 
Lopez \textit{et al.}~\cite{gpu-hv} leveraged GPUs to accelerate HV contribution calculations, improving performance in the SMS-EMOA algorithm. 
Furthermore, Hussain and Fujimoto~\cite{GPU-MOPSO} introduced a fast CUDA-based implementation of MOPSO on GPUs.

More recently, frameworks based on Google’s JAX~\cite{jax2018github} have opened new possibilities for GPU acceleration in EMO. 
Examples include EvoJAX~\cite{evojax}, evosax~\cite{evosax}, and EvoX~\cite{evox}, all of which provide open-source platforms for GPU-accelerated evolutionary algorithms. 
While EvoJAX and evosax focus on accelerating evolutionary strategies, EvoX is designed as a distributed GPU-accelerated framework that supports general evolutionary computation. 
Building upon EvoX, a recent effort has been made to tensorize the RVEA~\cite{rvea} for GPU acceleration~\cite{tensorrvea}.

Despite these advances, GPU-accelerated EMO algorithms remain in their infancy.
Current research has primarily focused on specific implementations that provide isolated performance improvements. Moreover, many of these implementations rely heavily on CUDA programming~\cite{nvidia-cuda} and are not open-source, thus making them less accessible, particularly for beginners.

\section{Tensorization Methodology}
\label{sec:methodology1}

In this section, we present how to adopt the general tensorization methodology in EMO algorithms. 
Specifically, we begin by defining the notation and preliminary concepts used throughout this article, including the basic definitions of \emph{tensor} and \emph{tensorization}. 
Next, we demonstrate how to transform atomic operations such as basic operations and control flow operations into tensors.
Furthermore, we discuss why and how tensorization matters for GPU acceleration.

\subsection{Preliminaries}

A \emph{tensor} is a multidimensional array that generalizes scalars, vectors, and matrices to higher dimensions~\cite{tensor}. Formally, a $k$-th order tensor is an element of the tensor product of $k$ vector spaces: \(\mathcal{T} \in \mathbb{R}^{d_1 \times d_2 \times \cdots \times d_k}\),
where $d_i$ represents the dimension along the $i$-th mode (axis) of the tensor. Scalars are zero-order tensors, vectors are first-order tensors, and matrices are second-order tensors. Correspondingly, \emph{tensorization} refers to the process of transforming algorithmic data structures and operations into tensor representations. This transformation facilitates efficient parallel computation, particularly on massively parallel hardware such as GPUs, by leveraging the inherent parallelism in tensor operations to improve computational performance and scalability.

In this article, scalars (0-order tensors) are denoted by lowercase letters (e.g., \(a\)), vectors (1st-order tensors) are denoted by italicized bold lowercase letters (e.g., \(\bm{a}\)), matrices (2nd-order tensors) are denoted by italicized bold uppercase letters (e.g., \(\bm{A}\)), and higher-order tensors are denoted by calligraphic letters (e.g., \(\mathcal{T}\)). 
The tensor variables related to EMO algorithms are summarized in Table~\ref{tab:notation}.
Correspondingly, a tensorized MOP can be formulated as:
\begin{equation}
    \label{eq:mops}
    \underset{\bm{X}}{\text{minimize}} \quad \bm{F}(\bm{X}) = (\bm{f}_1(\bm{X}), \bm{f}_2(\bm{X}), \ldots, \bm{f}_m(\bm{X})),
\end{equation}
where \(\bm{X} \in \mathbb{R}^{n \times d}\) is the solution tensor for \(n\) individuals and \(d\) dimensions, and \(\bm{F} \in \mathbb{R}^{n \times m}\) is the corresponding objective tensor.

\begin{table}[t]
    \centering
    \caption{Tensor Variables in EMO Algorithms}
    \label{tab:notation}
    \begin{tabular}{cc}
    \toprule
    \textbf{Notation} & \textbf{Description} \\ 
    \midrule
    \(n\) & Population size \\ 
    \(m\) & Number of objectives \\ 
    \(d\) & Problem dimension \\ 
    \(\bm{X}\) & Solution tensor \\ 
    \(\bm{F}\) & Objective tensor \\ 
    \(\bm{R}, \bm{W}\) & Reference and weight tensors \\ 
    \(\bm{U}, \bm{L}\) & Upper and lower bound tensors \\ 
    \bottomrule
    \end{tabular}
\end{table}

\begin{table}[htb]
    \centering
    \caption{Basic Tensor Operations}
    \label{tab:operation}
    \begin{tabular}{c p{6.5cm}}
    \toprule
    \textbf{Operation} & \multicolumn{1}{c}{\textbf{Description}} \\ 
    \midrule
    $\bm{A} \cdot \bm{B}$ & Tensor multiplication: \newline Product of two tensors $\bm{A}$ and $\bm{B}$. \\ 
    $\bm{A} \odot \bm{B}$ & Hadamard product: \newline Element-wise multiplication of $\bm{A}$ and $\bm{B}$. \\ 
    $H(\bm{A})$ & Heaviside step function: \newline Returns 1 if $\bm{A}_{ij} \geq 0$, 0 otherwise. \\ 
    $\mathds{1}_{\bm{A}}$ & Indicator function: Returns 1 if $\bm{A}_{ij}$ is true, 0 otherwise. \\

    \midrule
    sort & Arranges elements in ascending order. \\
    argsort & Returns indices of sorted elements. \\ 
    % lexsort & Sorts arrays using ordered keys. \\
    min, max & Returns the smallest or largest element along the specified axis: column-wise if $\text{axis}=0$, row-wise if $\text{axis}=1$.\\
    argmin & Returns the indices of the smallest elements. \\ 
    $\texttt{vmap}$ & Vectorization map: \newline Applies a function across an array axis. \\ 
    \bottomrule
    \end{tabular}
\end{table}

\subsection{Tensorization of Data Structures in EMO Algorithms} 

In EMO algorithms,  the candidate solutions and their corresponding objective values are two critical data structures, as expressed in (2).
They can be encoded as tensors \(\bm{X}\) and \(\bm{F}\), respectively. 
Moreover, decomposition-based algorithms use additional structures such as reference vectors or weights, which are similarly encoded as tensors \(\bm{R}\) and \(\bm{W}\), respectively. 
Here, \(\bm{R} \in \mathbb{R}^{r \times m}\) typically represents \(r\) reference vectors, each corresponding to a different point in the objective space, and \(\bm{W} \in \mathbb{R}^{n \times m}\) denotes a weight tensor with \(n\) different weights.
These tensor representations allow for the parallel and batch processing of operations along the population dimension, capitalizing on the independence of individuals within the population. 
Consequently, the EMO algorithm can efficiently process one entire population at a time.

\subsection{Tensorization of Operations in EMO Algorithms}

After establishing an effective tensorized data structure, the next crucial step is to consider tensorization for core operations in EMO algorithms, such as crossover, mutation, and selection mechanisms (e.g., environmental selection). These operations consist of numerous atomic operations, including both basic tensor operations and control flow operations.

\subsubsection{Basic Tensor Operations}

Basic tensor operations serve as the foundation for transforming more complex operations in EMO algorithms. Table \ref{tab:operation} summarizes these operations, including tensor multiplication, Hadamard product, Heaviside step function, and indicator function, as well as advanced functions like sort, argsort, min, max, and argmin. These functions are essential for implementing selection and ranking strategies, enabling GPU-accelerated nondominated sorting, and diversity maintenance in EMO algorithms.

\subsubsection{Control Flow Operations}

The tensorization of control flow operations, including \emph{loops} and \emph{branches}, is a key challenge in transforming traditional EMO algorithms into their tensor-based counterparts. 
Control flow operations are commonly used to implement iterative procedures and define selection rules based on specific conditions. However, traditional control flow operations such as loops (\texttt{for} and \texttt{while}) and \texttt{if-else} branches introduce sequential dependencies that hinder parallel execution and reduce the efficiency of GPU computations. Tensorizing these operations requires replacing them with tensor-based operations that can execute in parallel.

\emph{Loop} operations are often used in EMO algorithms for tasks like calculating distances or updating solutions. However, with a large population size, sequential processing becomes inefficient. These loops can be replaced by either vectorized mapping functions (e.g., \texttt{vmap}), or by using broadcasting combined with basic operations. The \texttt{vmap} function automatically applies a specified function across all elements in a given tensor dimension, eliminating explicit loops. Mathematically, \texttt{vmap} can be expressed as:
\begin{equation}
    \texttt{vmap}(f)(\bm{A}) = [ f(\bm{A}_1), f(\bm{A}_2), \dots, f(\bm{A}_n)]^\top,
\end{equation}
where \( \bm{A} \in \mathbb{R}^{n \times m} \) is the input tensor and \( f \) is the function applied to each individual \( \bm{A}_i \). The function \( f \) operates independently on each element, with \texttt{vmap} managing parallel computation and concatenation of results.
Alternatively, broadcasting can be used to perform the same operations without any loops. Broadcasting works by expanding the dimensions of tensors to align them, enabling element-wise operations to be performed simultaneously across the entire population. 

\emph{Branch}, such as \texttt{if-else} operations, are another type of control flow that poses challenges for tensorization. In traditional implementations, \texttt{if-else} operations introduce branching and disrupt parallel execution. To address this issue, tensorization replaces branch with element-wise masking operations such as \texttt{where}. For example, a traditional branch assigning values to a population matrix based on a threshold can be written as:
\begin{equation}
    \bm{Y}_{ij} = \begin{cases} \bm{A}_{ij}, & \text{if } \bm{M}_{ij} > \tau \\ \bm{B}_{ij}, & \text{otherwise} \end{cases},
\end{equation}
where \( \bm{A} \), \( \bm{B} \) are input tensors, \(\tau\) is a threshold, and \(\bm{M}\) is the mask tensor.
It can be replaced by a tensorized masking operation:
\begin{equation}
    \bm{Y} = \mathds{1}_{\bm{M}>\tau} \odot \bm{A} + (1 - \mathds{1}_{\bm{M}>\tau}) \odot \bm{B},
\end{equation}
which can be implemented as \(\texttt{where}(\bm{M} > \tau, \bm{A}, \bm{B})\).

\begin{figure}[t]
\begin{lstlisting}[caption={Comparison between conventional and tensor-based implementations of Pareto dominance detection.}, label={lst:matrix_tensor}, frame=lines, framerule=0.8pt]
import torch

# Conventional implementation
def dominance_detection_conventional(P):
    n = P.size(0)
    dom = torch.zeros(n, n, dtype=torch.bool)
    for i in range(n):
        for j in range(n):
            if i != j:
                if (P[i] <= P[j]).all() and 
                    (P[i] < P[j]).any():
                    dom[i, j] = True
    return dom
        
# Tensor-based implementation
@torch.compile
def dominance_detection_tensor(P):
    P1 = P.unsqueeze(1)
    P2 = P.unsqueeze(0)
    return (P1 <= P2).all(dim=2) & \
           (P1 < P2).any(dim=2)

\end{lstlisting}
\end{figure}

\subsubsection{Advantages over Conventional Operations}

Tensorization offers several key advantages over conventional EMO implementations. 
First, it provides greater flexibility by handling multi-dimensional data, whereas conventional matrix operations are often limited to two dimensions.
Second, tensorization enhances computational efficiency by enabling parallel processing and removing the need for explicit loops and conditional branches. 
Finally, tensorization simplifies the code, making it more concise and easier to maintain.
For instance, as shown in Listing~\ref{lst:matrix_tensor}, conventional Pareto dominance detection relies on nested \texttt{for} loops and \texttt{if-else} statements to compare individuals. 
In contrast, the tensorized version uses element-wise operations, broadcasting, and masking to perform these comparisons in parallel, significantly reducing code complexity and boosting performance.

\subsection{Discussion}

\subsubsection{Why Is Tensorization Crucial for GPU Acceleration} 
With thousands of cores designed to manage multiple tasks simultaneously, GPUs are intrinsically tailored for parallel computing. 
The architectures of GPUs, such as the SIMT (Single Instruction, Multiple Threads) model, enable efficient execution of tensor operations. 
Moreover, the specialized features like NVIDIA's Tensor Cores further enhance performance by accelerating matrix multiplication and accumulation tasks. 
This makes the tensorization methodology ideally suited for GPUs, which inherently involve large-scale parallel computations.

\subsubsection{What Algorithms Are Suitable for Tensorization} 
Algorithms with independent computations and minimal branching are ideal for tensorization, as they can be easily parallelized. 
In contrast, algorithms that depend on sequential processes, frequent branching, or recursion pose challenges for parallelization. 
For example, in traditional MOEA/D, the aggregate function computation relies on results from previous iterations, thus making direct tensorization challenging. 
Nonetheless, by restructuring and decoupling such algorithms, the tensorization methodology can still be effectively applied.

\section{Tensorization Implementation in Representative EMO Algorithms}
\label{sec:methodology2}

In this section, we present the application of tensorization methodology in three representative EMO algorithms: NSGA-III, MOEA/D, and HypE. 
The genetic operators, including mating selection, crossover, and mutation, are common across most EMO algorithms and follow similar tensorization procedures, which are elaborated in Section S.I of the Supplementary Document. 
Here, we focus on the tensorized implementation of the environmental selection operators specific to each algorithm.

It is important to note that both the environmental selection in NSGA-III and the Monte Carlo-based selection in HypE are inherently highly parallelizable, which allows for straightforward tensorization.
In contrast, the MOEA/D algorithm presents a unique challenge due to its fundamentally sequential nature. 
As shown in Algorithm~\ref{alg:original_moead}, each subproblem in MOEA/D involves four interdependent steps that must be executed in sequence. 
This sequential dependency prevents direct tensorization and requires a reconfiguration of the entire process to enable parallel computation, which will be elaborated in Section~\ref{subsec:moead}.

\subsection{Tensorized Environmental Selection in NSGA-III}
\label{subsec:nsga3}

The key components of environmental selection in NSGA-III include \textit{nondominated sorting}, \textit{normalization}, \textit{association}, \textit{niche count calculation}, and \textit{niche selection}.
The tensorization process of each component is elaborated as follows.
The pseudocode of both the original and tensorized algorithms is provided in Section S.II of the Supplementary Document.

\subsubsection{Nondominated Sorting}

Given the combined objective tensor \(\bm{F} \in \mathbb{R}^{2n \times m}\), representing the objective tensor of both parent and offspring populations, the nondominated rank is computed iteratively. 
The primary goal is to assign a nondominated rank to each individual, where lower ranks correspond to better solutions. 

First, the dominance relation tensor \(\bm{D} \in {\{0,1\}}^{2n \times 2n}\) is computed using \texttt{vmap} or broadcasting for parallel processing. 
Each element \(\bm{D}_{ij}\) indicates whether the solution \(\bm{F}_i\) dominates \(\bm{F}_j\):
\begin{equation}
\bm{D}_{ij} = \bm{F}_i \prec \bm{F}_j, \quad i, j = 1, 2, \dots, 2n.
\end{equation}
Next, we calculate the dominance count tensor \(\bm{c} \in \mathbb{Z}^{2n}\), which indicates how many individuals each individual dominates:
\begin{equation}
\bm{c} = \sum_{j=1}^{2n} \bm{D}_{ij}.
\end{equation}
The rank tensor \(\bm{r} \in \mathbb{Z}^{2n}\) is initialized to zeros, and the rank counter \(k\) is set to zero. The boolean tensor \(\bm{p} \in {\{0,1\}}^{2n}\), the set of all nondominated solutions at the current rank, is obtained by \(\bm{p} = \mathds{1}_{\bm{c}=0}\).

In each iteration, individuals sharing the same dominance rank are identified and processed collectively. This method ensures that even with a large population size, the number of iterations remains relatively low. Additionally, the \texttt{while} function is optimized for accelerated computation. The rank tensor \(\bm{r}\) is updated as follows:
\begin{equation}
    \bm{r} = H(\bm{p}) \cdot k + H(1-\bm{p}) \odot \bm{r}.
\end{equation}
After rank assignment, the dominance count \(\bm{c}\) is updated by:
\begin{equation}
    \bm{c} = \bm{c} - \sum_{i=1}^{2n} \bm{p}_i \cdot \bm{D}_{ij} - \bm{p}.
\end{equation}
The process repeats until all individuals are ranked, which means that all elements in \(\bm{p}\) are zero.
Once all ranks are assigned, the tensor \(\bm{r}\) is sorted to determine the rank \(l\) of the \(n\)-th individual.

\subsubsection{Normalization}
After performing nondominated sorting, the objective tensor \(\bm{F}\) undergoes a normalization process, similar to that in the original NSGA-III algorithm. This normalization ensures that the objectives are comparable by mapping them onto a hyperplane, enabling the algorithm to maintain diversity across generations. 
%The normalization pseudocode is shown in Algorithm~\ref{alg:normalize} in the Appendix.

\subsubsection{Association}
In this step, each individual in the population is associated with the closest reference point. The distance between the normalized objective tensor \(\bm{F}^\prime\) and the reference tensor \(\bm{R}\) is computed using the perpendicular distance:
\begin{equation}
    \bm{D} = \|\bm{F}^\prime\| \cdot \sqrt{1 - \left(\bm{F}^\prime \cdot \bm{R}^\top/(\|\bm{F}^\prime\| \cdot \|\bm{R}\|)\right)^2}.
\end{equation}
Based on the distance tensor \(\bm{D}\), the index of the closest reference point \(\bm{\pi}\) is the index of the minimum value in each row of \(\bm{D}\), and the corresponding distance \(\bm{d}\) represents the minimum value in each row. 

\subsubsection{Niche Count Calculation}
For each reference point, the niche count is computed, which indicates how many individuals are associated with that reference point. The niche count tensor \(\bm{\rho}\) is calculated as:
\begin{equation}
    \bm{\rho}_j = \sum_{i=1}^{2n} H(l - \bm{r}_i) \cdot \mathds{1}_{\bm{\pi}_i = j}, \quad j=1,\dots,n_r,
\end{equation}
where \(n_r\) is the number of reference points. The tensor \(\bm{\rho}_l\) represents the niche count for the last front (i.e., the niche count corresponding to the front when \(\bm{r}_i = l\)):
\begin{equation}
    \bm{\rho}_{l,j} = \sum_{i=1}^{2n} \mathds{1}_{\bm{r}_i = l} \cdot \mathds{1}_{\bm{\pi}_i = j}, \quad j=1,\dots,n_r.
\end{equation}
The total number of selected individuals \(n_s\) is then updated as \(n_s = \sum \bm{\rho}\).

\subsubsection{Niche Selection}
In the niche selection process of NSGA-III, the distance tensor \(\bm{D}^\prime\) represents \(\bm{D}\) adjusted by niche counts for each reference tensor. This tensor is used to identify individuals closest to underpopulated niches. The index of the selected individual \(\bm{q}\) is determined by minimizing the distance in each row of \(\bm{D}^\prime\), i.e., \(\bm{q} = \arg\min_j(\bm{D}^\prime)\). After selecting the individual, the rank tensor \(\bm{r}\) is updated by setting \(\bm{r}[\bm{q}] = l-1\), with further updates in subsequent iterations reflecting the inclusion of individuals in the current front.

Once the selection is complete, the indices of the top \(n\) individuals for the next generation are determined by:
\begin{equation}
    \bm{i}_\text{next} = \text{sort}(H(l-\bm{r}) \odot \bm{a} + \infty \cdot (1-H(l-\bm{r})))[:n],
\end{equation}
where \(\bm{a} = [0,1,\dots, n-1]\).
The final solution tensors \(\bm{X}_\text{next}\) and \(\bm{F}_\text{next}\) are formed by selecting individuals using the sorted indices: \(\bm{X}_\text{next} = \bm{X}[\bm{i}_\text{next}]\) and \(\bm{F}_\text{next} = \bm{F}[\bm{i}_\text{next}]\).

\renewcommand{\algorithmicrequire}{\textbf{Input:}}
\renewcommand{\algorithmicensure}{\textbf{Output:}}
\begin{algorithm}
    \caption{Main Framework of Original MOEA/D}
    \label{alg:original_moead}
    \begin{algorithmic}[1]
    \REQUIRE The maximal number of generations $t_\text{max}$; $n$ weight vectors; the neighborhood size $T$ of each weight vector;
    \ENSURE Final population;
    
    \STATE \textbf{Initialization};

    \FOR{$t = 1$ to $t_\text{max}$}
        \FOR{$i = 1$ to $n$}
            \STATE Reproduction;
            \STATE Fitness Evaluation;
            \STATE Ideal Point Update;
            \STATE Neighborhood Update;
        \ENDFOR
    \ENDFOR
    \end{algorithmic}
\end{algorithm}

\subsection{Tensorized Environmental Selection in MOEA/D}
\label{subsec:moead}

\begin{figure*}[htbp]
    \centering
    \includegraphics[width=\textwidth]{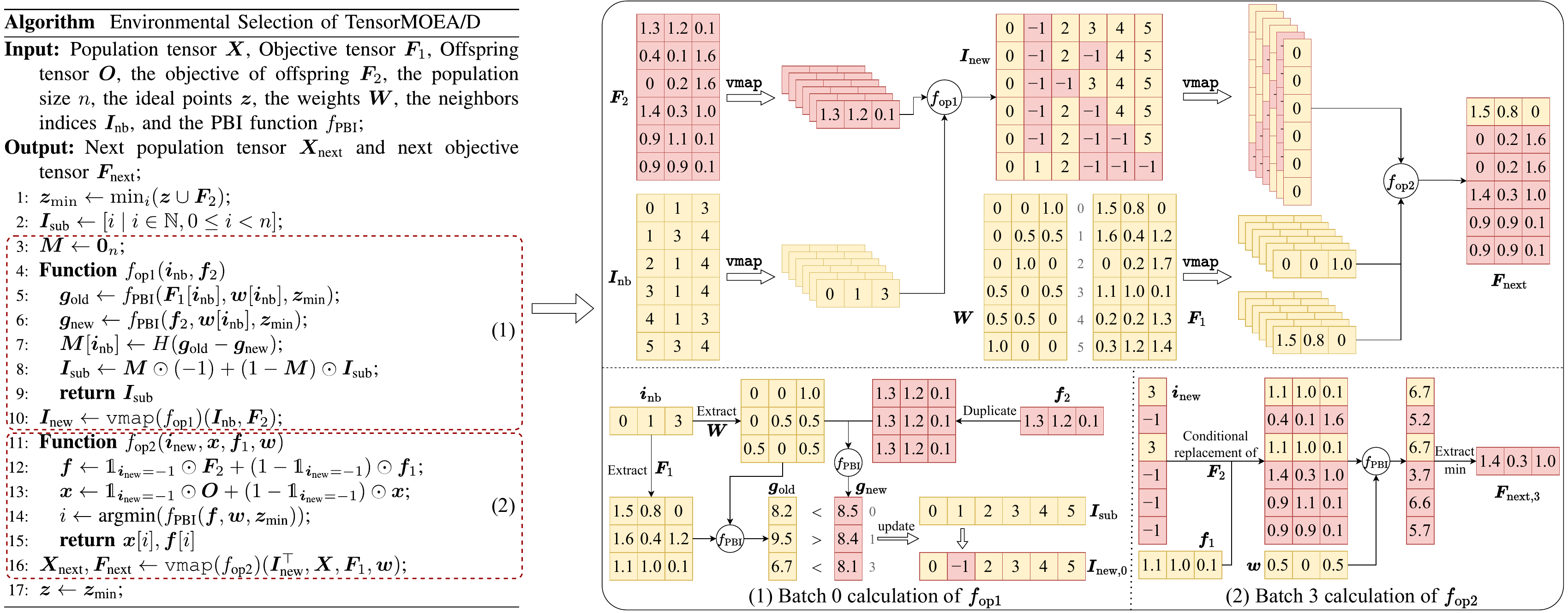}
    % \captionsetup{skip=-0.4pt}
    \caption{Overview of the environmental selection in the TensorMOEA/D algorithm. \textbf{Left}: Pseudocode of the algorithm. \textbf{Right}: Tensor dataflow of module (1) and module (2). The upper part of the right figure shows the overall tensor dataflow for modules (1) and (2), while the lower part presents the batch calculation tensor dataflow, with module (1) on the left and module (2) on the right.}
    \label{fig:moead_framework}
\end{figure*}

In the original MOEA/D algorithm, as shown in Algorithm~\ref{alg:original_moead}, the reproduction, fitness evaluation, ideal point update, and neighborhood update are executed \emph{sequentially} within a single loop. This method requires processing individuals one by one in a specific order, which can significantly impede the execution speed of the algorithm. To address this limitation, we apply tensorization methodology to decouple these four steps in the inner loop (i.e., environmental selection), treating them as independent operations. This adjustment enables parallel processing of all individuals in the tensorized version, referred to as TensorMOEA/D.

In TensorMOEA/D, reproduction generates \( n \) individuals simultaneously based on the neighborhood, contrasting with the original MOEA/D, which produces one individual at a time. The environmental selection process is further divided into two main steps: \textit{comparison and population update}, and \textit{elite selection}. These two steps are primarily implemented using two \texttt{vmap} operations, with the tensorization process detailed as follows.
The pseudocode of both the original and tensorized algorithms is provided in Section S.III of the Supplementary Document.

\subsubsection{Comparison and Population Update}

The primary objective of this step is to determine the indices for the updated population by comparing the aggregated function values of the old and new populations. This process ultimately generates an updated index tensor \(\bm{I}_\text{new}\), where each row corresponds to a subpopulation that mirrors the structure of the original population, containing indices of \(n\) individuals. Positions that require updates are indicated by \(-1\).

Given the solution tensor \(\bm{X}\), objective tensor \(\bm{F}_1\), offspring tensor \(\bm{O}\), the objective of offspring \(\bm{F}_2\), the ideal points \(\bm{z}\), the weights \(\bm{W}\), and the neighbors indices \(\bm{I}_\text{nb}\), the process begins by calculating the minimal objective values to update the reference points, \(\bm{z}_{\min}\), by finding the minimum objective values between the current population and the offspring.

Next, the subpopulation indices \(\bm{I}_{\text{sub}}\) are created to track population updates. The \texttt{vmap} function is utilized to calculate the update indices \(\bm{I}_\text{new}\):
\begin{equation}
    \bm{I}_\text{new} = \texttt{vmap}(f_\text{op1})(\bm{I}_\text{nb}, \bm{F}_2),
\end{equation}
where \(f_\text{op1}(\bm{i_{\text{nb}}}, \bm{f}_2) = \bm{I}_\text{sub}\). This process is conducted in parallel by batching the rows of \(\bm{I}_\text{nb}\) and \(\bm{F}_2\), with \(\bm{i_{\text{nb}}}\) and \(\bm{f}_2\) representing a single batch of these tensors. Each batch undergoes the \(f_\text{op1}\) operation, and all batches are processed simultaneously to yield results for all updates. A visual representation of this operation can be found in Fig.~\ref{fig:moead_framework}, with an example for the first batch illustrated in the lower left corner of the tensor dataflow on the right.

In the \(f_\text{op1}\) function, \(\bm{i_{\text{nb}}}\) is used as an index to extract the corresponding rows from \(\bm{F}_1\) and \(\bm{W}\), and the entries of \(\bm{f}_2\) are replicated to obtain the inputs needed for the aggregation function. The aggregation function employs the penalty-based boundary intersection (PBI) function:
\begin{equation}
    f_{\text{PBI}}(\bm{f}, \bm{w}, \bm{z}) = d_1 + \theta \cdot d_2,
\end{equation}
where \(d_1 = \frac{\left\|(\bm{f}-\bm{z})^\top \cdot \bm{w}\right\|}{\|\bm{w}\|}\), \(d_2 = \left\|(\bm{f}-\bm{z}) - \left(d_1 \cdot \bm{w}\right)\right\|\), and \(\theta\) is a preset penalty parameter.
The old and new aggregated values \(\bm{g}_{\text{old}}\) and \(\bm{g}_{\text{new}}\) are computed by applying the PBI function. The mask \(\bm{M}\) stores the comparison results and is initially a zero tensor, updated as follows: 
\begin{equation}
    \bm{M}[\bm{i}_\text{nb}] = H(\bm{g}_{\text{old}} - \bm{g}_{\text{new}}).
\end{equation}
Finally, the subpopulation indices \(\bm{I}_{\text{sub}}\) are updated based on \(\bm{M}\):
\begin{equation}
    \bm{I}_\text{sub} = \bm{M} \odot (-1) + (1 - \bm{M}) \odot \bm{I}_\text{sub},
\end{equation}
where positions that need updates are assigned a value of \(-1\).

\subsubsection{Elite Selection}
The purpose of this step is to select the best individuals along \( n \) distinct weighted directions based on the aggregated function values. Each direction yields a single elite individual, resulting in a total of \( n \) individuals that form the population of next generation.

To efficiently update the solution and objective tensor for the entire population, the function \texttt{vmap} is applied to parallelize the computation over all rows of the input:
\begin{equation}
    \bm{X}_\text{next}, \bm{F}_\text{next} = \texttt{vmap}(f_\text{op2})(\bm{I}_\text{new}^\top, \bm{X}, \bm{F}_1, \bm{W}),
\end{equation}
where \texttt{vmap} maps the function \(f_\text{op2}(\bm{i}_\text{new}, \bm{x}, \bm{f}_1, \bm{w})\) to each row of the provided inputs in parallel. Fig.~\ref{fig:moead_framework} shows this process in the bottom-right corner, highlighting the computation for batch 3 (i.e., the $4$th row).

The function \(f_\text{op2}\) is defined to update the population and objective tensor based on the new indices \(\bm{i}_\text{new}, \bm{f}_1, \bm{x}, \text{ and } \bm{w}\). Specifically, the objective values \(\bm{f}\) are updated as \(\bm{f} = \mathds{1}_{\bm{i}_\text{new}=-1} \odot \bm{F}_2 + (1-\mathds{1}_{\bm{i}_\text{new}=-1}) \odot \bm{f}_1\). And the individual is updated as \(\bm{x} = \mathds{1}_{\bm{i}_\text{new}=-1} \odot \bm{O} + (1-\mathds{1}_{\bm{i}_\text{new}=-1}) \odot \bm{x}\). The index \(i = \text{argmin}(f_\text{PBI}(\bm{f}, \bm{w}, \bm{z}_{\text{min}}))\) is used to select the best solution in \(\bm{x}\).
Finally, the ideal points \(\bm{z}\) are updated to the minimum objective values \(\bm{z}_\text{min}\) for the next iteration.

\subsection{Tensorized Environmental Selection in HypE}
\label{subsec:hype}

The environmental selection in HypE primarily involves nondominated sorting and HV calculation. In this article, we utilize the Monte Carlo estimation method for HV calculation that is well-suited for tensorization. This Monte Carlo-based HV calculation consists of four key steps: \textit{sampling bound determination}, \textit{sampling weights calculation}, \textit{dominance score and distance update}, and \textit{hypervolume calculation}. The implementation details for these steps are as follows. 
The pseudocode of both the original and tensorized algorithms are provided in Section S.IV of the Supplementary Document.

\subsubsection{Sampling Bound Determination}
The lower and upper bounds for sampling are determined by calculating the minimum objective values, \(\bm{f}_l = \min_i(\bm{F})\), and using the reference point, \(\bm{f}_u = \bm{v}_\text{ref}\). These bounds define the hyperrectangle for sampling. Next, uniformly sample points from the hyperrectangle defined by \(\bm{f}_l\) and \(\bm{f}_u\) to generate a sample tensor \(\bm{S}\) of dimension \(s \times m\). The initial distance score \(\bm{v}_\text{ds}\) is then initialized to zeros with dimensions \(1 \times s\).

\subsubsection{Sampling Weights Calculation}
The sampling weights \(\bm{\alpha}\) are calculated as follows:
\begin{equation}
    \bm{\alpha}_j = \prod_{i=1}^{j} \bm{\lambda}_i / j, \quad j = 1, 2, \dots, k,
\end{equation}
where \(\bm{\lambda}\) is defined as \(\bm{\lambda} = [1, (k - \bm{l})/(n_1 - \bm{l})]\). Here, \(\bm{l} = [i \mid i \in \mathbb{N}, 1 \leq i < n_1]\), and \(n_1\) is the number of rows in \(\bm{F}\).

\subsubsection{Dominance Score and Distance Updating}
Calculate the dominance scores for each sample \(i\) using the function \(f_{\text{pds}}\):
\begin{equation}
    f_{\text{pds}}(\bm{f}) = \mathds{1}_{\sum_{j=1}^{m} H(\bm{S}_{ij} - \bm{f}) = m}, 
\end{equation}
where \(i=1, 2, \dots, s\). The dominance scores \(\bm{T}_\text{pds}\) are then computed by applying \(\texttt{vmap}\) in parallel to the function \(f_{\text{pds}}\) across all rows of the objective tensor \(\bm{F}\).

The distance score \(\bm{v}_\text{ds}\) is then updated based on a temporary matrix \(\bm{T}_\text{temp}\), which is computed as \(\bm{T}_\text{pds}\) combined with \(\bm{v}_\text{ds}\) and a tensor of ones \(\mathds{1}_{n \times 1}\):
\begin{equation}
    \bm{v}_\text{ds} = \text{maximum}\left(\sum_{i=1}^{n_1} \left(\bm{T}_\text{temp}\right)_i - 1, 0\right),
\end{equation}
where \(\bm{T}_\text{temp}\) is calculated as \(\bm{T}_\text{pds} \odot (\mathbf{1}_{n \times 1} \cdot \bm{v}_\text{ds} + 1) + (1-\bm{T}_\text{pds}) \odot (\mathbf{1}_{n \times 1} \cdot \bm{v}_\text{ds})\), and \(\text{maximum}(\cdot, 0)\) is an element-wise operation that compares each element with 0, returning the element itself if it is greater than or equal to 0, and 0 otherwise.

\subsubsection{Hypervolume Calculation}
The HV contributions are calculated using the function \(f_{\text{hv}}\), which sums the contributions from each sample:
\begin{equation}
    f_{\text{hv}}(\bm{t}_\text{pds}) = \sum_{i=1}^{s} \left(\bm{\alpha}[\delta] \odot \mathds{1}_{\bm{t}_\text{pds} \neq -1}\right)_i,
\end{equation}
where \(\bm{\delta} = \bm{t}_\text{pds} \odot \bm{v}_\text{ds} - (1-\bm{t}_\text{pds})\).
This function is then applied in parallel to each row of the point dominance score tensor \(\bm{T}_\text{pds}\) using \(\texttt{vmap}\), which efficiently computes the HV contributions across all samples.

Finally, the total HV \(\bm{v}_\text{hv}\) is obtained by aggregating these contributions and normalizing:
\begin{equation}
    \bm{v}_\text{hv} = \bm{v}_\text{hv} \cdot \prod_{i=1}^{m} \left(\bm{v}_{\text{ref},i} - \bm{f}_{l,i}\right) / s.
\end{equation}

For other indicator-based EMO algorithms, the indicator is typically calculated using mathematical expressions, which facilitates straightforward tensorization. In the case of more complex indices, such as HV, Monte Carlo sampling can be employed to enhance algorithm efficiency by approximating these values. Additionally, to improve efficiency and scalability, the niche selection phase can use a one-shot method~\cite{hype} for parallel selection.

\begin{figure*}[htbp]
    \centering
    \includegraphics[width=0.9\textwidth]{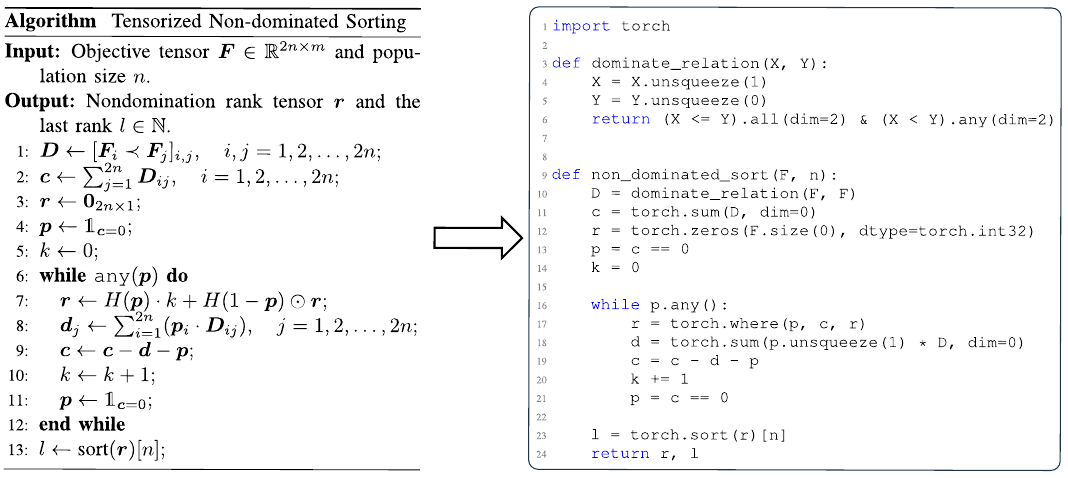}
    % \captionsetup{skip=-0.5pt}
    \caption{The seamless transformation of the tensorized nondominated sorting from pseudocode (Left) to Python code (Right).}
    \label{fig:ndsort_transfer}
    % \vspace{-5pt}
\end{figure*}

\subsection{Discussion}
The tensorization methodology offers unique transformative benefits in EMO algorithm design and implementation. One of the key advantages is the seamless transformation of EMO algorithms from mathematical formulations into efficient code implementations. This bridge between algorithm design and programming is particularly valuable in GPU computing, where tensor operations can be leveraged for significant performance gains. 
For instance, Fig.~\ref{fig:ndsort_transfer} illustrates how pseudocode for tensorized nondominated sorting can be directly translated into Python code.
This straightforward translation reduces the gap between high-level algorithmic design and practical implementation, thereby enabling practitioners to focus more on the theoretical aspects without the burden of intricate code optimization.

Tensorization also provides a high level of conciseness, significantly reducing code complexity compared to traditional iterative pseudocode.
By representing population-wide operations as single tensor expressions, tensorization minimizes the need for loops and conditional statements, making the codebase more compact and readable. 
This conciseness not only eases code maintenance but also reduces the risk of programming errors by limiting procedural complexity.

Moreover, tensorization facilitates reproducibility in the field of EMO.
As tensorized code relies on structured mathematical expressions, it becomes easier for researchers and developers to replicate results and benchmark different methods. 
This standardization paves the way for creating robust and high-performance libraries that can be shared and reused across various applications, thereby ultimately advancing research and industrial applications in EMO.

\section{Multiobjective Robot Control Benchmark}
\label{sec:mrcb}

Traditional EMO benchmarks, such as ZDT~\cite{ZDT}, DTLZ~\cite{DTLZ}, WFG~\cite{WFG}, LSMOP~\cite{lsmop_problem}, and MaF~\cite{MaF}, primarily focus on numerical optimization problems. While these benchmarks are effective for evaluating the basic mechanisms of EMO algorithms, they are limited in their ability to leverage hardware acceleration, thereby reducing their relevance in GPU computing environments. 

In contrast, multiobjective robot control tasks present more realistic and computationally intensive challenges that better reflect real-world applications, which is particularly significant in the emerging area of \emph{embodied artificial intelligence} (Embodied AI)~\cite{EmbodiedAI}.
These tasks provide complex and dynamic environments where multiple objectives must be balanced, such as energy efficiency and stability, making them well-suited for testing the adaptability and robustness of EMO algorithms. However, these tasks have been underexplored in the EMO community due to a lack of suitable benchmarks and the significant computational cost of running these environments.

To address this gap, we introduce the multiobjective robot control benchmark test suite, dubbed \emph{MoRobtrol}, which reformulates nine tasks from the Brax environment~\cite{brax} into MOPs.
As a GPU-accelerated physics simulation engine, Brax provides a substantial performance improvement over CPU-based platforms such as OpenAI Gym~\cite{openai-gym} and Mo-Gymnasium~\cite{mogym}. 
By leveraging Brax's GPU computing capabilities, MoRobtrol enables scalable and rapid evaluations, making it an ideal benchmark for testing EMO algorithms in computationally demanding settings in practice.

As illustrated in Fig. S.1, the MoRobtrol benchmark includes nine robot control tasks: MoHalfcheetah, MoHopper, MoSwimmer, MoInvertedDoublePendulum (MoIDP), MoWalker2d, MoPusher, MoReacher, MoHumanoid, and MoHumanoidStandup (MoHumanoid-s). These tasks involve optimizing multiple conflicting objectives, such as speed, energy consumption, and distance to target, reflecting trade-offs commonly encountered in robotics applications.
Specifically, in MoRobtrol, the parameters being optimized are the weights of a multilayer perceptron (MLP), a common design in control policy modeling for evolutionary reinforcement learning (EvoRL)~\cite{evorl}. 
These parameters are optimized by EMO algorithms to enable agents to maximize performance across conflicting objectives.

Table~\ref{tab:mopset} provides an overview of the nine tasks, including the MLP structure, number of parameters (\(d\)), and specific objectives for each task. 
The key objectives across tasks include forward reward (\(f_\text{v}\)), control cost (\(f_\text{c}\)), height (\(f_\text{h}\)), distance penalty (\(f_\text{dp}\)), speed penalty (\(f_\text{sp}\)), distance reward (\(f_\text{d}\)), and near reward (\(f_\text{n}\)). 
The number of objectives \(m\) varies depending on the task, allowing for detailed evaluations of algorithmic performance in diverse, real-world-inspired control scenarios.
Detailed mathematical definitions are provided in Section S.VI of the Supplementary Document. 

\begin{table}[t]
    \centering
    \caption{Overview of Multiobjective Robot Control Problems in the Proposed MoRobtrol Benchmark Test Suite}
    \label{tab:mopset}
    \begin{tabular}{cccc} 
        \toprule
        Problem & MLP architecture\(^\dagger\) & \(d\) & \(m\) \\
        \midrule
        MoHalfcheetah & \(17\times16\times6\) & 390 & \(f_\text{v}\), \(f_\text{c}\) \\
        MoHopper      & \(11\times16\times3\) & 243 & \(f_\text{v}\), \(f_\text{h}\), \(f_\text{c}\) \\
        MoSwimmer     & \(8\times16\times2\) & 178 & \(f_\text{v}\), \(f_\text{c}\) \\
        MoIDP         & \(8\times16\times1\) & 161 & \(f_\text{dp}\), \(f_\text{sp}\) \\
        MoWalker2d    & \(17\times16\times6\) & 390 & \(f_\text{v}\), \(f_\text{c}\) \\
        MoPusher      & \(23\times16\times7\) & 503 & \(f_\text{n}\), \(f_\text{d}\), \(f_\text{c}\) \\
        MoReacher     & \(11\times16\times2\) & 226 & \(f_\text{d}\), \(f_\text{c}\) \\
        MoHumanoid    & \(244\times16\times17\) & 4209 & \(f_\text{v}\), \(f_\text{c}\) \\
        MoHumanoid-s  & \(244\times16\times17\) & 4209 & \(f_\text{v}\), \(f_\text{c}\) \\
        \bottomrule
        \multicolumn{4}{l}{\footnotesize \(^\dagger\) All MLP networks use the \texttt{tanh} activation function.}
    \end{tabular}
    \vspace{-5mm}
\end{table}

\section{Experimental Study}
\label{sec:experiments}
In this section, we conduct experiments to evaluate the performance of the tensorized EMO algorithms, including TensorNSGA-III, TensorMOEA/D, and TensorHypE. The experiments are categorized into three main aspects: acceleration performance, benchmarking on standard EMO test problems, and evaluation in multiobjective robot control tasks. 
All experiments are conducted on an RTX 4090 GPU server with AMD EPYC 7543 CPUs using the EvoX~\cite{evox} framework.

\subsection{Acceleration Performance}
To verify the acceleration performance of the three proposed algorithms, we have conducted two sub-experiments. The first sub-experiment doubles the population size and observes the average runtime per generation. The second sub-experiment doubles the problem dimension and observes the average runtime per generation. We compare the performance of NSGA-III, MOEA/D, and HypE before and after tensorization on both CPU and GPU platforms using the DTLZ1~\cite{DTLZ} problem.

Additionally, further experiments have been conducted to compare the performance of the tensorized algorithms in comparison with CUDA-accelerated algorithms in EvoTorch~\cite{evotorch}, as well as to investigate the impact of different types of GPUs on performance. Detailed results of these experiments can be found in Section S.VII-B and Section S.VII-C of Supplementary Document, respectively.

\subsubsection{Experimental Settings}
In the two sub-experiments, the tensorized and non-tensorized\footnote{The non-tensorized algorithms are partially tensorized for efficiency, as the original versions are time-consuming for large populations.} algorithms are independently repeated 10 times on both CPU and GPU devices, with each algorithm evolving for 100 generations. The average runtime per generation is then calculated. In the first sub-experiment, the DTLZ1 problem has a dimension \(d = 500\), with \(m = 3\), and the population size \(n\) is doubled incrementally from 128 to 32768. In the second sub-experiment, the DTLZ1 problem is configured with \(m = 3\), \(n = 100\), and \(d\) is incrementally doubled from 1024 to 1048576.

\subsubsection{Comparison Results}
\begin{figure}[H]
    
    \centering
        \begin{subfigure}[b]{0.24\textwidth}
            \includegraphics[width=\linewidth]{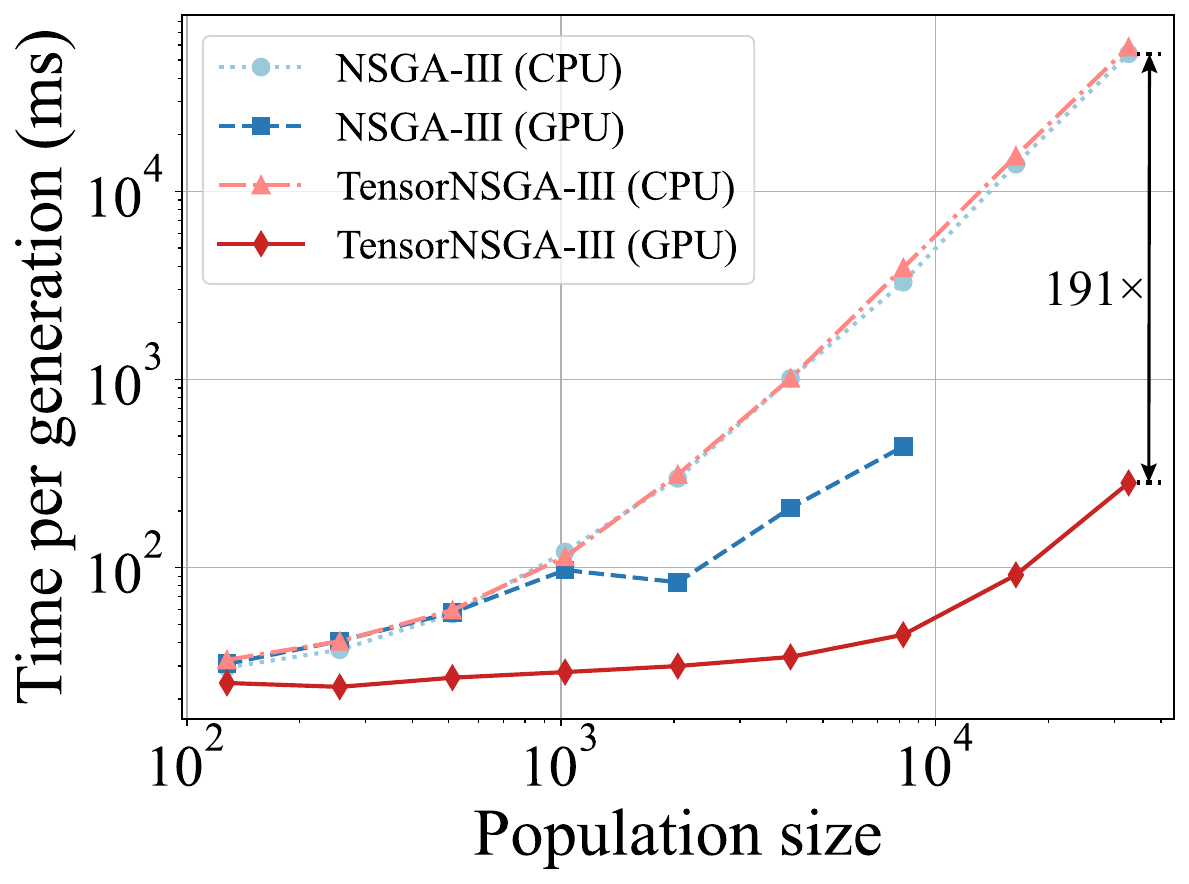}
            \captionsetup{skip=-0.5pt}
            \caption{}
            \label{fig:image31}
        \end{subfigure}
        \begin{subfigure}[b]{0.235\textwidth}
            \includegraphics[width=\linewidth]{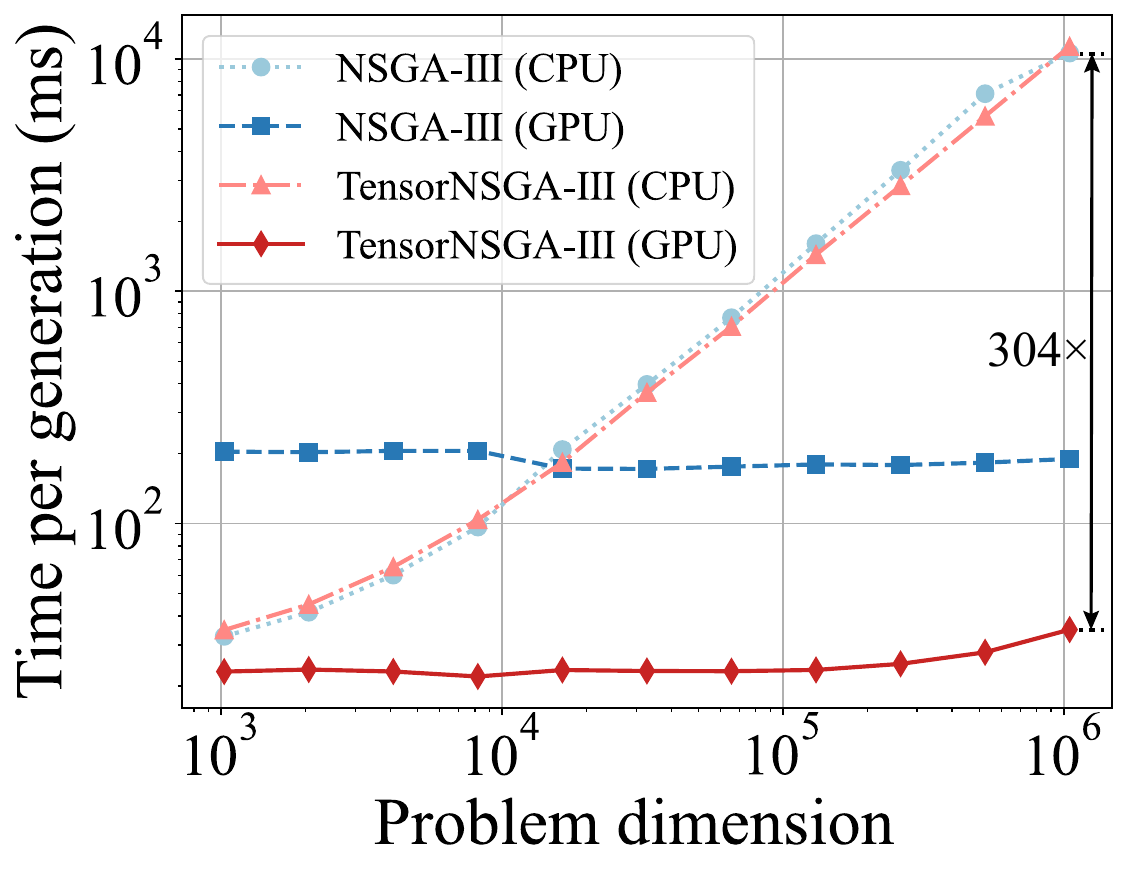}
            \captionsetup{skip=-0.5pt}
            \caption{}
            \label{fig:image32}
        \end{subfigure}

        \begin{subfigure}[b]{0.24\textwidth}
            \includegraphics[width=\linewidth]{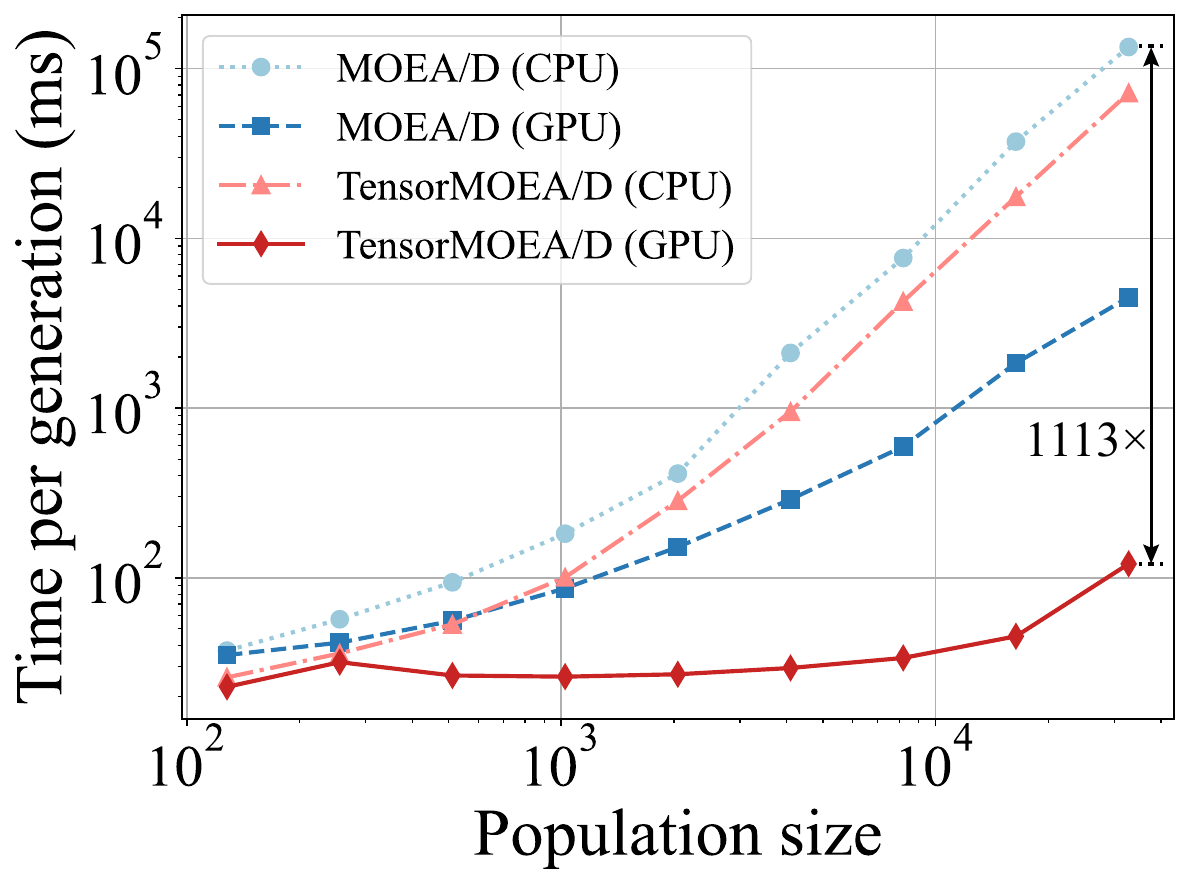}
            \captionsetup{skip=-0.5pt}
            \caption{}
            \label{fig:image33}
        \end{subfigure}
        \begin{subfigure}[b]{0.235\textwidth}
            \includegraphics[width=\linewidth]{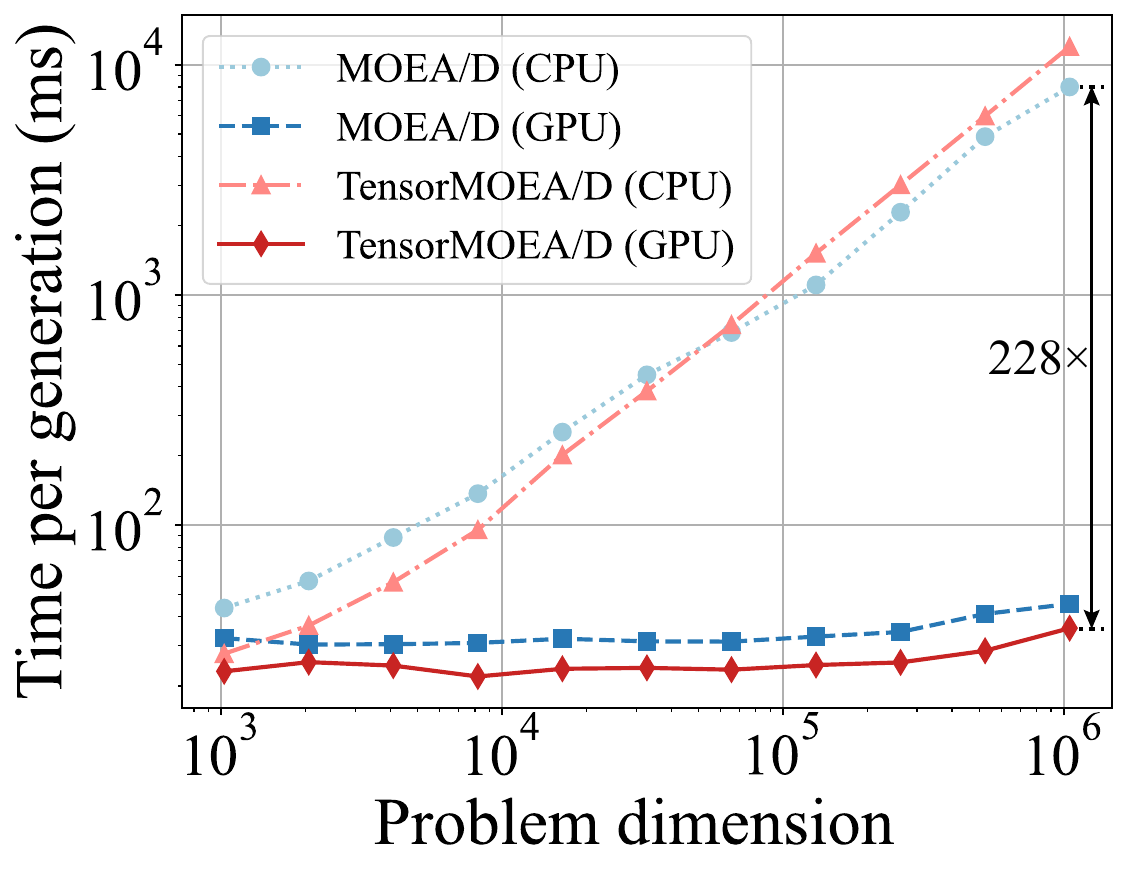}
            \captionsetup{skip=-0.5pt}
            \caption{}
            \label{fig:image34}
        \end{subfigure}

        \begin{subfigure}[b]{0.24\textwidth}
            \includegraphics[width=\linewidth]{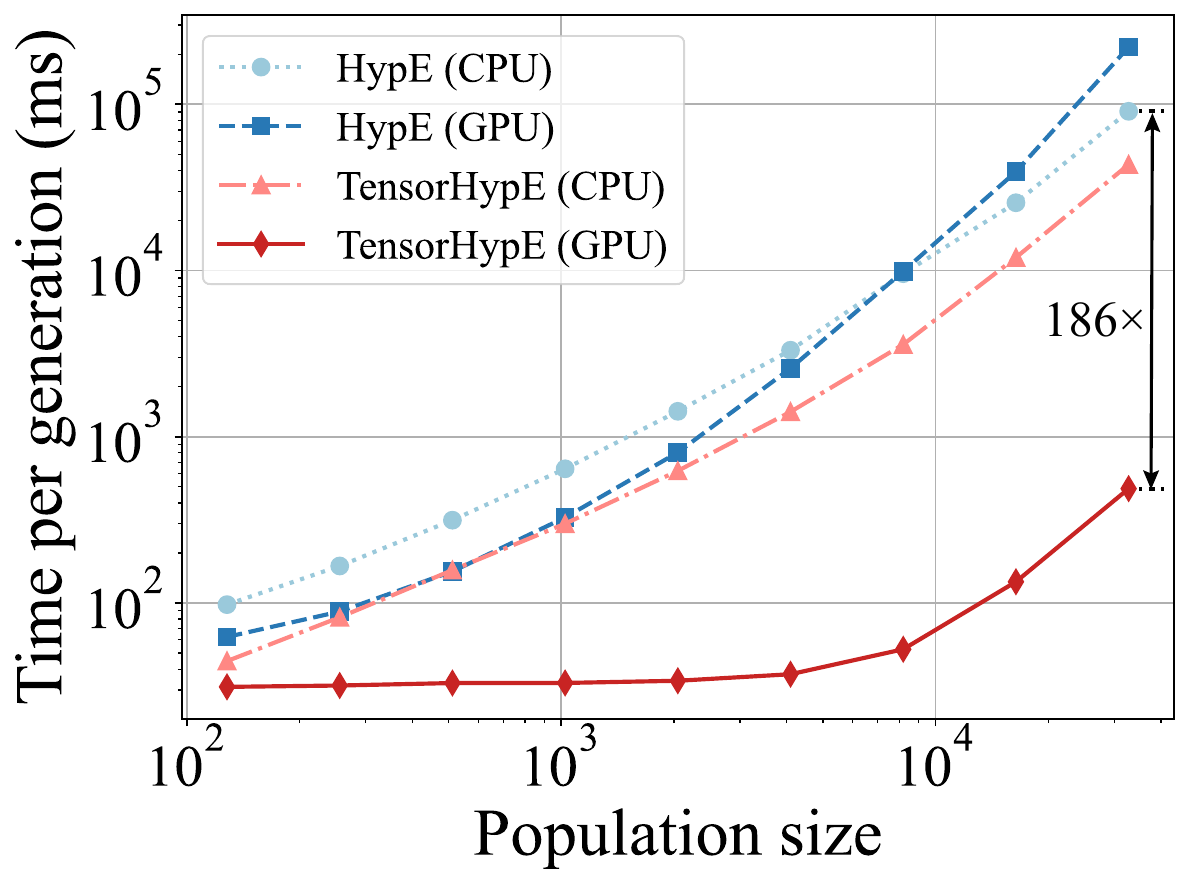}
            \captionsetup{skip=-0.5pt}
            \caption{}
            \label{fig:image35}
        \end{subfigure}
        \begin{subfigure}[b]{0.235\textwidth}
            \includegraphics[width=\linewidth]{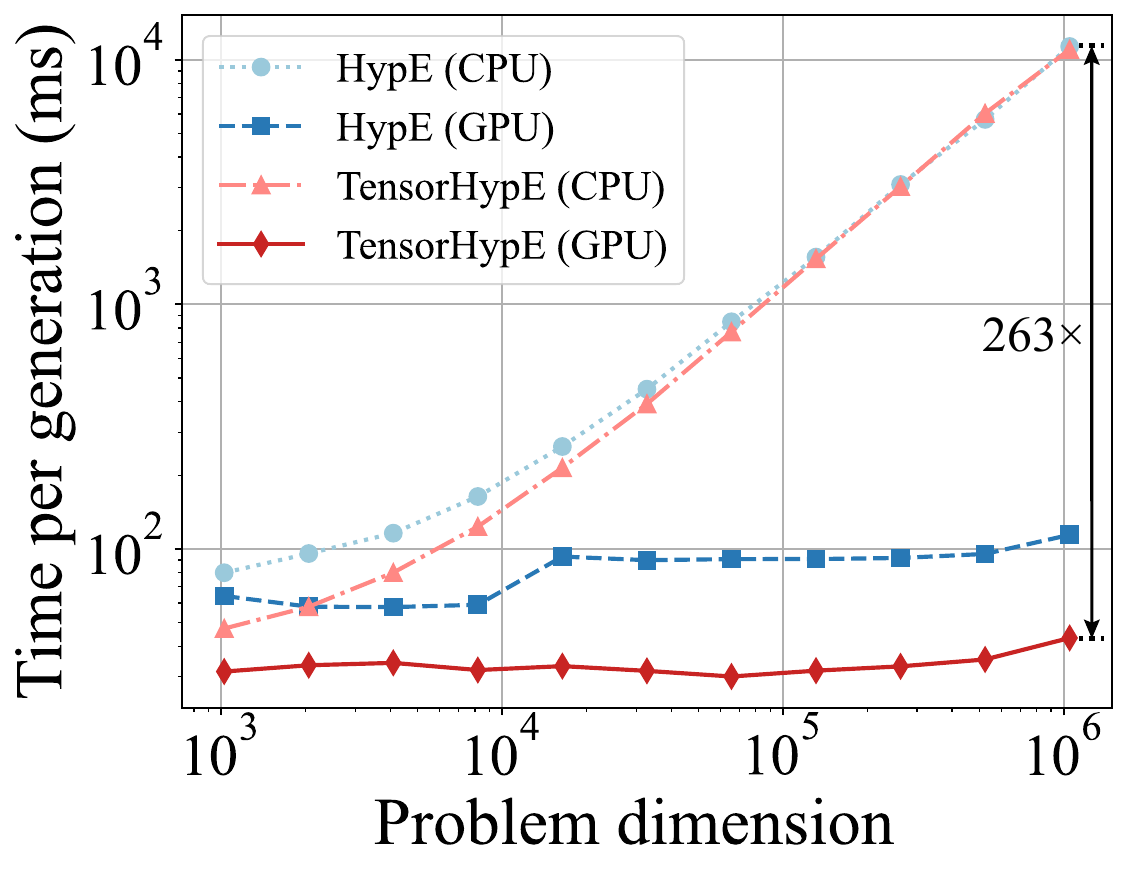}
            \captionsetup{skip=-0.5pt}
            \caption{}
            \label{fig:image36}
        \end{subfigure}
    \captionsetup{skip=-0.5pt}
    \caption{Comparative acceleration performance of NSGA-III, MOEA/D, and HypE with their tensorized counterparts on CPU and GPU platforms. (a) and (b) illustrate performance for NSGA-III, (c) and (d) for MOEA/D, and (e) and (f) for HypE, across varying population sizes and problem dimensions.}
    \label{fig:performences}
\end{figure}

Fig.~\ref{fig:performences} shows that TensorNSGA-III, TensorMOEA/D, and TensorHypE consistently achieve faster runtimes on GPU compared to their non-tensorized versions on CPUs. When the population size \(n\) reaches 32768, TensorNSGA-III, TensorMOEA/D, and TensorHypE attain speedups of approximately 191\(\times\), 1113\(\times\), and 186\(\times\), respectively, compared to their CPU-based counterparts. As the problem dimension increases to 1048576, these speedups rise to 304\(\times\), 228\(\times\), and 263\(\times\). Although the runtime of tensorized algorithms may increase with \(n\), they consistently outperform the non-tensorized versions. Additionally, as problem dimensions scale up, the runtime of tensorized algorithms remains relatively stable, maintaining a significant performance advantage over the original algorithms.

% \begin{figure}[htbp]
%     \centering

%     \subfloat[]{%
%         \includegraphics[width=0.24\textwidth]{figures/NSGA3_DTLZ1_pop.pdf}%
%         \label{fig:image31}}
%     \hfill
%     \subfloat[]{%
%         \includegraphics[width=0.235\textwidth]{figures/NSGA3_DTLZ1_dim.pdf}%
%         \label{fig:image32}}
%     \hfill
%     \vspace{-2ex}
%     \subfloat[]{%
%         \includegraphics[width=0.24\textwidth]{figures/TensorMOEAD_DTLZ1_pop.pdf}%
%         \label{fig:image33}}
%     \hfill
%     \subfloat[]{%
%         \includegraphics[width=0.235\textwidth]{figures/TensorMOEAD_DTLZ1_dim.pdf}%
%         \label{fig:image34}}
%     \hfill
%     \vspace{-2ex}
%     \subfloat[]{%
%         \includegraphics[width=0.24\textwidth]{figures/HypE_DTLZ1_pop.pdf}%
%         \label{fig:image35}}
%     \hfill
%     \subfloat[]{%
%         \includegraphics[width=0.235\textwidth]{figures/HypE_DTLZ1_dim.pdf}%
%         \label{fig:image36}}
%     \vspace{-1ex}
%     \caption{\normalsize Comparative acceleration performance of NSGA-III, MOEA/D, and HypE with their tensorized counterparts on CPU and GPU platforms across varying population sizes and problem dimension.}
%     \label{fig:performences}
% \end{figure}

\begin{table*}[htb]
    \centering
    \caption{Statistical Results (Mean and Standard Deviation) of the IGD and Runtime (s) for Non-Tensorized and Tensorized EMO Algorithms in LSMOP1--LSMOP9. All Experiments Are on an RTX 4090 GPU and the Best Results Are Highlighted.}
    \label{table:precision}
    \begin{tabular}{c c c c c c}
    \toprule
    \textbf{Algorithm} & \textbf{Problem} & \textbf{IGD (Non-Tensorized)} & \textbf{IGD (Tensorized)} & \textbf{Time (Non-Tensorized)} & \textbf{Time (Tensorized)} \\
    \midrule
    \multirow{9}{*}{\textbf{NSGA-III}} 
    & LSMOP1 & \textbf{8.2378e$-$01 (5.2173e$-$03)} & 1.0399e$+$00 (1.2327e$-$01) & 3.5048e$+$03 (2.7933e$+$00) & \textbf{7.8573e$+$00 (2.0971e$-$01)} \\
    & LSMOP2 & \textbf{3.5883e$-$01 (5.9506e$-$06)} & \textbf{3.5883e$-$01 (2.0165e$-$05)} & 6.0457e$+$01 (1.2916e$+$00) & \textbf{1.6127e$+$00 (1.8569e$-$02)} \\
    & LSMOP3 & \textbf{1.4106e$+$00 (7.5580e$-$02)} & 7.3244e$+$00 (5.3558e$-$01) & 6.1996e$+$01 (1.6457e$+$00) & \textbf{2.4987e$+$00 (3.4992e$-$02)} \\
    & LSMOP4 & \textbf{5.9798e$-$01 (1.5968e$-$04)} & 5.9815e$-$01 (2.0699e$-$04) & 6.5350e$+$01 (1.8958e$+$00) & \textbf{1.7237e$+$00 (1.5717e$-$02)} \\
    & LSMOP5 & \textbf{5.6862e$-$01 (2.3841e$-$03)} & 6.4269e$-$01 (6.6591e$-$03) & 6.0026e$+$01 (1.0193e$+$00) & \textbf{7.0027e$+$00 (8.2602e$-$02)} \\
    & LSMOP6 & \textbf{2.8277e$+$00 (5.9075e$-$01)} & 2.3737e$+$01 (2.9479e$+$00) & 6.3764e$+$01 (1.8220e$+$00) & \textbf{3.5406e$+$00 (5.6013e$-$02)} \\
    & LSMOP7 & \textbf{1.8425e$+$00 (7.5222e$-$03)} & 1.8474e$+$00 (3.7155e$-$03) & 6.2175e$+$01 (1.2735e$+$00) & \textbf{5.0992e$+$00 (1.3959e$-$01)} \\
    & LSMOP8 & \textbf{3.0317e$-$01 (2.9719e$-$02)} & 3.5186e$-$01 (1.3261e$-$02) & 6.6262e$+$01 (1.4221e$+$00) & \textbf{5.6878e$+$00 (4.9742e$-$02)} \\
    & LSMOP9 & \textbf{7.4488e$-$01 (9.4755e$-$03)} & 7.6837e$-$01 (2.9024e$-$03) & 6.1194e$+$01 (1.6353e$+$00) & \textbf{3.4327e$+$00 (2.2911e$-$02)} \\
    \midrule
    \multirow{9}{*}{\textbf{MOEA/D}} 
    & LSMOP1 & \textbf{7.3475e$-$01 (3.8276e$-$04)} & 7.4365e$-$01 (2.1169e$-$03) & 9.4239e$+$01 (3.0883e$+$00) & \textbf{5.4412e$+$00 (2.8179e$-$01)} \\
    & LSMOP2 & \textbf{3.5885e$-$01 (3.4131e$-$05)} & 3.5888e$-$01 (5.9703e$-$06) & 9.9946e$+$01 (2.7744e$+$00) & \textbf{5.7516e$+$00 (6.0279e$-$01)} \\
    & LSMOP3 & 1.0815e$+$00 (1.0631e$-$01) & \textbf{8.2722e$-$01 (5.3689e$-$02)} & 1.0148e$+$02 (2.6072e$+$00) & \textbf{7.0391e$+$00 (4.7843e$-$01)} \\
    & LSMOP4 & 5.9570e$-$01 (1.5407e$-$03) & \textbf{5.9420e$-$01 (1.1813e$-$03)} & 1.0919e$+$02 (2.2606e$+$00) & \textbf{7.1559e$+$00 (3.8251e$-$01)} \\
    & LSMOP5 & 4.0554e$-$01 (8.8890e$-$03) & \textbf{3.7065e$-$01 (8.5792e$-$03)} & 9.7131e$+$01 (2.9509e$+$00) & \textbf{7.1893e$+$00 (4.6098e$-$01)} \\
    & LSMOP6 & \textbf{1.6366e$+$00 (7.7256e$-$02)} & 1.8622e$+$00 (1.2612e$-$01) & 9.6220e$+$01 (2.9977e$+$00) & \textbf{6.8919e$+$00 (2.4831e$-$01)} \\
    & LSMOP7 & \textbf{1.5534e$+$00 (2.6469e$-$01)} & 1.7345e$+$00 (3.6746e$-$03) & 1.0456e$+$02 (2.4456e$+$00) & \textbf{7.0900e$+$00 (4.0993e$-$01)} \\
    & LSMOP8 & 2.1207e$-$01 (6.4136e$-$03) & \textbf{2.0107e$-$01 (1.8164e$-$02)} & 1.0154e$+$02 (2.0132e$+$00) & \textbf{7.1273e$+$00 (4.2536e$-$01)} \\
    & LSMOP9 & 5.1964e$-$01 (1.7163e$-$02) & \textbf{5.0269e$-$01 (1.0548e$-$02)} & 1.0443e$+$02 (2.9910e$+$00) & \textbf{7.2184e$+$00 (7.7100e$-$01)} \\
    \midrule
    \multirow{9}{*}{\textbf{HypE}} 
    & LSMOP1 & \textbf{8.2757e$-$01 (1.2610e$-$03)} & \textbf{8.2747e$-$01 (1.0132e$-$03)} & 6.3918e$+$01 (6.1344e$-$01) & \textbf{1.8814e$+$00 (9.9671e$-$02)} \\
    & LSMOP2 & \textbf{3.5912e$-$01 (5.5267e$+$00)} & \textbf{3.5912e$-$01 (4.9532e$-$05)} & 6.4140e$+$01 (4.4810e$-$01) & \textbf{1.5211e$+$00 (2.8512e$-$02)} \\
    & LSMOP3 & \textbf{4.6906e$+$00 (3.2754e$-$01)} & \textbf{4.6906e$+$00 (3.2754e$-$01)} & 6.4090e$+$01 (4.1599e$-$01) & \textbf{1.7588e$+$00 (4.3342e$-$02)} \\
    & LSMOP4 & \textbf{5.9841e$-$01 (2.3223e$-$04)} & \textbf{5.9847e$-$01 (2.5082e$-$04)} & 6.3900e$+$01 (2.4586e$-$01) & \textbf{1.6785e$+$00 (4.7110e$-$02)} \\
    & LSMOP5 & \textbf{5.2160e$-$01 (1.3777e$-$03)} & \textbf{5.2153e$-$01 (1.6406e$-$03)} & 6.4216e$+$01 (2.9358e$-$01) & \textbf{2.0728e$+$00 (1.6839e$-$02)} \\
    & LSMOP6 & \textbf{3.0407e$+$00 (1.2805e$-$01)} & \textbf{3.0267e$+$00 (1.1215e$-$01)} & 6.5536e$+$01 (3.2336e$-$01) & \textbf{3.4693e$+$00 (7.3326e$-$02)} \\
    & LSMOP7 & \textbf{1.8223e$+$00 (2.5992e$-$03)} & \textbf{1.8220e$+$00 (2.5472e$-$03)} & 6.4099e$+$01 (2.5939e$-$01) & \textbf{1.7594e$+$00 (2.9685e$-$02)} \\
    & LSMOP8 & \textbf{3.6062e$-$01 (9.5863e$-$03)} & \textbf{3.6060e$-$01 (9.5862e$-$03)} & 6.4537e$+$01 (3.3889e$-$01) & \textbf{2.3856e$+$00 (2.8679e$-$02)} \\
    & LSMOP9 & \textbf{6.5873e$-$01 (3.0162e$-$03)} & \textbf{6.5797e$-$01 (3.4461e$-$03)} & 6.3643e$+$01 (2.6093e$-$01) & \textbf{1.5741e$+$00 (1.2765e$-$02)} \\
    \bottomrule
    \end{tabular}
    
\end{table*}

Notably, when \(n\) exceeds 16384, the runtime for NSGA-III on GPU surpasses the preset threshold of 5 hours, resulting in missing data points for larger populations. For the HypE algorithm, once \(n\) exceeds 1024, its average runtime per generation on GPU begins to exceed that of TensorHypE on CPU, and for \(n > 8192\), the GPU version of HypE becomes even slower than the original CPU-based implementation. 

These performance drops can be attributed to two main factors. First, the original implementations of NSGA-III and HypE do not fully leverage the multicore parallel processing capabilities of GPUs, leading to underutilization of GPU cores. Second, the data transfer overhead between CPU and GPU further reduces efficiency, particularly for large population sizes where the NSGA-III and HypE algorithms incur higher computational costs on GPU than on CPU.

Additionally, the acceleration performance varies across tensorized algorithms. MOEA/D and HypE, due to their simpler operations and fewer conditional branches, achieve greater acceleration after tensorization compared to their original versions. Conversely, NSGA-III shows more limited acceleration, as its more complex operations involve intricate loops and branches, which are less amenable to GPU parallelization.

\subsection{Performance in Numerical Optimization}
To verify the precision before and after tensorization, the three proposed algorithms and their original versions are comprehensively tested on the LSMOP~\cite{lsmop_problem} and DTLZ~\cite{DTLZ} test suites. The detailed results for the DTLZ are provided in Section S.VII-D of Supplementary Document.

\subsubsection{Experimental Settings}
In this experiment, all algorithms are independently repeated 31 times on 9 LSMOP problems and 7 DTLZ problems. Each algorithm is run for 100 generations with a population size of 10000. Each problem has a dimension of 5000. Performance is measured using the average IGD~\cite{igd} and average runtime over 31 runs. A Wilcoxon rank-sum test is used to compare tensorized and non-tensorized algorithms. If the test shows no significant difference (i.e., \(p > 0.05\)), both performance indicators are highlighted in bold.

% and Table S.I
\subsubsection{Comparison Results}
Table~\ref{table:precision} demonstrates that the tensorized algorithms maintain comparable precision to their non-tensorized counterparts while achieving significantly faster runtimes. The average IGD between the tensorized and non-tensorized algorithms consistently remain within the same order of magnitude, indicating similar levels of solution quality. In some instances, the tensorized algorithms even exhibit superior indicator performance. Additionally, the average runtime for tensorized algorithms is consistently lower than that for non-tensorized algorithms, underscoring the efficiency gains enabled by tensorization. Notably, when comparing performance within equivalent time frames, tensorized algorithms consistently achieve better indicator performance, highlighting their effectiveness in optimizing both time and solution quality metrics.

However, some performance degradation is observed in certain cases. In TensorNSGA-III, batch random operations simulate the niche selection process of the original algorithm, which differs from the original method’s precise operation on individual solutions. This discrepancy can lead to performance degradation, as the batch approach may not capture niche selection as effectively. Similarly, TensorMOEA/D adopts batch offspring generation and updates, which accelerate computation per generation but may result in lower performance for the same number of generations. Despite this, TensorMOEA/D consistently achieves or sometimes surpasses the performance of the original algorithm when comparing performance over equivalent time periods. As for HypE, since its method of calculating the HV is largely similar to that of the original algorithm, the primary difference lies in computation speed.

\begin{figure*}[!htb]
    \centering
    
        \begin{subfigure}[b]{0.28\textwidth}
            \includegraphics[width=\linewidth]{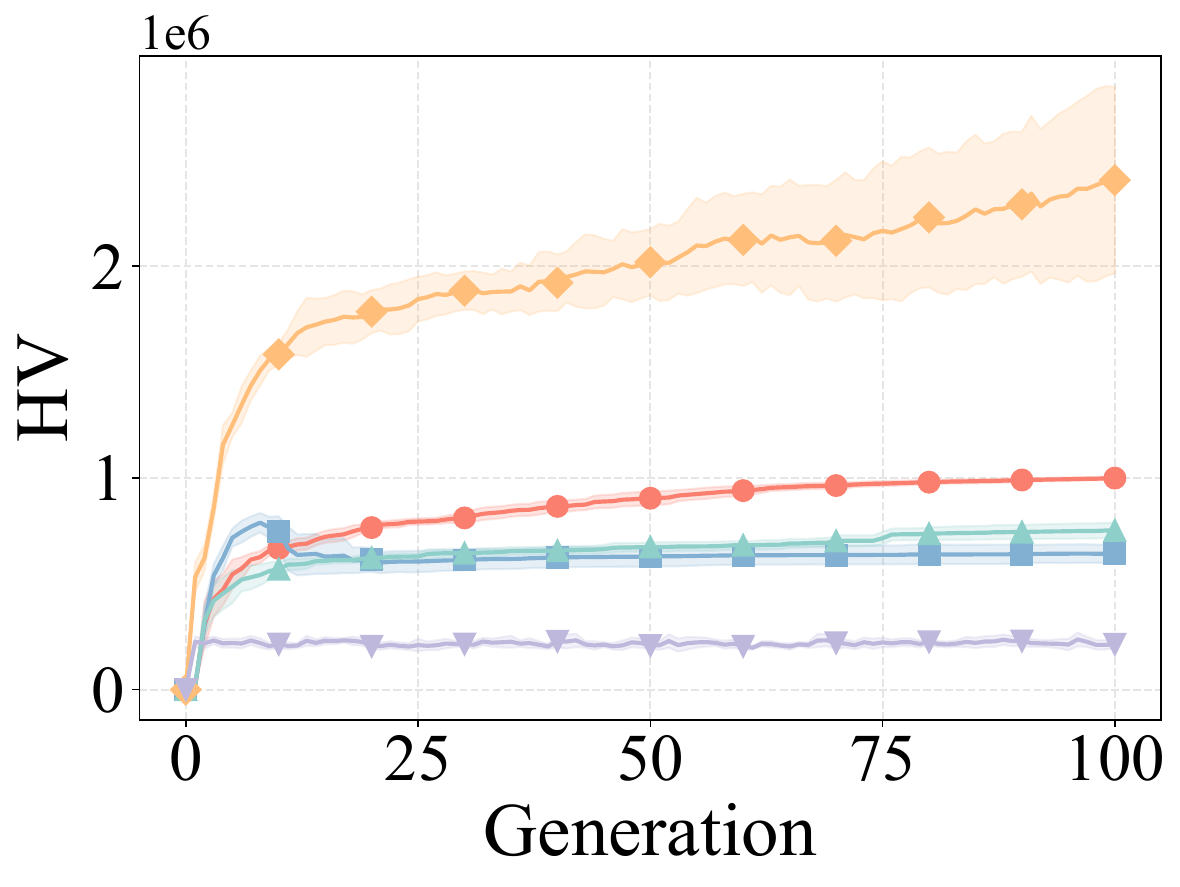}
            \captionsetup{skip=-0.5pt}
            \caption{HV on MoHalfcheetah}
            \label{fig:image1}
        \end{subfigure}
        \hspace{5mm}
        \begin{subfigure}[b]{0.28\textwidth}
            \includegraphics[width=\linewidth]{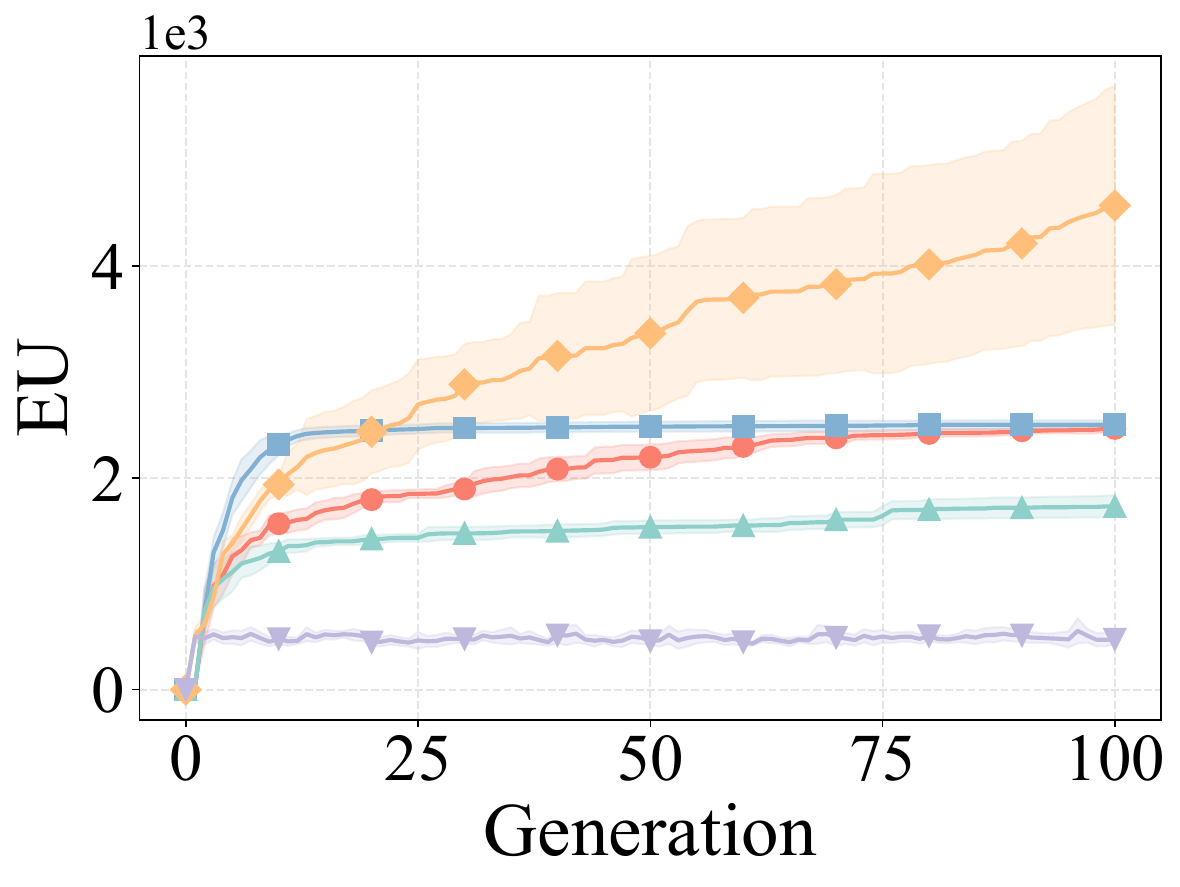}
            \captionsetup{skip=-0.5pt}
            \caption{EU on MoHalfcheetah}
            \label{fig:image2}
        \end{subfigure}
        \hspace{5mm}
        \begin{subfigure}[b]{0.28\textwidth}
            \includegraphics[width=\linewidth]{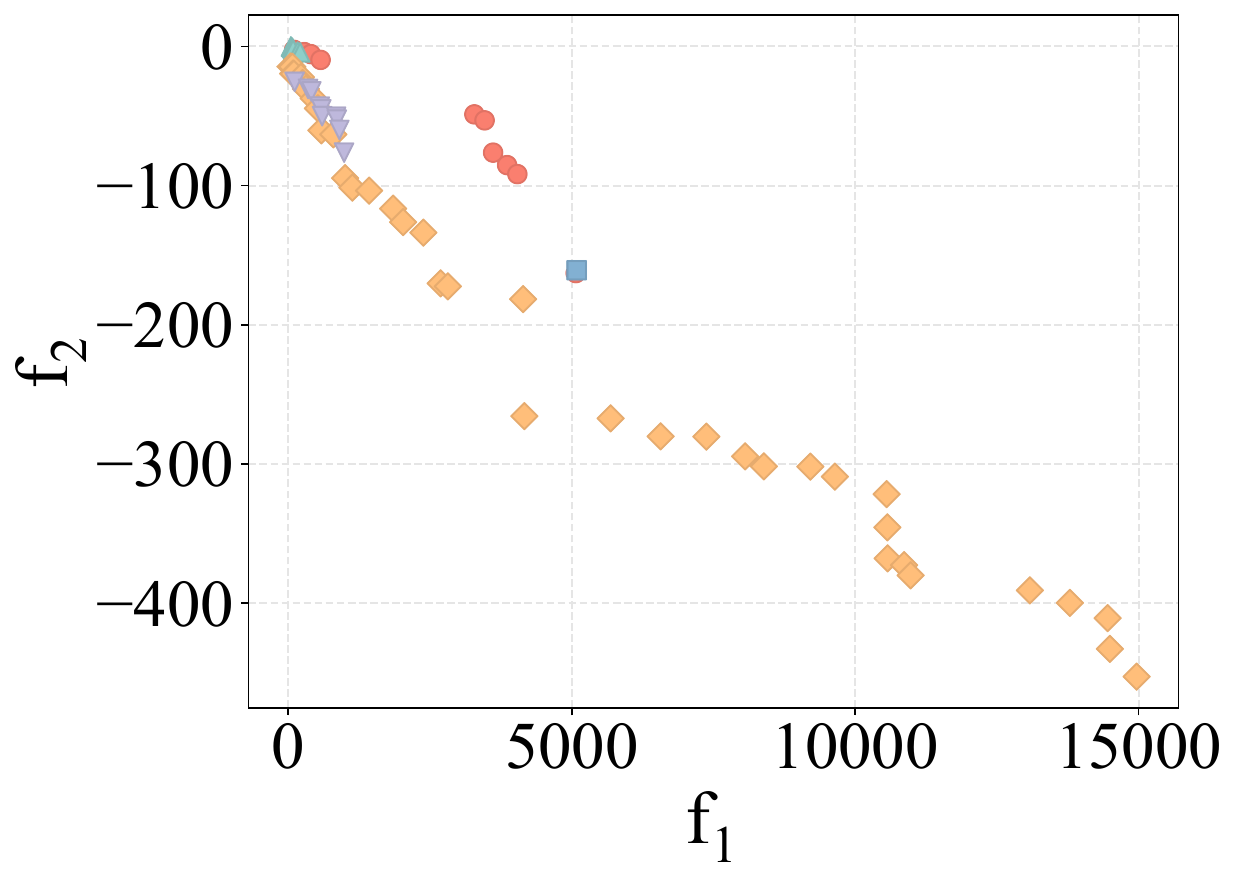}
            \captionsetup{skip=-0.5pt}
            \caption{Final results on MoHalfcheetah}
            \label{fig:image3}
        \end{subfigure}

        \begin{subfigure}[b]{0.27\textwidth}
            \includegraphics[width=\linewidth]{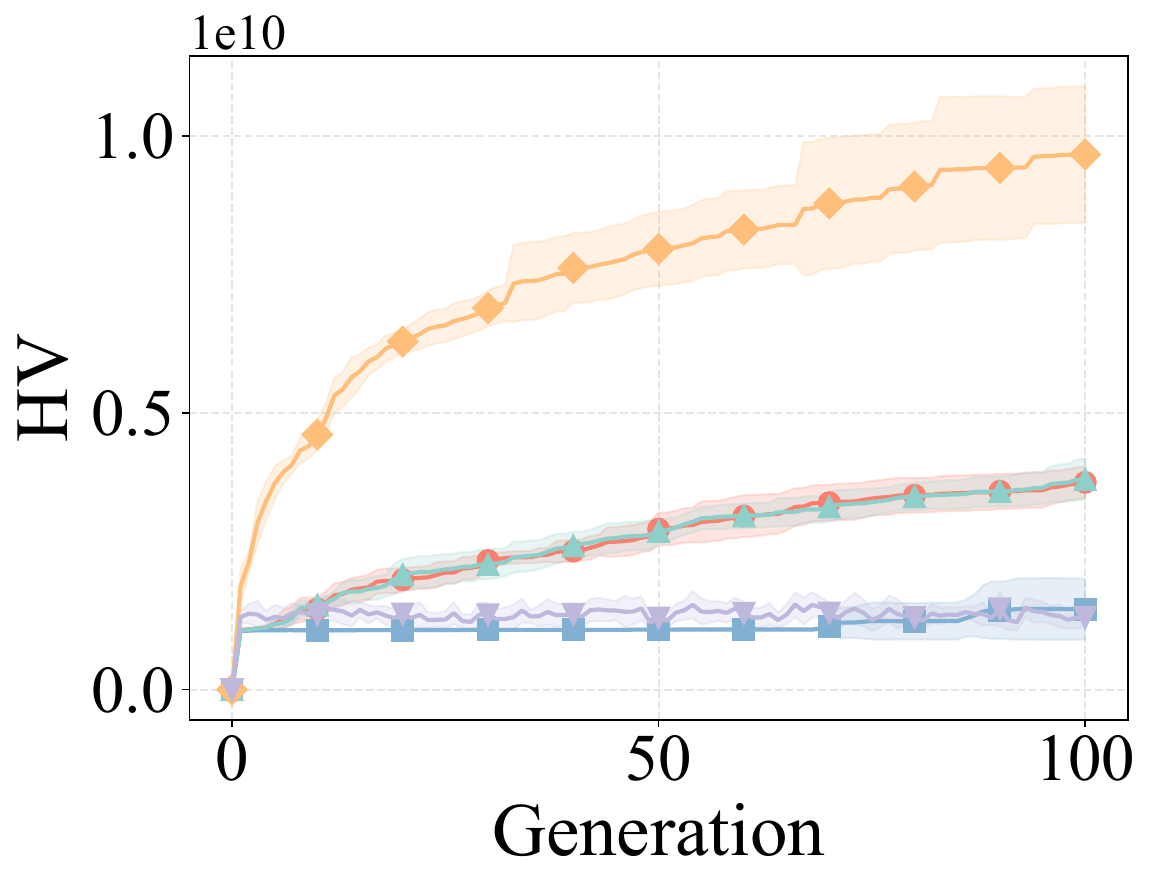}
            \captionsetup{skip=-0.5pt}
            \caption{HV on MoHopper}
            \label{fig:image4}
        \end{subfigure}
        \hspace{5mm}
        \begin{subfigure}[b]{0.28\textwidth}
            \includegraphics[width=\linewidth]{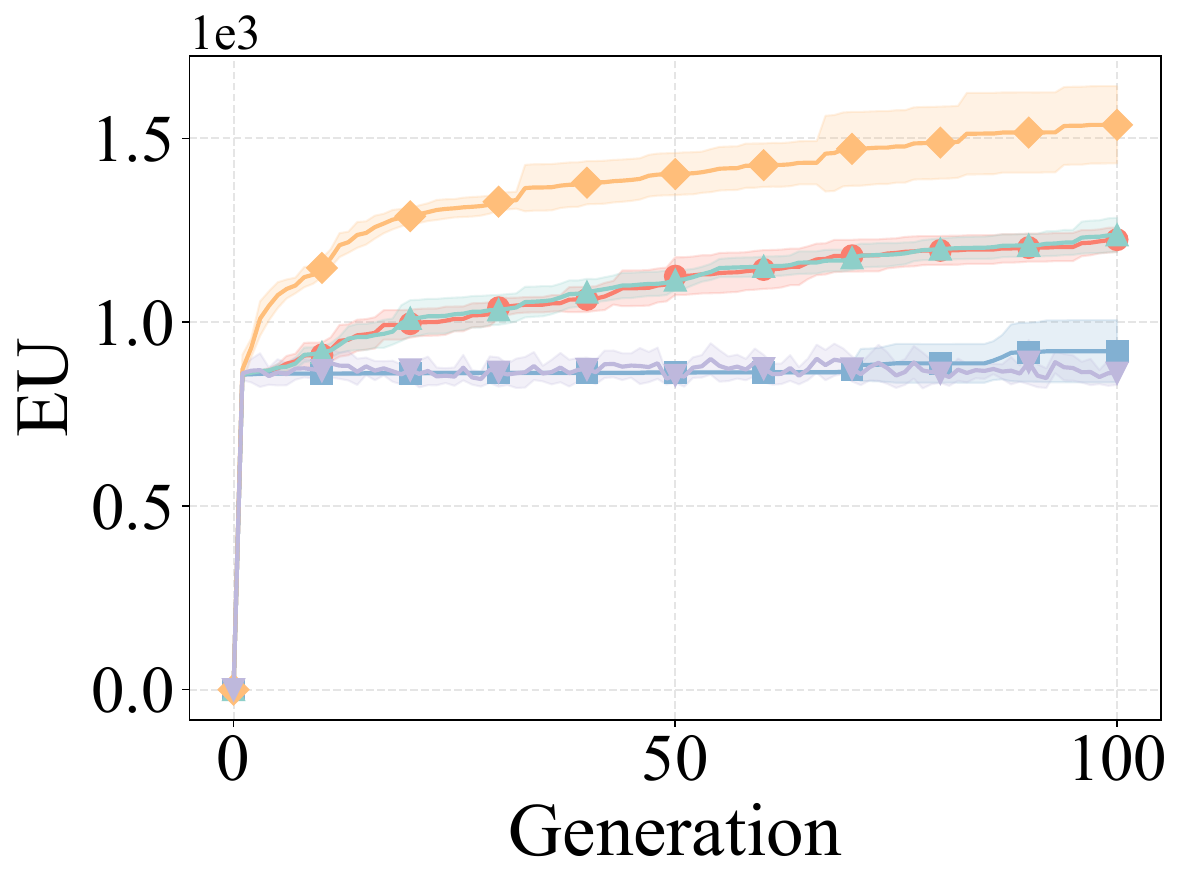}
            \captionsetup{skip=-0.5pt}
            \caption{EU on MoHopper}
            \label{fig:image5}
        \end{subfigure}
        \hspace{5mm}
        \begin{subfigure}[b]{0.28\textwidth}
            \includegraphics[width=\linewidth]{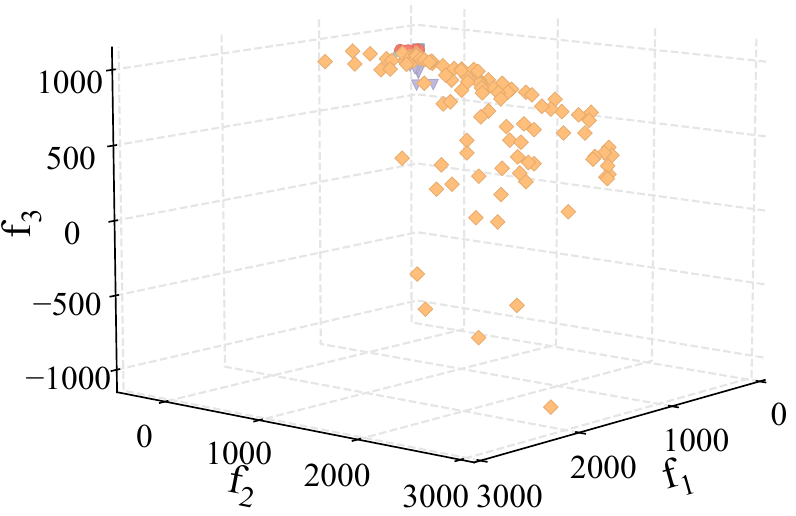}
            \captionsetup{skip=-0.5pt}
            \caption{Final results on MoHopper}
            \label{fig:image6}
        \end{subfigure}

        \begin{subfigure}[b]{0.27\textwidth}
            \includegraphics[width=\linewidth]{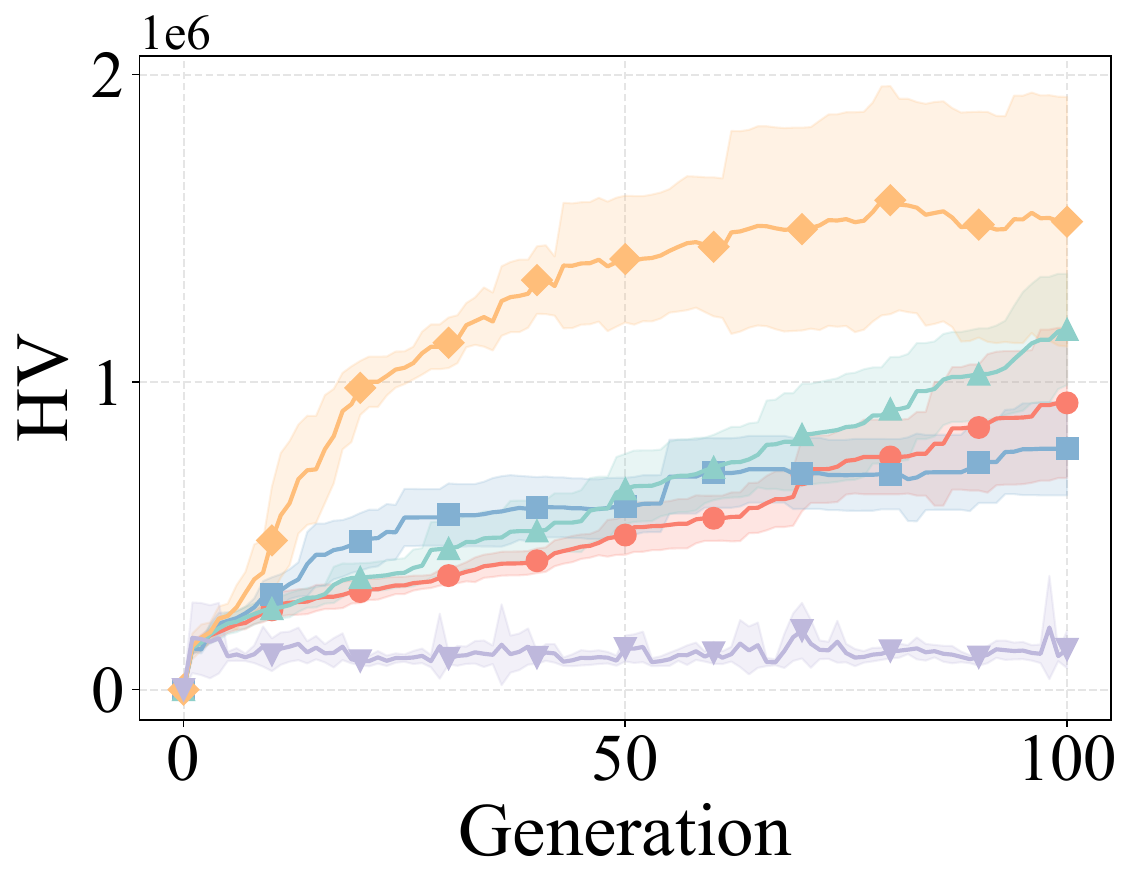}
            \captionsetup{skip=-0.5pt}
            \caption{HV on MoWalker2d}
            \label{fig:image28}
        \end{subfigure}
        \hspace{5mm}
        \begin{subfigure}[b]{0.28\textwidth}
            \includegraphics[width=\linewidth]{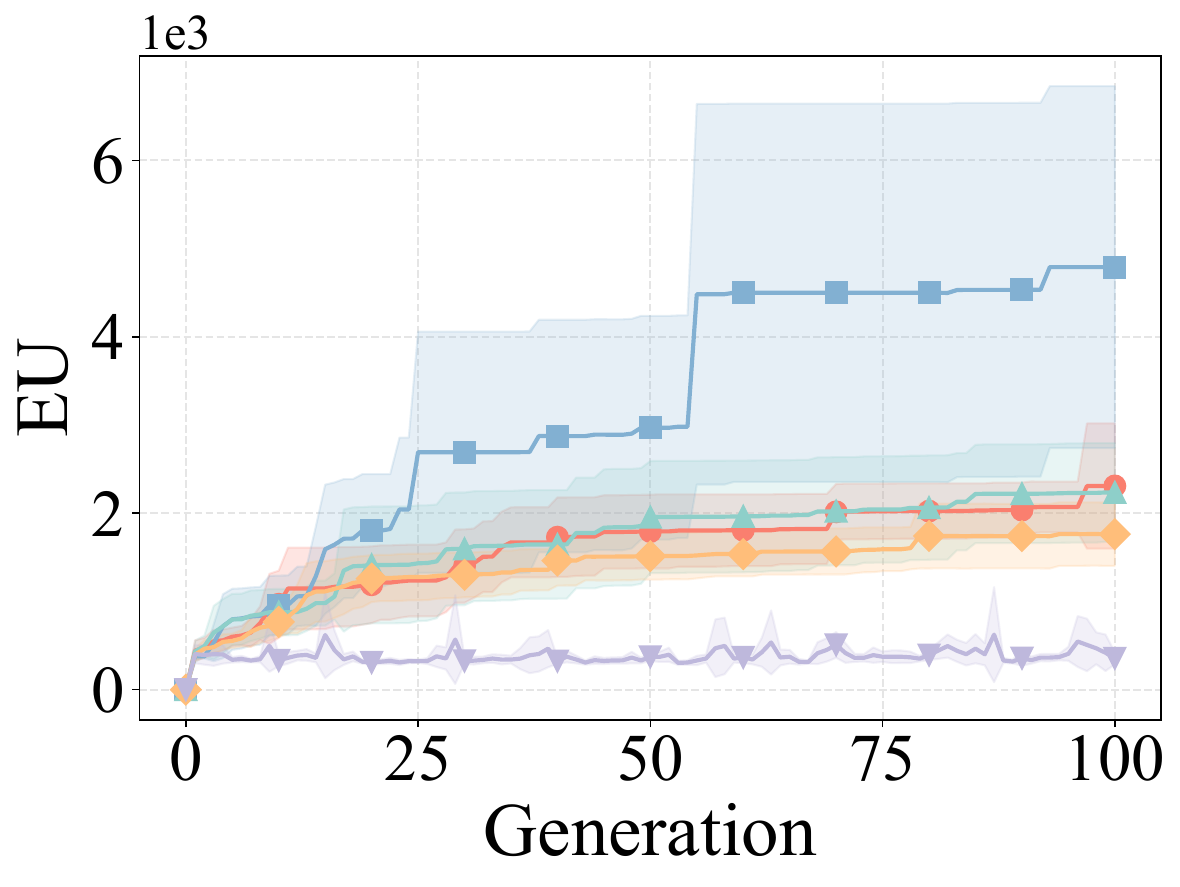}
            \captionsetup{skip=-0.5pt}
            \caption{EU on MoWalker2d}
            \label{fig:image29}
        \end{subfigure}
        \hspace{5mm}
        \begin{subfigure}[b]{0.28\textwidth}
            \includegraphics[width=\linewidth]{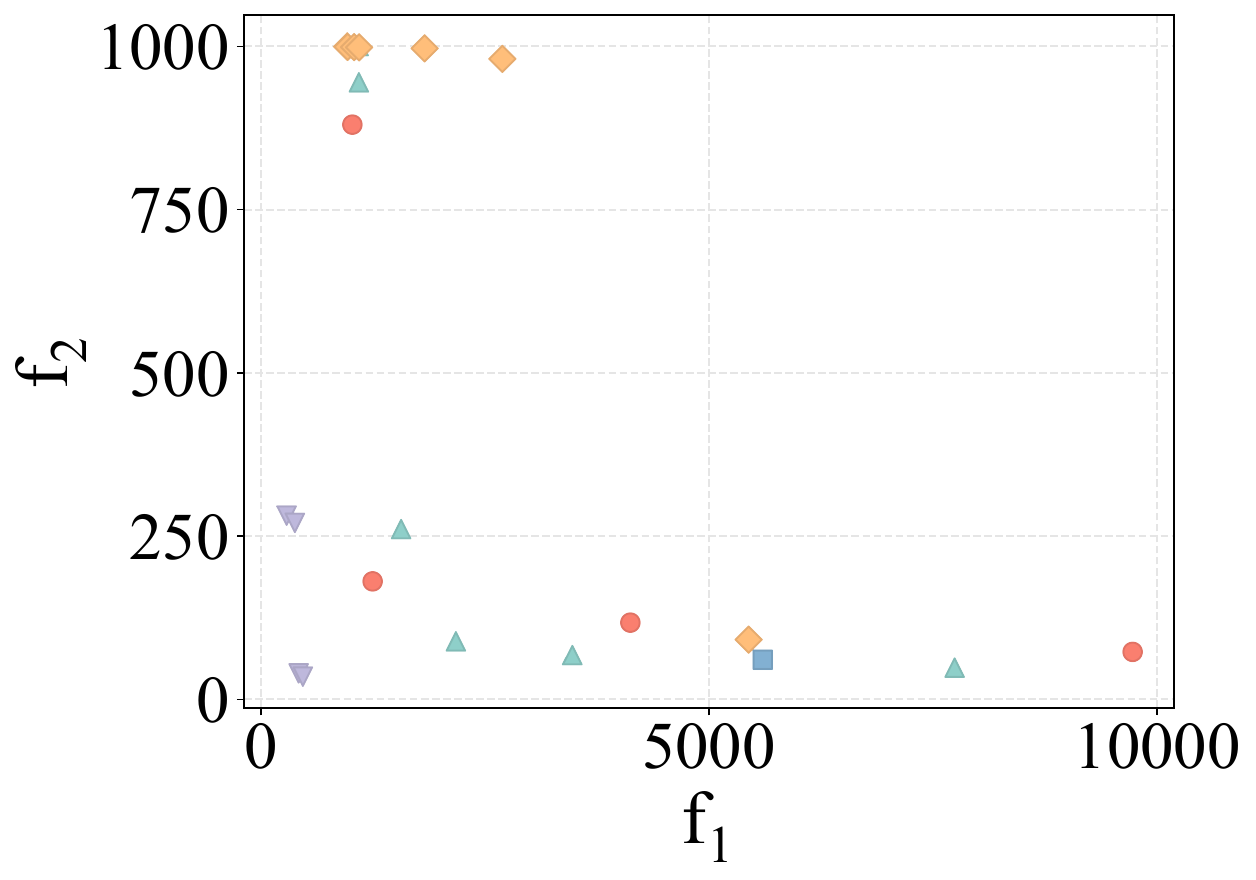}
            \captionsetup{skip=-0.5pt}
            \caption{Final results on MoWalker2d}
            \label{fig:image30}
        \end{subfigure}

        \begin{subfigure}[b]{0.9\textwidth}
            \centering
            \includegraphics[width=\textwidth]{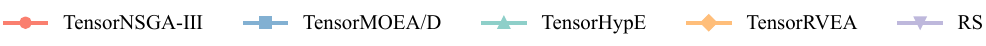}
            \captionsetup{skip=-0.5pt}
        \end{subfigure}
        \captionsetup{skip=-0.5pt}
        \caption{Comparative performance (HV, EU, and visualization of final results) of TensorNSGA-III, TensorMOEA/D, TensorHypE, TensorRVEA, and random search (RS) across varying problems: MoHalfcheetah (390D), MoHopper (243D), and MoWalker2d (390D). \textit{Note:} Higher values for all metrics indicate better performance.}

    \label{fig:neo1}
\end{figure*}

\begin{figure*}[!htb]
    \centering

        \begin{subfigure}[b]{0.28\textwidth}
            \includegraphics[width=\linewidth]{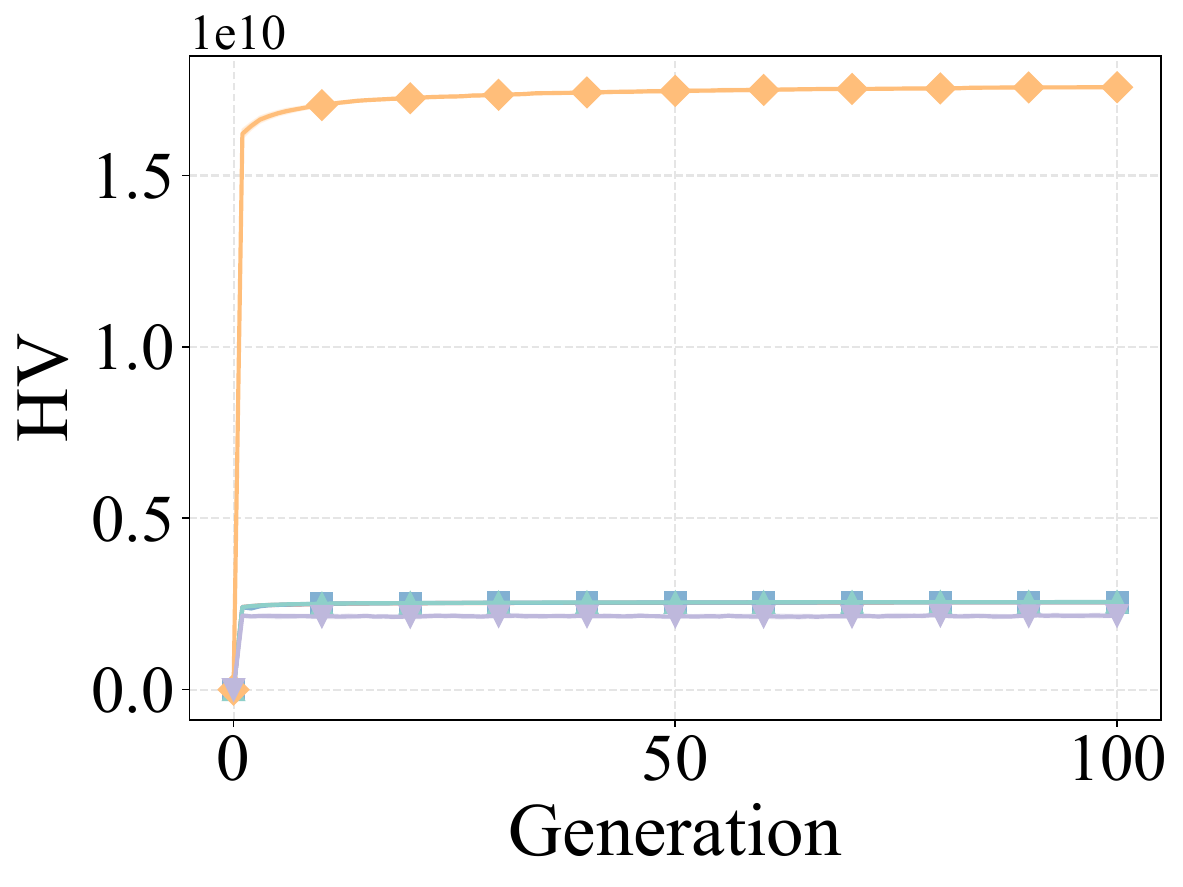}
            \captionsetup{skip=-0.5pt}
            \caption{HV on MoPusher}
            \label{fig:image16}
        \end{subfigure}
        \hspace{5mm}
        \begin{subfigure}[b]{0.28\textwidth}
            \includegraphics[width=\linewidth]{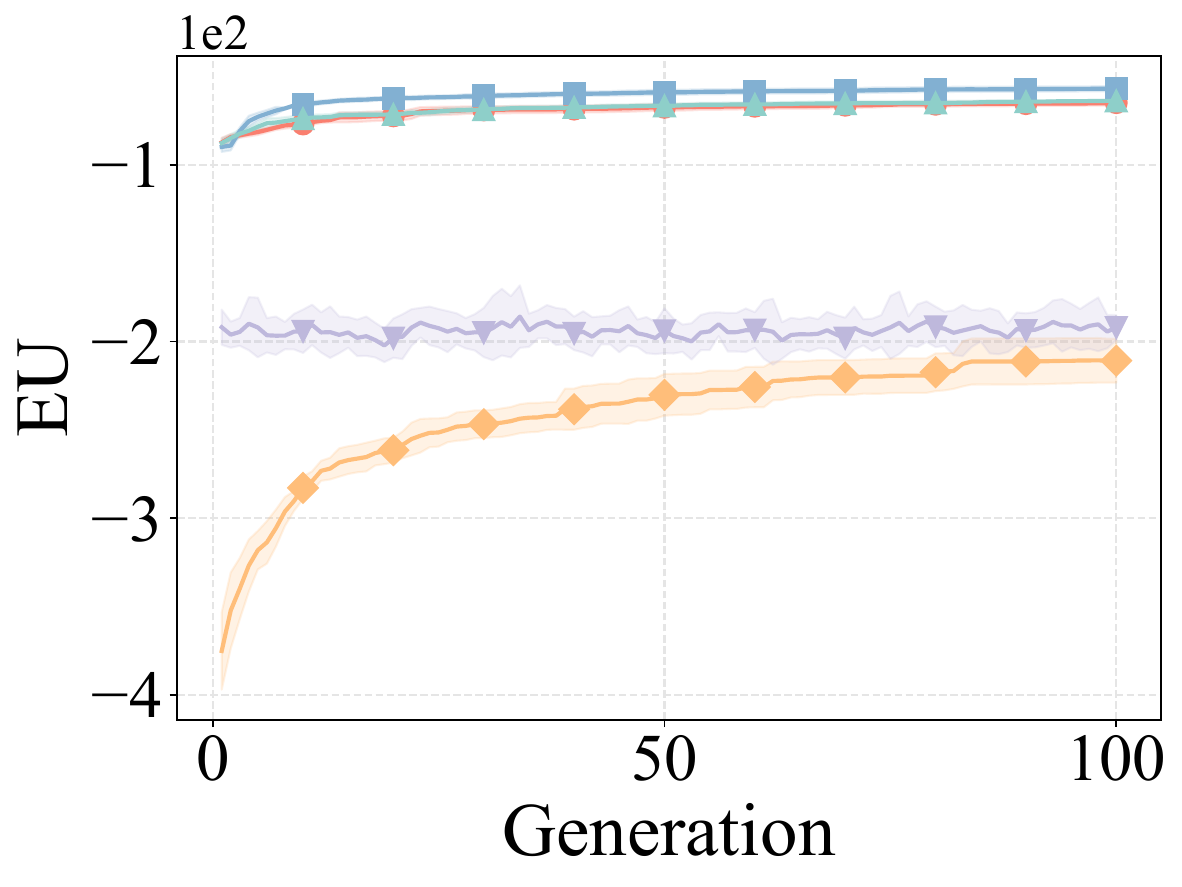}
            \captionsetup{skip=-0.5pt}
            \caption{EU on MoPusher}
            \label{fig:image17}
        \end{subfigure}
        \hspace{5mm}
        \begin{subfigure}[b]{0.28\textwidth}
            \includegraphics[width=\linewidth]{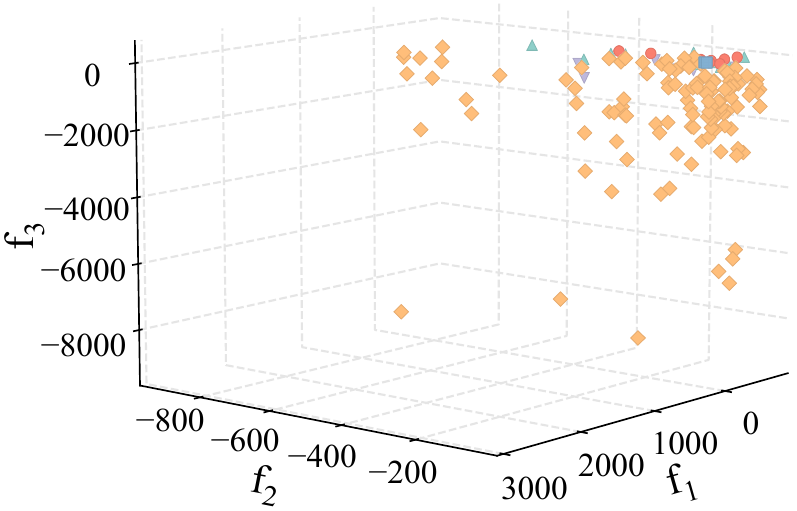}
            \captionsetup{skip=-0.5pt}
            \caption{Final results on MoPusher}
            \label{fig:image18}
        \end{subfigure}

        \begin{subfigure}[b]{0.27\textwidth}
            \includegraphics[width=\linewidth]{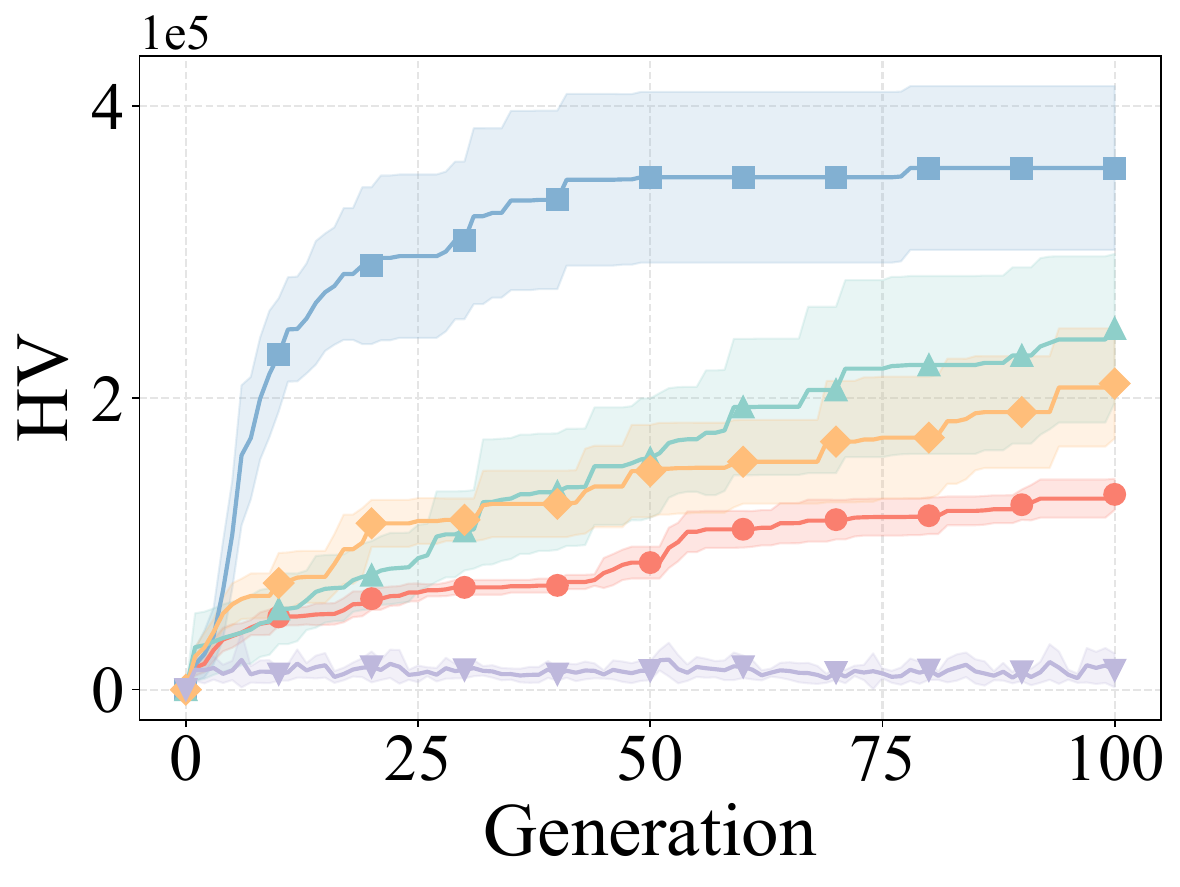}
            \captionsetup{skip=-0.5pt}
            \caption{HV on MoHumanoid}
            \label{fig:image22}
        \end{subfigure}
        \hspace{5mm}
        \begin{subfigure}[b]{0.27\textwidth}
            \includegraphics[width=\linewidth]{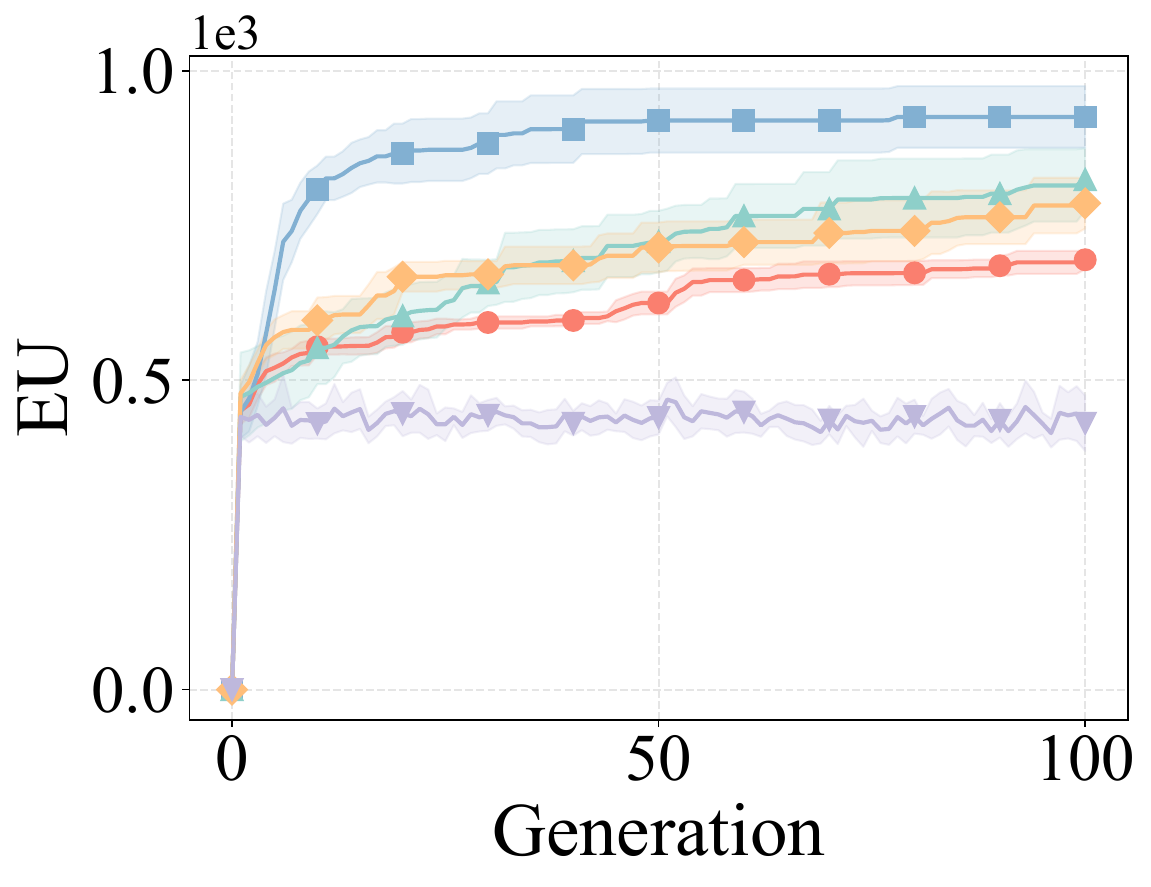}
            \captionsetup{skip=-0.5pt}
            \caption{EU on MoHumanoid}
            \label{fig:image23}
        \end{subfigure}
        \hspace{5mm}
        \begin{subfigure}[b]{0.26\textwidth}
            \includegraphics[width=\linewidth]{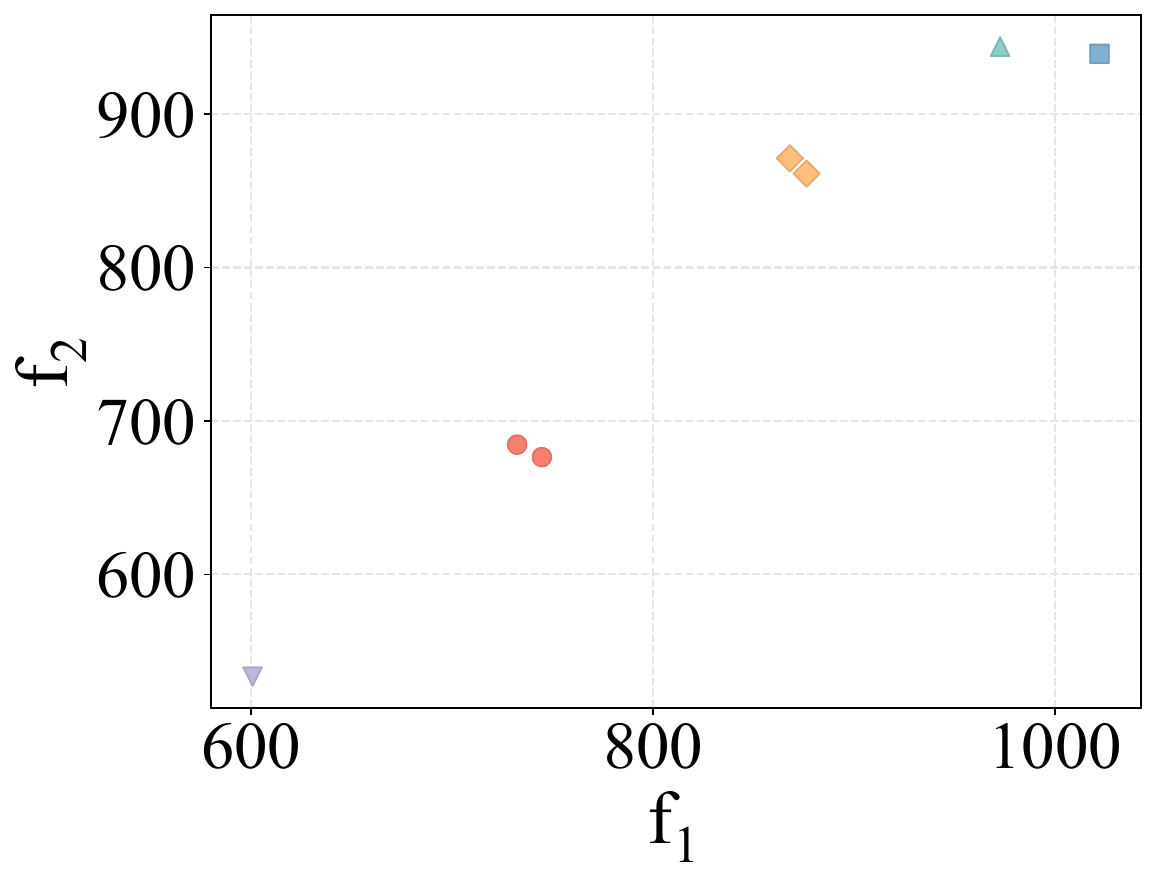}
            \captionsetup{skip=-0.5pt}
            \caption{Final results on MoHumanoid}
            \label{fig:image24}
        \end{subfigure}

        \begin{subfigure}[b]{0.28\textwidth}
            \includegraphics[width=\linewidth]{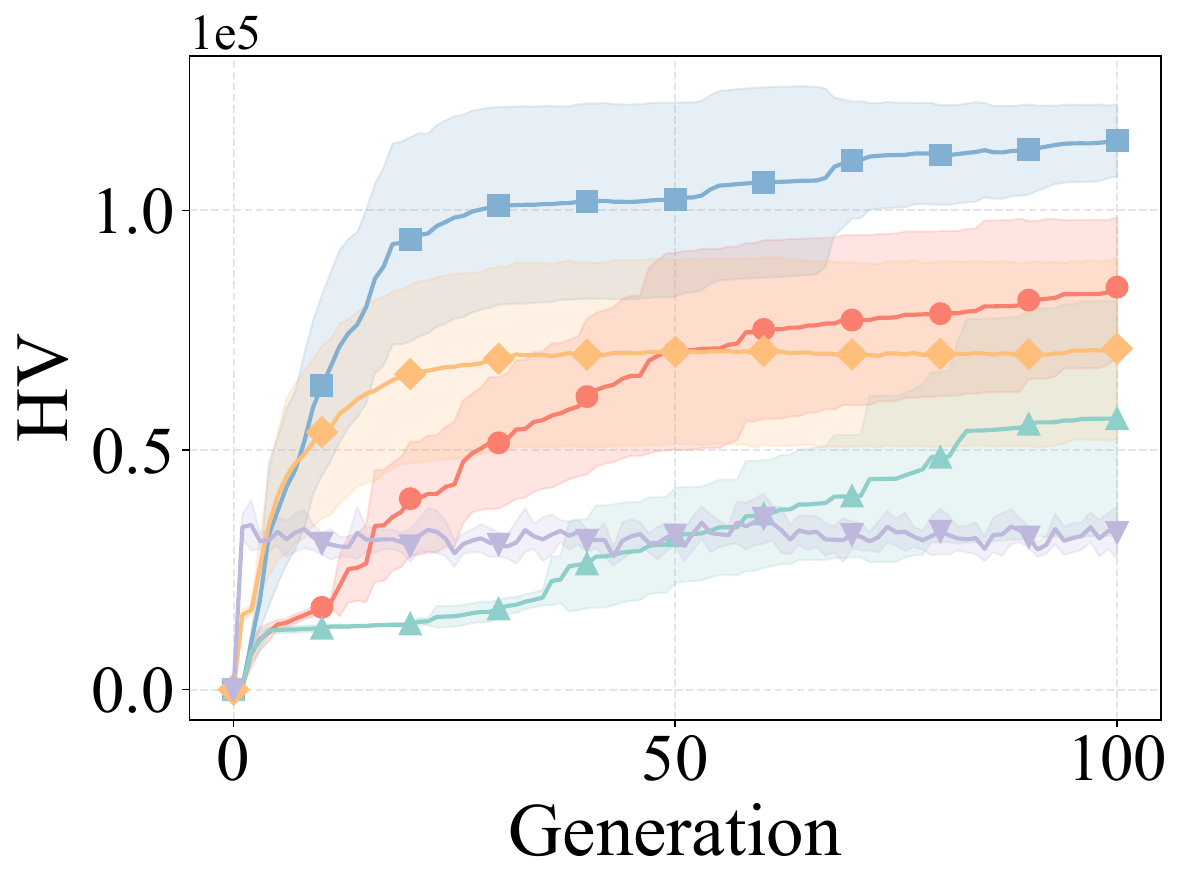}
            \captionsetup{skip=-0.5pt}
            \caption{HV on MoHumanoid-s}
            \label{fig:image25}
        \end{subfigure}
        \hspace{5mm}
        \begin{subfigure}[b]{0.28\textwidth}
            \includegraphics[width=\linewidth]{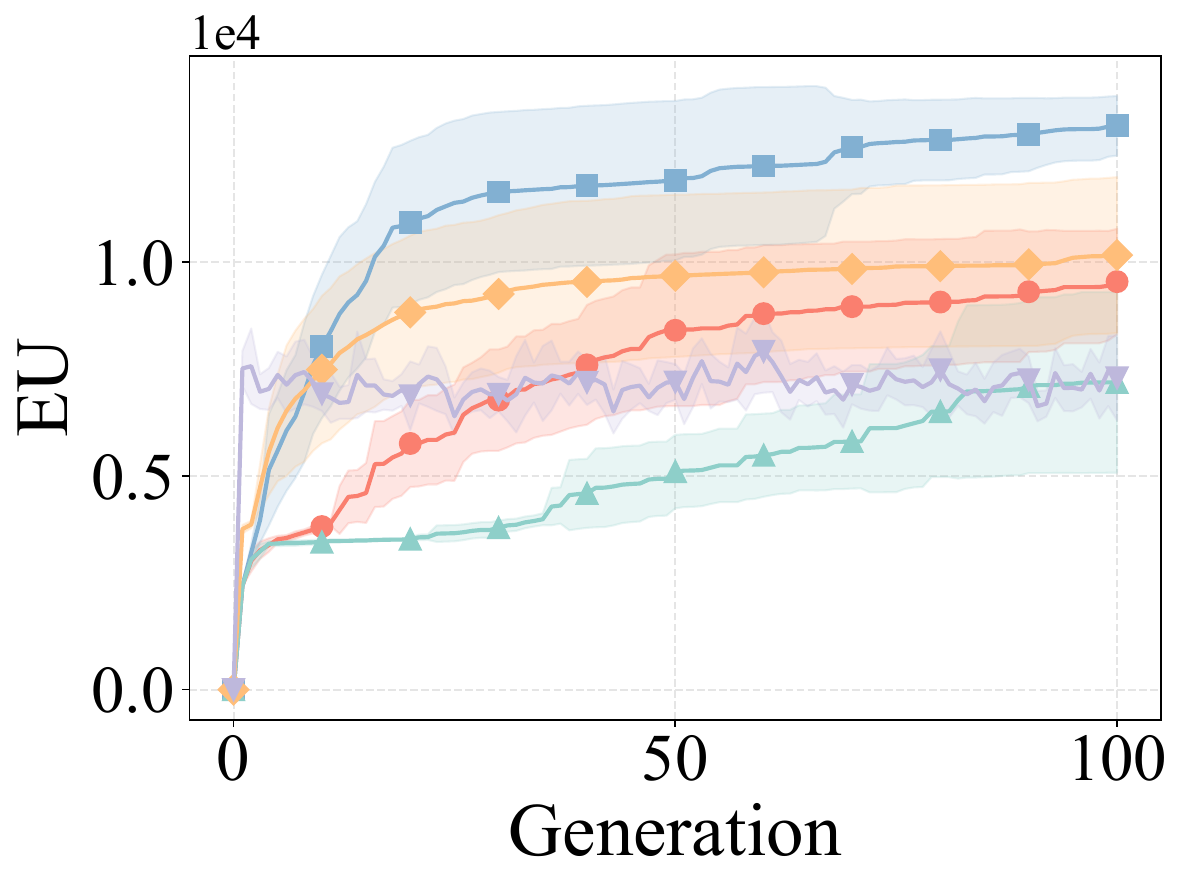}
            \captionsetup{skip=-0.5pt}
            \caption{EU on MoHumanoid-s}
            \label{fig:image26}
        \end{subfigure}
        \hspace{5mm}
        \begin{subfigure}[b]{0.265\textwidth}
            \includegraphics[width=\linewidth]{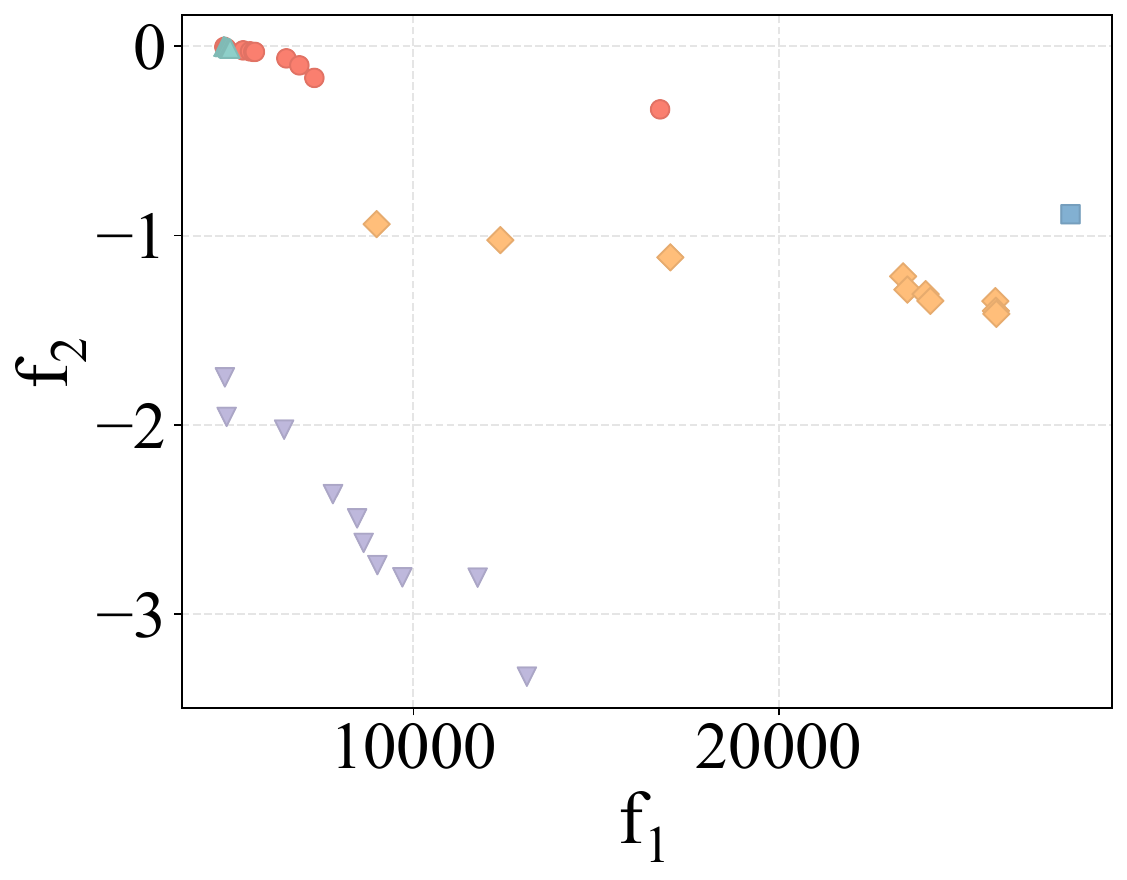}
            \captionsetup{skip=-0.5pt}
            \caption{Final results on MoHumanoid-s}
            \label{fig:image27}
        \end{subfigure}

        \begin{subfigure}[b]{0.9\textwidth}
            \centering
            \includegraphics[width=\textwidth]{figures/legend_with_markers.pdf}

        \end{subfigure}
    \captionsetup{skip=-0.5pt}
    \caption{Comparative performance (HV, EU, and visualization of final results) of TensorNSGA-III, TensorMOEA/D, TensorHypE, TensorRVEA, and random search (RS) across varying problems: MoPusher (503D), MoHumanoid (4209D), and MoHumanoid-s (4209D). \textit{Note:} Higher values for all metrics indicate better performance.}
    \label{fig:neo2}
\end{figure*}

\begin{figure*}[!htb]
    \centering

        \begin{subfigure}[b]{0.28\textwidth}
            \includegraphics[width=\linewidth]{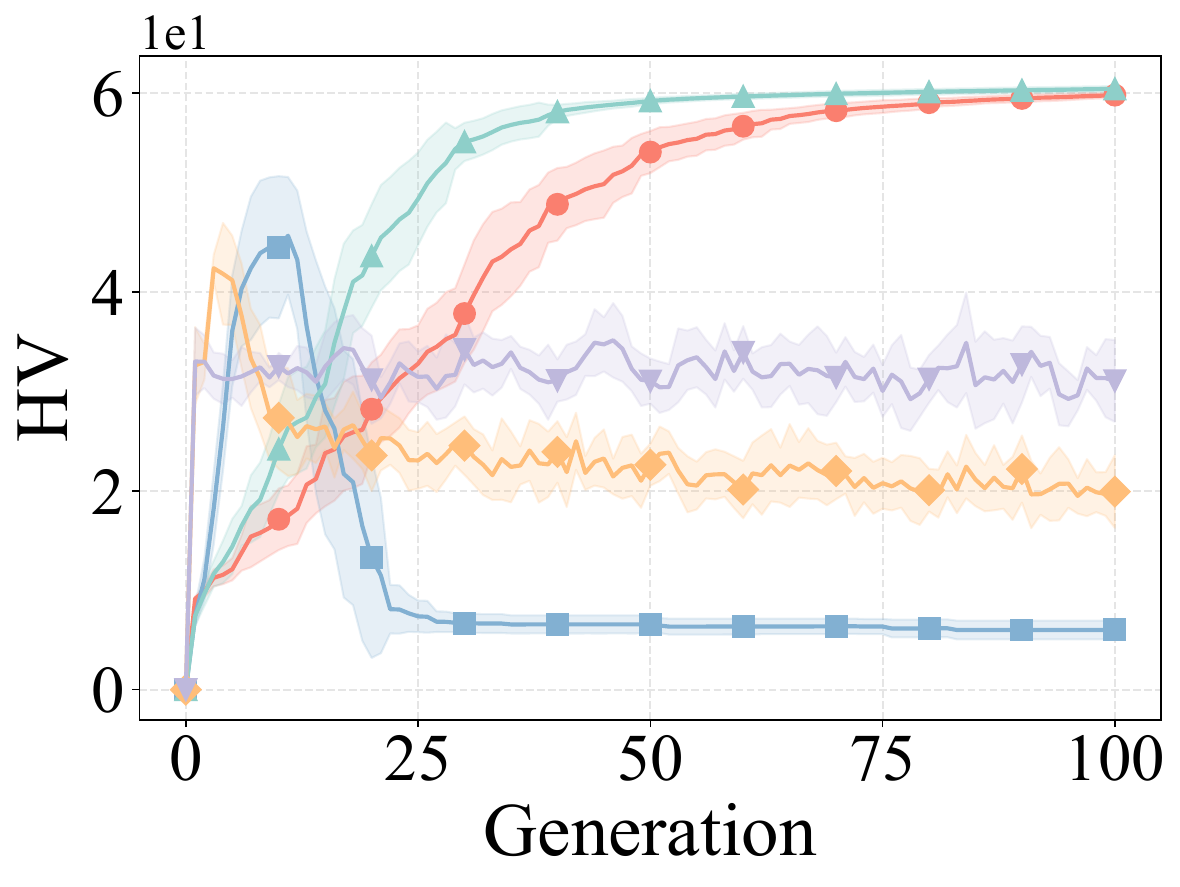}
            \captionsetup{skip=-0.5pt}
            \caption{HV on MoSwimmer}
            \label{fig:image7}
        \end{subfigure}
        \hspace{5mm}
        \begin{subfigure}[b]{0.27\textwidth}
            \includegraphics[width=\linewidth]{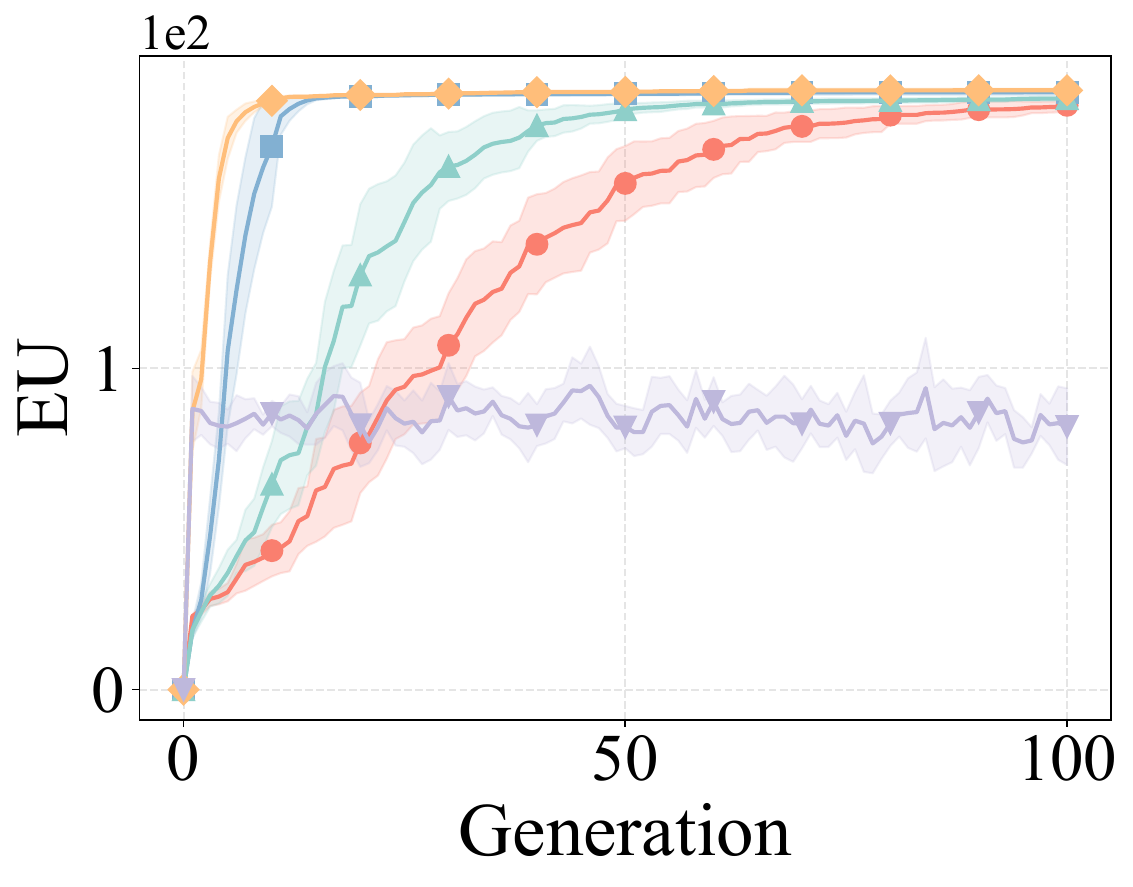}
            \captionsetup{skip=-0.5pt}
            \caption{EU on MoSwimmer}
            \label{fig:image8}
        \end{subfigure}
        \hspace{5mm}
        \begin{subfigure}[b]{0.28\textwidth}
            \includegraphics[width=\linewidth]{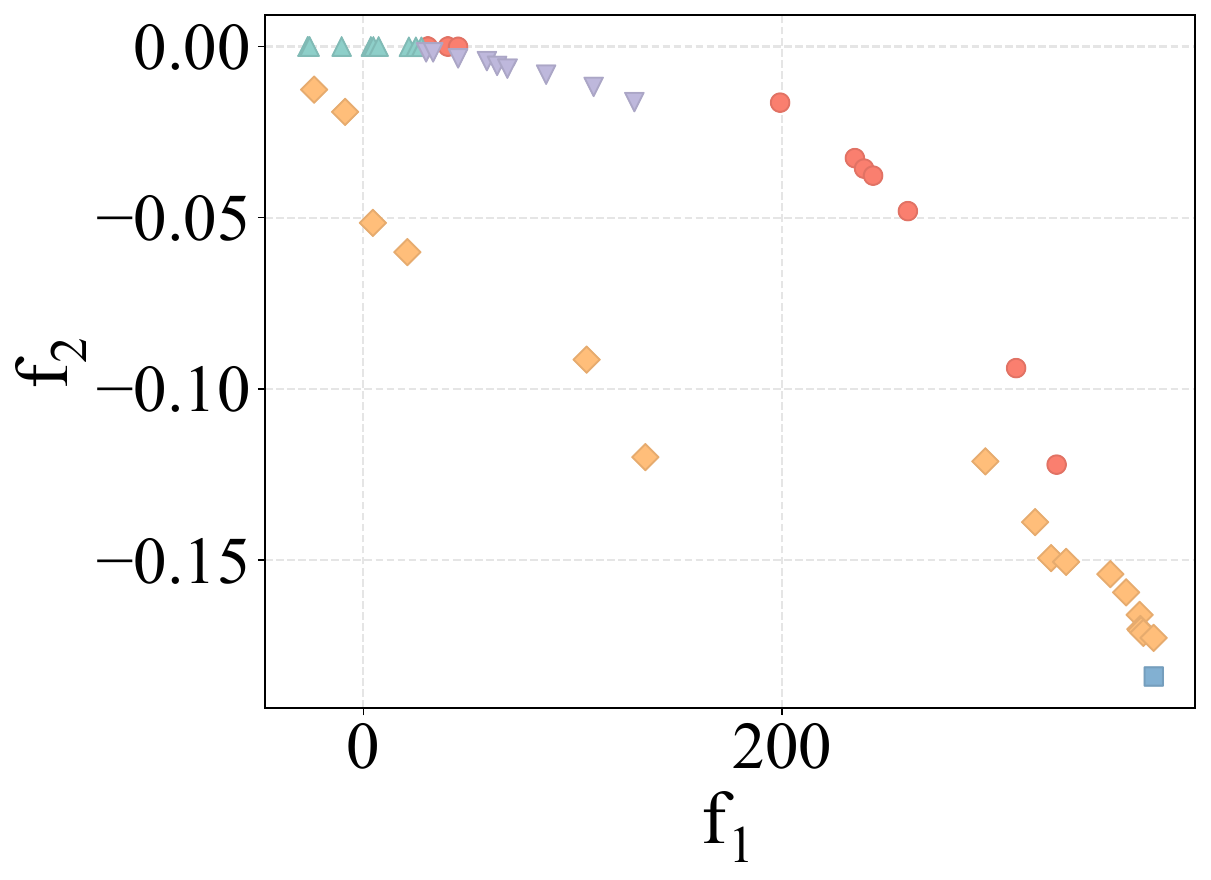}
            \captionsetup{skip=-0.5pt}
            \caption{Final results on MoSwimmer}
            \label{fig:image9}
        \end{subfigure}

        \begin{subfigure}[b]{0.28\textwidth}
            \includegraphics[width=\linewidth]{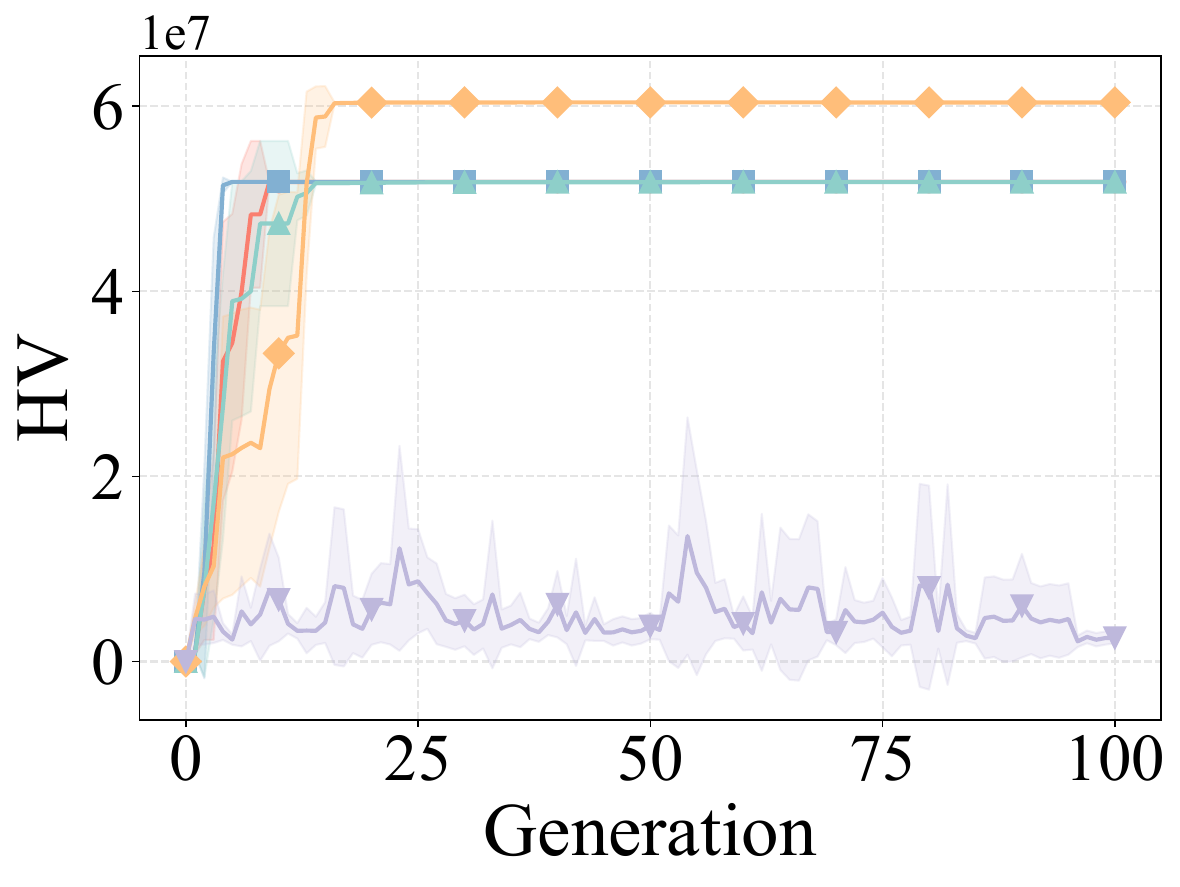}
            \captionsetup{skip=-0.5pt}
            \caption{HV on MoIDP}
            \label{fig:image10}
        \end{subfigure}
        \hspace{5mm}
        \begin{subfigure}[b]{0.27\textwidth}
            \includegraphics[width=\linewidth]{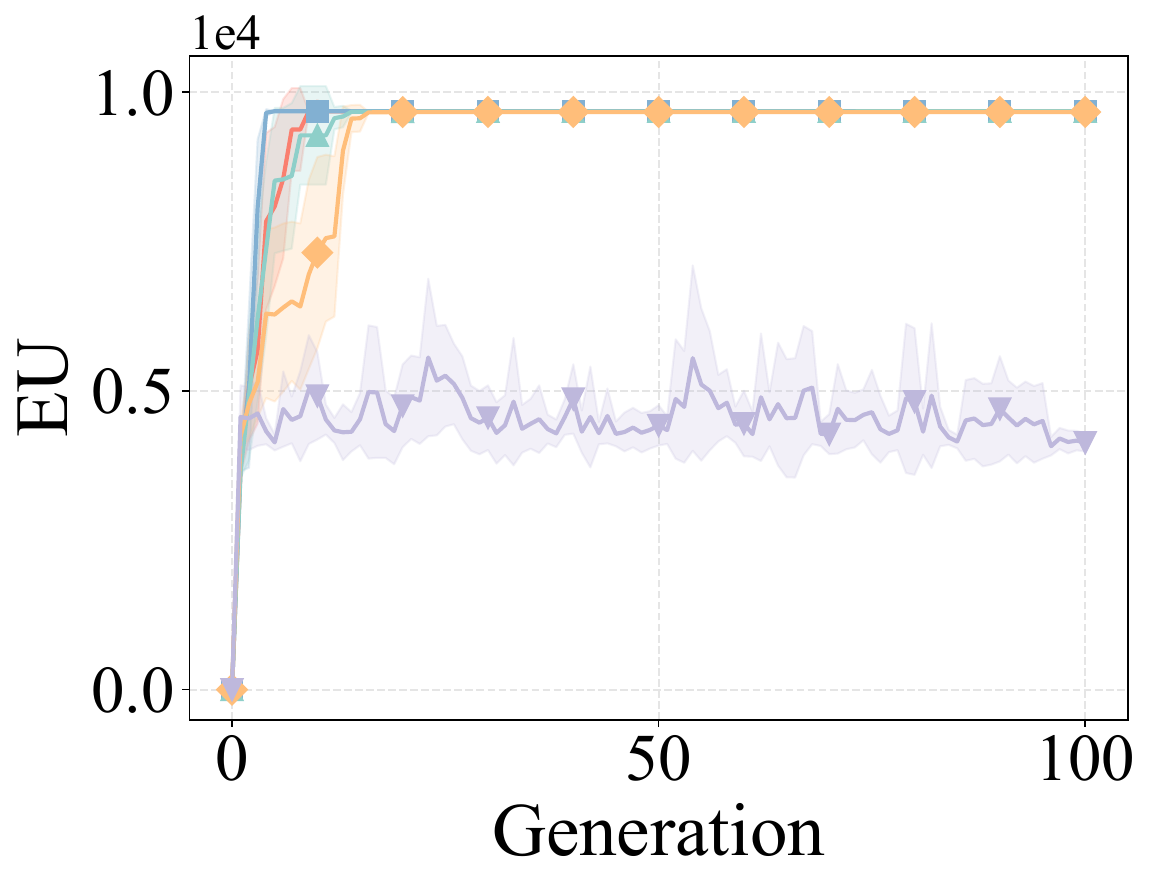}
            \captionsetup{skip=-0.5pt}
            \caption{EU on MoIDP}
            \label{fig:image11}
        \end{subfigure}
        \hspace{5mm}
        \begin{subfigure}[b]{0.28\textwidth}
            \includegraphics[width=\linewidth]{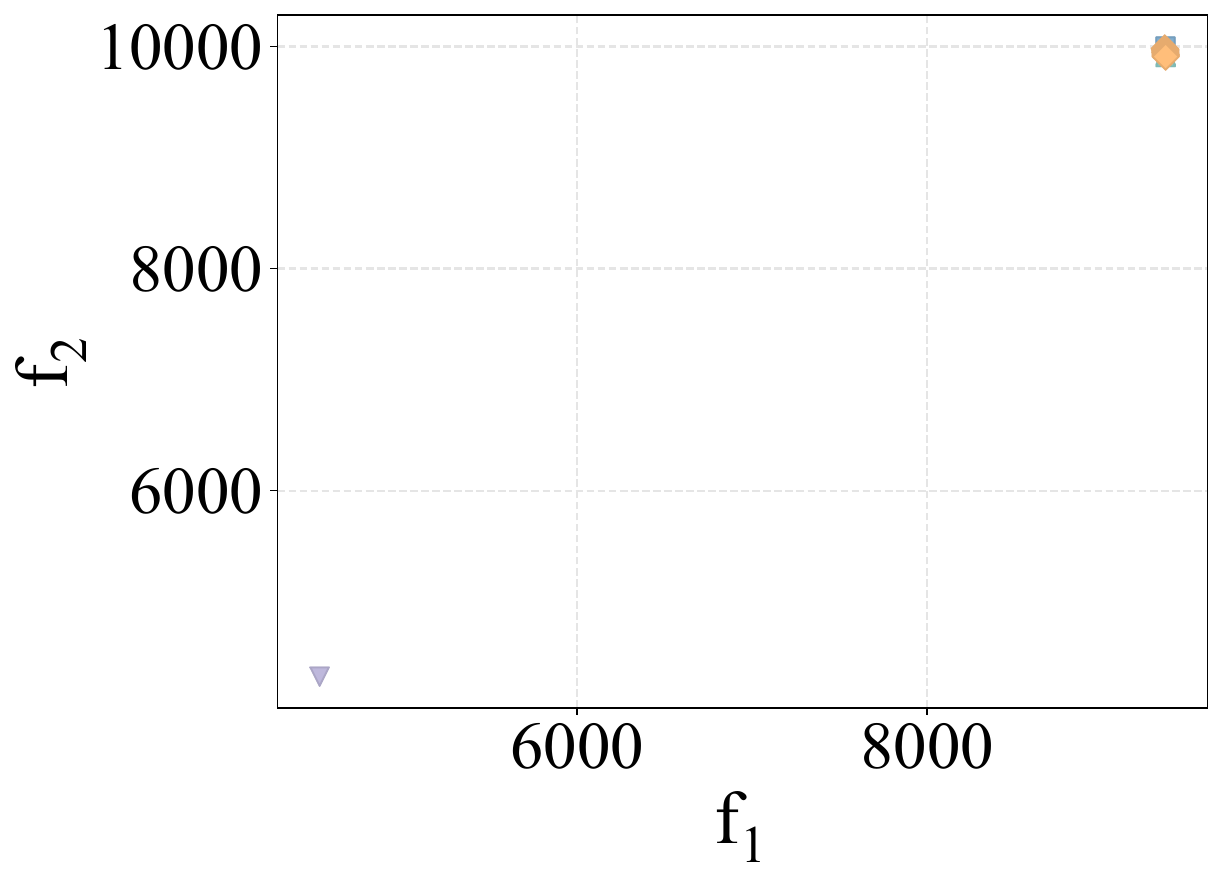}
            \captionsetup{skip=-0.5pt}
            \caption{Final results on MoIDP}
            \label{fig:image12}
        \end{subfigure}

        \begin{subfigure}[b]{0.28\textwidth}
            \includegraphics[width=\linewidth]{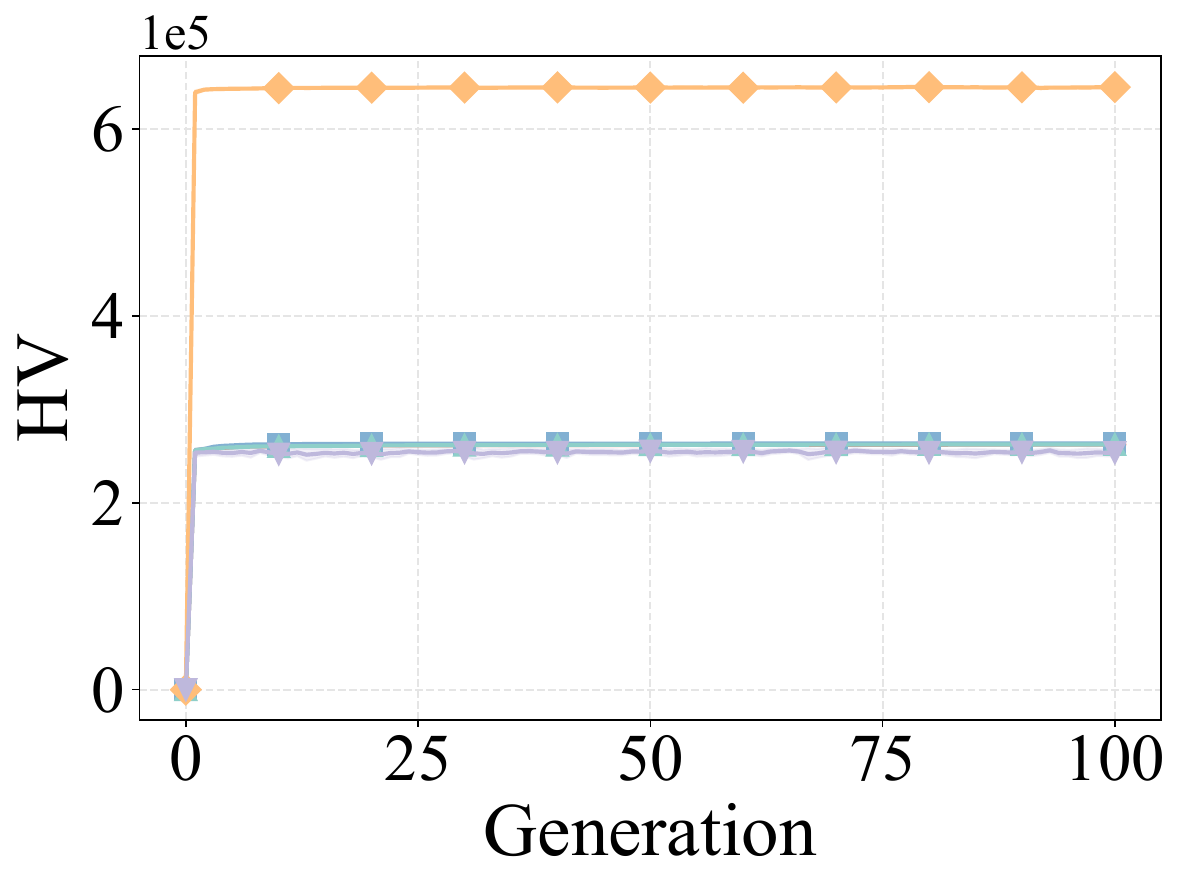}
            \captionsetup{skip=-0.5pt}
            \caption{HV on MoReacher}
            \label{fig:image19}
        \end{subfigure}
        \hspace{5mm}
        \begin{subfigure}[b]{0.28\textwidth}
            \includegraphics[width=\linewidth]{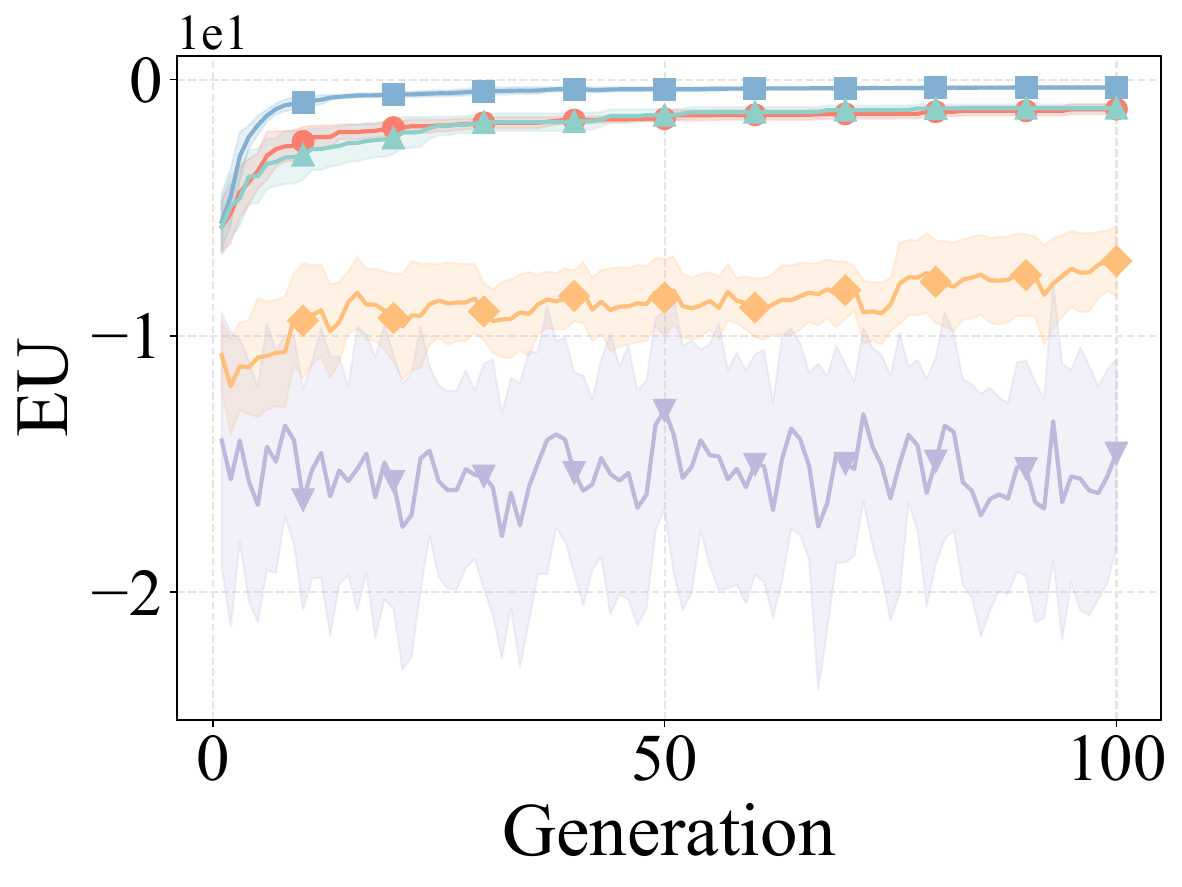}
            \captionsetup{skip=-0.5pt}
            \caption{EU on MoReacher}
            \label{fig:image20}
        \end{subfigure}
        \hspace{5mm}
        \begin{subfigure}[b]{0.27\textwidth}
            \includegraphics[width=\linewidth]{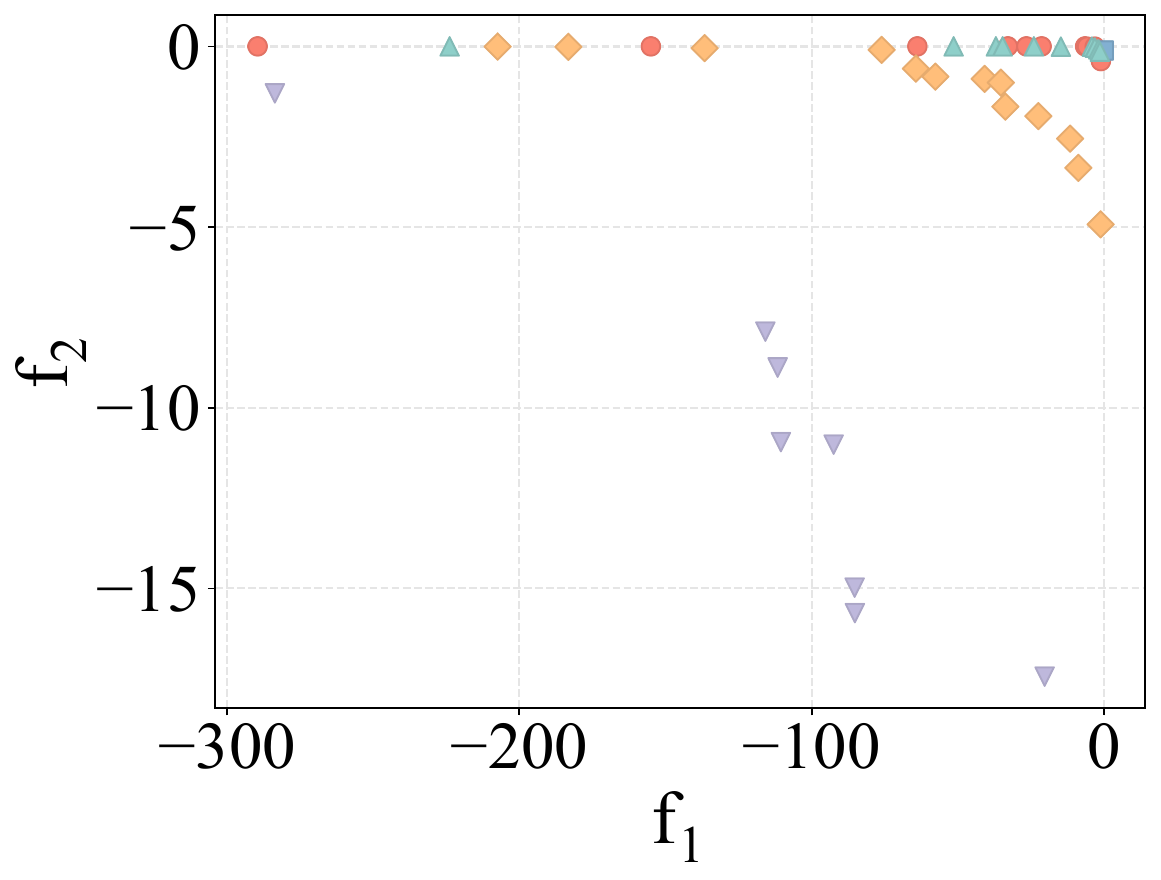}
            \captionsetup{skip=-0.5pt}
            \caption{Final results on MoReacher}
            \label{fig:image21}
        \end{subfigure} 

        \begin{subfigure}[b]{0.9\textwidth}
            \centering
            \includegraphics[width=\textwidth]{figures/legend_with_markers.pdf}
            \captionsetup{skip=-0.5pt}
        \end{subfigure}
    \captionsetup{skip=-0.5pt}
    \caption{Comparative performance (HV, EU, and visualization of final results) of TensorNSGA-III, TensorMOEA/D, TensorHypE, TensorRVEA, and random search (RS) across varying problems: MoSwimmer (178D), MoIDP (161D), and MoReacher (226D). \textit{Note:} Higher values for all metrics indicate better performance.}
    \label{fig:neo3}
\end{figure*}

\subsection{Performance in Multiobjective Robot Control Benchmark}
In this experiment, we evaluate the advantages of tensorization and the effectiveness of the three proposed tensorized algorithms in solving the multiobjective robot control tasks using the proposed MoRobtrol benchmark test suite. 
Following the paradigm of EvoRL, the EMO algorithms evolve a population of MLP neural networks, with each MLP serving as a policy model within the simulation environment of each task. 

\subsubsection{Experimental Settings}
In this experiment, we apply TensorNSGA-III, TensorMOEA/D, TensorHypE, TensorRVEA~\cite{tensorrvea}, and random search (RS) algorithms to solve 9 multiobjective robot control problems in MoRobtrol. 
Each algorithm is repeated 10 times with a population size of 10000. 
Performance is evaluated using HV~\cite{hv}, expected utility (EU)~\cite{eu}, and visualization of the final nondominated solutions. 
These indicators are calculated based on the nondominated solutions of each generation. 
Details on the TensorRVEA algorithm and the reference points for HV calculation are provided in Sections S.V and S.VII-E of the Supplementary Document, respectively.

\subsubsection{Comparison Results}
As shown in Fig.~\ref{fig:neo1}, TensorRVEA achieves the highest HV on the MoHalfcheetah, MoHopper, and MoWalker2d problems, followed closely by TensorHypE and TensorNSGA-III, which show similar performance. TensorHypE outperforms TensorNSGA-III on MoWalker2d, while TensorNSGA-III performs better on MoHalfcheetah. For EU and final results, TensorRVEA demonstrates high EU scores and strong diversity in MoHalfcheetah and MoHopper. On MoWalker2d, TensorRVEA achieves better HV, while TensorMOEA/D scores higher in EU, indicating a better preference under uniform weights. TensorNSGA-III performs best on MoWalker2d for the first objective, achieving very high speeds. Although TensorMOEA/D does not perform as well as the other algorithms, it still significantly surpasses random search.

As shown in Fig.~\ref{fig:neo2} and Fig.~\ref{fig:neo3}, TensorRVEA achieves higher HV on MoPusher and MoReacher, although its EU is lower than that of the other algorithms. On MoReacher, solutions from TensorNSGA-III, TensorMOEA/D, and TensorHypE dominate parts of TensorRVEA’s solutions, though TensorRVEA still maintains superior HV. TensorMOEA/D performs significantly better on the MoHumanoid and MoHumanoid-s problems, highlighting its effectiveness in handling large-scale problems.

Another important observation is that TensorRVEA and TensorMOEA/D can exhibit an initial rise in HV, followed by a decline before stabilizing, as seen in problems like MoSwimmer in Fig.~\ref{fig:neo3}. This phenomenon can be attributed to the weight or reference tensors guiding the population to optimize in specific directions early in the process, leading to premature convergence and a subsequent decline in HV as diversity decreases. By contrast, on the MoIDP problem, TensorRVEA achieves the highest HV, although its EU and visualization results remain comparable to those of the other algorithms.

Overall, decomposition-based algorithms such as TensorRVEA and TensorMOEA/D exhibit superior performance in large-scale multiobjective robot control tasks, particularly in handling high-dimensional decision spaces and maintaining solution diversity.

\section{Conclusion}
\label{sec:conclusion}
This article introduces a tensorization approach to address the computational limitations of traditional CPU-based EMO algorithms, enhancing both speed and scalability. 
We applied this approach across three representative EMO algorithm classes: 1) dominance-based (NSGA-III), 2) decomposition-based (MOEA/D), and 3) indicator-based (HypE), which demonstrated substantial performance improvements on GPU platforms. 
Our results confirm that tensorized algorithms can significantly accelerate computations while maintaining solution quality comparable to their original CPU-based counterparts.

To demonstrate the applicability of tensorized EMO algorithms in GPU computing environments, we also developed MoRobtrol, a comprehensive benchmark test suite that reformulates complex multiobjective robot control tasks from the physics simulation environments into MOPs. 
MoRobtrol underscores the potential of tensorized EMO algorithms to efficiently address the high computational demands of Embodied AI, illustrating their relevance to dynamic real-world applications.

While tensorization has substantially improved algorithmic efficiency, opportunities remain to further optimize speed and memory use. 
Future work will focus on refining key operators such as nondominated sorting and exploring new tensorized operators optimized for multi-GPU environments to maximize performance. 
Additionally, leveraging large population data to strengthen search strategies and integrating deep learning techniques may further extend the capabilities of EMO algorithms in tackling large-scale challenges.

% \section*{Acknowledgments}
% This should be a simple paragraph before the References to thank those individuals and institutions who have supported your work on this article.

\bibliography{reference} 
\bibliographystyle{IEEEtran}

\newpage

\twocolumn[
\begin{@twocolumnfalse}
\begin{center}
    \fontsize{24}{29}\selectfont  Bridging Evolutionary Multiobjective Optimization and GPU Acceleration via Tensorization \\
    (Supplementary Document)
    \vspace{0.5em} % Optional: Adjust vertical space here
    % \fontsize{11}{13}\selectfont Zhenyu Liang, Hao Li, Naiwei Yu, Kebin Sun, Ran Cheng \\ 
    \vspace{1em} % Optional: Adjust vertical space here
\end{center}
\end{@twocolumnfalse}
]

\setcounter{algorithm}{0}
\setcounter{table}{0}
\setcounter{figure}{0}
\setcounter{section}{0}
\setcounter{equation}{0}
\renewcommand\thealgorithm{S.\arabic{algorithm}}
\renewcommand\thetable{S.\Roman{table}}
\renewcommand{\figurename}{\normalsize Fig.}
\renewcommand\thefigure{S.\arabic{figure}}
\renewcommand\thesection{S.\Roman{section}}
\renewcommand\theequation{S.\arabic{equation}}

\vspace{1em} % Adds some vertical space between the title and the content

\setcounter{section}{0}
\section{Genetic Operators in EMO Algorithms}
\label{appendix:genetic_operators}

This section provides a comprehensive overview of the genetic operators used in EMO algorithms, including mating selection, original crossover and mutation, and the tensorization of these crossover and mutation operations.

\subsection{Mating Selection}

Mating selection is a critical component of EMO algorithms, whereby individuals are selected from the existing population to act as parents for the offspring generation. Classical methods, including roulette wheel selection, tournament selection, and random selection, are commonly utilized in this context. These methods, which leverage probability and statistical principles to varying degrees for solution selection, are well-suited for tensorization.

\subsection{Original Crossover and Mutation}

In EMO algorithms, genetic operators play a pivotal role in both the exploration and exploitation of the search space. Among the most commonly used operators in genetic algorithms tailored for real-coded variables are simulated binary crossover (SBX) and polynomial mutation. These operators are specifically engineered to maintain a balance between exploration and exploitation, proving particularly effective for problems characterized by continuous decision variables. The conventional implementation is as follows.

\subsubsection{Simulated Binary Crossover}

SBX operates under the principle of creating offspring that are close to the parent solutions, analogous to the single-point crossover used in binary-coded genetic algorithms. Given two parent solutions \( \mathbf{x}_1 = (x_1^1, x_1^2,\dots, x_1^d) \) and \( \mathbf{x}_2 = (x_2^1, x_2^2,\dots, x_2^d) \), SBX generates two offspring \( \mathbf{c}_1 \) and \( \mathbf{c}_2 \) using the following equations:
\begin{equation}
    \mathbf{c}_1^i = 0.5 \left[ (1 + \beta)\mathbf{x}_1^i + (1 - \beta)\mathbf{x}_2^i \right],
\end{equation}
\begin{equation}
    \mathbf{c}_2^i = 0.5 \left[ (1 - \beta)\mathbf{x}_1^i + (1 + \beta)\mathbf{x}_2^i \right],
\end{equation}
where \( i \) is the index of each decision variable, \( \beta \) is the spread factor, calculated based on a probability distribution and a user-defined parameter \( \eta_c \), typically known as the distribution index. The spread factor \( \beta \) is calculated as follows:
\begin{equation}
    \beta = \begin{cases}
        (2\mu)^{\frac{1}{\eta_c+1}}, & \mu \leq 0.5 \\
        \left( \frac{1}{2(1-\mu)} \right)^{\frac{1}{\eta_c+1}}, & \mu > 0.5
    \end{cases},
\end{equation}
where \( \mu \) is a random number with values between 0 and 1.

\subsubsection{Polynomial Mutation}

Polynomial Mutation is designed to introduce minor perturbations in the offspring, promoting diversity in the population. For a given parent solution \( \mathbf{x} \), the mutated solution \( \mathbf{y} \) is generated as follows:

\begin{equation}
    \mathbf{y}^i = \mathbf{x}^i + \overline{\delta} (\mathbf{u}^i - \mathbf{l}^i),
\end{equation}
where \( \mathbf{u}^i\) and \( \mathbf{l}^i \) are the upper and lower bounds of the \(i\)-the decision variable, respectively, and \( \overline{\delta} \) is the mutation step size. The value of \( \overline{\delta}\) is determined by:

\begin{equation}
    \overline{\delta} = \begin{cases}
        \left[2\mu + (1-2\mu )(1-\delta_1)^{\eta}\right]^{\frac{1}{\eta}}-1, & \mu \leq 0.5 \\
        1-\left[2-2\mu + (2\mu-1)(1-\delta_2)^{\eta}\right]^{\frac{1}{\eta}}, & \mu > 0.5
    \end{cases},
\end{equation}
where \( \eta = \eta_m + 1 \), with \( \eta_m \) being the distribution index for mutation. The term \( \mu \) represents a uniformly distributed random number within the range [0,1]. Additionally, \( \delta_1 = \frac{\mathbf{x}^i - \mathbf{l}^i}{\mathbf{u}^i - \mathbf{l}^i} \) and \( \delta_2 = \frac{\mathbf{u}^i - \mathbf{x}^i}{\mathbf{u}^i - \mathbf{l}^i} \) define the normalized differences relevant to the mutation process.

\subsection{Tensorization of Crossover and Mutation}

Inspired by the matrix-based evolutionary computation framework~\cite{mec}, which represents the entire population as a matrix to enable efficient parallel execution of evolutionary operators, we adopt a similar tensorization strategy to accelerate crossover and mutation in EMO.
The tensorized methods implemented in TensorRVEA~\cite{tensorrvea} and PlatEMO~\cite{platemo} facilitate the concurrent processing of genetic information across a population, significantly enhancing the speed and scalability of genetic operations within EMO frameworks. This demonstrates the efficacy of tensorization in contemporary computational environments. The tensorized implementation of crossover and mutation operators is as follows.

\subsubsection{Simulated Binary Crossover}

In a tensorized form, the crossover operation can be performed in parallel across the population. For instance, in the SBX, the tensorized population $\bm{X} \in \mathbb{R}^{n \times d}$ is divided into two parent tensors, $\bm{X}_1$ and $\bm{X}_2$, each with dimension $\left\lfloor \frac{n}{2} \right\rfloor \times d$. The tensorized formulations for SBX are as follows:
\begin{equation}
    \bm{X}_\text{c} = \begin{bmatrix}
        \left[ (1 + \bm{B}) \odot \bm{X}_1 + (1 - \bm{B}) \odot \bm{X}_2 \right] / 2 \\
        \left[ (1 - \bm{B}) \odot \bm{X}_1 + (1 + \bm{B}) \odot \bm{X}_2 \right] / 2
    \end{bmatrix},
\end{equation}
where $\bm{B}$ is a tensor of spread factors computed as:
\begin{equation}
    \begin{split}
        \bm{B} = & (2 \bm{M})^{\frac{1}{\eta_c+1}} \odot H(0.5 - \bm{M}) + \\
        & \left(1/(2 - 2\bm{M})\right)^{\frac{1}{\eta_c+1}} \odot (1 - H(0.5 - \bm{M}))
    \end{split},
\end{equation}
where $\eta_c$ is the distribution parameter of SBX, $\bm{M}$ is a tensor with the same dimension as $\bm{X}_1$ and $\bm{X}_2$, containing uniformly distributed random numbers in [0,1]. The $\odot$ operator denotes the Hadamard product, which performs element-wise multiplication. The $H(\cdot)$ function is the Heaviside step function, returning 1 for non-negative inputs and 0 otherwise.

\subsubsection{Polynomial Mutation}

Mutation operations introduce small random changes to solutions. In a tensorized form, mutation can be performed in parallel. For polynomial mutation:
\begin{equation}
    \bm{X}_\text{m} = \bm{X}_\text{c} + \overline{\bm{\Delta}} \odot (\bm{U} - \bm{L}),
\end{equation}
where the tensors $\bm{U}$ and $\bm{L}$ represent the upper and lower bounds of the population, respectively. $\bm{X}_\text{c}$ is the population after crossover. $\overline{\bm{\Delta}}$ is a size tensor of mutation steps, computed as:
\begin{equation}
    \overline{\bm{\Delta}}_1 = \left[ 2 \bm{M} + (1 - 2 \bm{M}) \odot (1 - \bm{\Delta}_1)^{\eta} \right]^{\frac{1}{\eta}},
\end{equation}
\begin{equation}
    \overline{\bm{\Delta}}_2 = 1 - \left[ 2 - 2 \bm{M} + (2 \bm{M} - 1) \odot (1 - \bm{\Delta}_2)^{\eta} \right]^{\frac{1}{\eta}},
\end{equation}
\begin{equation}
    \overline{\bm{\Delta}} = \overline{\bm{\Delta}}_1 \odot H(0.5 - \bm{M}) + \overline{\bm{\Delta}}_2 \odot (1 - H(0.5 - \bm{M})),
\end{equation}
where $\eta = \eta_m + 1$, and $\eta_m$ is the distribution index, $\bm{M}$ consists of uniformly distributed random numbers within [0,1], and $H$ is the Heaviside step function. Additionally, $\bm{\Delta}_1 = (\bm{X}_\text{c} - \bm{L})/(\bm{U} - \bm{L})$ and $\bm{\Delta}_2 = (\bm{U} - \bm{X}_\text{c})/(\bm{U} - \bm{L})$ are transformation tensors for normalizing crossover transformations, similar to mutation processes.

\section{Tensorization of NSGA-III}
\label{appendix:tensor_nsgaiii}

This section compares the original NSGA-III algorithm with its tensorized version, focusing on the environmental selection process. The pseudocode for both implementations is provided to demonstrate the key differences and the efficiency gains achieved through tensorization methodology.

\label{alg:original_environmental_selection_nsgaiii}

\begin{algorithm}
    \caption{Original Environmental Selection in NSGA-III}
    \label{alg:original_environmental_selection_nsgaiii}
    \begin{algorithmic}[1] % [1] enables line numbers
    \REQUIRE Combined population $P_t$ with size $2n$, population size $n$, set of reference points $\bm{Z}$.
    \ENSURE Next generation population $P_{t+1}$.
    
    \STATE Perform Nondominated Sorting on $P_t$ to identify the fronts $F_1, F_2, \dots$;
    \STATE Set $P_{t+1} = \emptyset$;
    \STATE $i \leftarrow 1$;
    
    \WHILE {$|P_{t+1}| + |F_i| \leq n$}
        \STATE $P_{t+1} \leftarrow P_{t+1} \cup F_i$;
        \STATE $i \leftarrow i + 1$;
    \ENDWHILE
    \STATE $S_t \leftarrow  P_{t+1} \cup F_i$;
    
    \IF {$|P_{t+1}| < n$}
        \STATE Normalize the objective values of solutions in $S_t$;
        \STATE Calculate the distance between $S_t$ and $\bm{Z}$;
        \STATE Associate all solutions in $S_t$ to a reference point in $\bm{Z}$ based on the minimum distance;
        \STATE Compute the niche counts for each reference point in $P_{t+1}$ and $S_t$;
        
        \WHILE {$|P_{t+1}| < n$}
            \STATE Identify the least crowded reference point $\bm{j} \in \bm{Z}$;
            \IF {There is no solution associated with $\bm{j}$ in $F_i$}
                \STATE $\bm{Z} \leftarrow \bm{Z} \setminus \{\bm{j}\}$;
            \ELSE
                \IF {There is no solution associated with $\bm{j}$ in $P_{t+1}$}
                    \STATE Select the solution $\mathbf{a} \in F_i$ with the minimum distance to $\bm{j}$;
                \ELSE
                    \STATE Randomly select a solution $\mathbf{a} \in F_i$ that is associated with $\bm{j}$;
                \ENDIF
                \STATE $P_{t+1} \leftarrow P_{t+1} \cup \{\mathbf{a}\}$;
                \STATE Update the niche count for $\bm{j}$ in both $P_{t+1}$ and $S_t$;
            \ENDIF
        \ENDWHILE
    \ENDIF
    \RETURN $P_{t+1}$
    \end{algorithmic}
\end{algorithm}

\begin{algorithm}
    \caption{Nondominated Sorting in Original NSGA-III}
    \label{alg:original_nondominated_sort}
    \begin{algorithmic}[1] % [1] enables line numbers
    \REQUIRE Population $P$ of size $n$, with each individual $\mathbf{p}_i$ having objective values $\mathbf{f}(\mathbf{p}_i)$
    \ENSURE Set of nondominated fronts $F_1, F_2, \dots$
    
    \STATE Initialize an empty list of fronts: $F = \emptyset$;
    \STATE Initialize an empty list of domination counts: $n_i = 0$ for all $\mathbf{p}_i \in P$;
    \STATE Initialize an empty list of dominated sets: $S_i = \emptyset$ for all $\mathbf{p}_i \in P$;
    
    \FOR {$\mathbf{p}_i$ in $P$}
        \FOR {$\mathbf{p}_j$ in $P$}
            \IF {$j \neq i$}
                \IF {$\mathbf{p}_i$ dominates $\mathbf{p}_j$}
                    \STATE $S_i \leftarrow S_i \cup \{\mathbf{p}_j\}$;
                \ELSIF {$\mathbf{p}_j$ dominates $\mathbf{p}_i$}
                    \STATE $n_i \leftarrow n_i + 1$;
                \ENDIF
            \ENDIF
        \ENDFOR
        \IF {$n_i = 0$}
            \STATE $F_i \leftarrow F_i \cup \{\mathbf{p}_i\}$;
        \ENDIF
    \ENDFOR
    
    \STATE $k \leftarrow 1$;
    
    \WHILE {$F_k \neq \emptyset$}
        \STATE $F_{k+1} = \emptyset$;
        \FOR {$\mathbf{p}_i$ in $F_k$}
            \FOR {$\mathbf{p}_j$ in $S_i$}
                \STATE $n_j \leftarrow n_j - 1$;
                \IF {$n_j = 0$}
                    \STATE $F_{k+1} \leftarrow F_{k+1} \cup \{\mathbf{p}_j\}$;
                \ENDIF
            \ENDFOR
        \ENDFOR
        \STATE $k \leftarrow k + 1$;
    \ENDWHILE
    \end{algorithmic}
\end{algorithm}

\subsection{Original Environmental Selection}
\label{appendix:original_nsgaiii}

In the original NSGA-III algorithm, environmental selection consists of several key steps. The process begins with nondominated sorting, which classifies the population into different Pareto fronts by comparing individuals. After sorting, the population undergoes a normalization step to scale the objective values. Following this, an association operation assigns individuals to predefined reference points. Finally, a niche-preservation method is applied to maintain diversity among the selected individuals by filling underrepresented niches. The nondominated sorting and subsequent operations, particularly the iterative loops, can become computationally expensive as the population size increases. The pseudocode for the original environmental selection process, including nondominated sorting, normalization, association, and niche-preservation, is provided in Algorithm~\ref{alg:original_environmental_selection_nsgaiii} and Algorithm~\ref{alg:original_nondominated_sort}.

\label{alg:non_dominated_sort}

\renewcommand{\algorithmicrequire}{\textbf{Input:}}
\renewcommand{\algorithmicensure}{\textbf{Output:}}
\begin{algorithm}
    \caption{Tensorized Nondominated Sorting}
    \label{alg:non_dominated_sort}
    \begin{algorithmic}[1]
        \REQUIRE Objective tensor \(\bm{F} \in \mathbb{R}^{2n \times m}\) and population size \(n\).
        \ENSURE Nondomination rank tensor \(\bm{r}\) and the last rank \(l \in \mathbb{N}\).

        \STATE \(\bm{D} \leftarrow [\bm{F}_i \prec \bm{F}_j]_{i, j}, \quad i,j = 1,2,\ldots,2n;\)

        \STATE \(\bm{c} \leftarrow \sum_{j=1}^{2n} \bm{D}_{ij}, \quad i=1,2,\dots,2n;\)

        \STATE \(\bm{r} \leftarrow \bm{0}_{2n \times 1};\)
        
        \STATE \( k \leftarrow 0;\)

        \STATE \(\bm{p} \leftarrow \mathds{1}_{\bm{c}=0};\)

        \WHILE{$\texttt{any}(\bm{p})$}

            \STATE \(\bm{r} \leftarrow H(\bm{p}) \cdot k + H(1-\bm{p}) \odot \bm{r};\)

            \STATE \(\bm{d}_j \leftarrow \sum_{i=1}^{2n} (\bm{p}_i \cdot \bm{D}_{ij}), \quad j=1,2,\dots,2n;\)

            \STATE \(\bm{c} \leftarrow \bm{c} - \bm{d} - \bm{p};\)

            \STATE \(k \leftarrow k + 1;\)
            \STATE \(\bm{p} \leftarrow \mathds{1}_{\bm{c}=0};\)

        \ENDWHILE

        \STATE \(l \leftarrow \text{sort}(\bm{r})[n];\)

    \end{algorithmic}
\end{algorithm}

\subsection{Tensorized Environmental Selection}
\label{appendix:tensorized_nsgaiii}

The tensorized version of NSGA-III leverages GPU parallelism to accelerate the entire environmental selection process, which includes nondominated sorting, normalization, association, niche count calculation, and niche selection. By transforming these operations into tensor form, batch processing can be performed across the population, significantly reducing computational complexity and execution time.

A related matrix-based method~\cite{mnd} has been proposed to accelerate nondominated sorting and population selection using simple and efficient matrix operations. This idea inspires our tensorization method, which extends such matrix-level parallelism to GPU-based tensor computations.
In the tensorized process, nondominated sorting is conducted in parallel, efficiently classifying individuals into Pareto fronts. This is followed by a normalization step to scale objective values, an association step to assign individuals to reference points, niche count calculation to track the distribution of individuals across niches, and finally, niche selection to maintain diversity. All these operations are performed in parallel, taking full advantage of the tensorization methodology. The pseudocode for the tensorized environmental selection, including all these steps, is provided in Algorithm~\ref{alg:environmental_selection_nsgaiii}.

\begin{algorithm}
    \caption{Environmental Selection in TensorNSGA-III}
    \label{alg:environmental_selection_nsgaiii}
    \begin{algorithmic}[1]
    \REQUIRE Shuffled solution tensor $\bm{X}$ and corresponding objective tensor $\bm{F}$,  reference tensors $\bm{R}$ with $n_r$ vectors, and population size $n$.
    \ENSURE Next solution tensor $\bm{X}_{\text{next}}$ and corresponding objective tensor $\bm{F}_{\text{next}}$.
    \STATE $\bm{r}, l \leftarrow \text{GPU-accelerated Nondominated Sorting}(\bm{F}, n);$ 
    \STATE $\bm{F} \leftarrow \bm{F} + \texttt{NaN} \odot H(\bm{r} - l);$
    \STATE $\bm{F}^\prime \leftarrow \text{Normalize}(\bm{F});$ \quad // refer to Algorithm~\ref{alg:normalize}
    \STATE // Association: 
    \STATE $\bm{D} \leftarrow \|\bm{F}^\prime\| \cdot \sqrt{1 - \left( \frac{\bm{F}^\prime \cdot \bm{R}^\top}{\|\bm{F}^\prime\| \cdot \|\bm{R}\|} \right)^2};$
    \STATE $\bm{\pi} \leftarrow \arg\min_j(\bm{D})$;
    \STATE $\bm{d} \leftarrow \min_j(\bm{D})$;    
    \STATE // Niche count calculation of reference tensor:
    \STATE $\bm{\rho}_j \leftarrow \sum_{i=1}^{2n} H(l - \bm{r}_i) \cdot \mathds{1}_{\bm{\pi}_i = j}, \quad j=1,\dots,n_r$;
    \STATE $\bm{\rho}_{l,j} \leftarrow \sum_{i=1}^{2n} \mathds{1}_{\bm{r}_i = l} \cdot \mathds{1}_{\bm{\pi}_i = j}, \quad j=1,\dots,n_r$;
    \STATE $n_s \leftarrow \sum \bm{\rho}$;
    \STATE // Niche selection:
    \STATE $\bm{\rho} \leftarrow \infty \cdot \mathds{1}_{\bm{\rho}_l=0};$
    \STATE $\bm{\pi} \leftarrow \bm{\pi} + \infty \cdot \mathds{1}_{\bm{r} \neq l};$
    \STATE $\bm{i} \leftarrow [1,2, \dots ,n_r];$
    \STATE $\bm{\rho}_s \leftarrow \bm{i} \odot \mathds{1}_{\bm{\rho}=0} + \infty \cdot \mathds{1}_{\bm{\rho} \neq 0};$
    \STATE $\bm{M}_{ij} \leftarrow \mathds{1}_{\bm{\rho}_{s,i}=\bm{\pi}_j}$;
    \STATE $\bm{D}^\prime \leftarrow \bm{d} \odot \bm{M} + \infty \cdot (1 - \bm{M})$;
    \STATE $\bm{q} \leftarrow \arg\min_j(\bm{D}^\prime)$;
    \STATE $\bm{q} \leftarrow \infty \cdot \mathds{1}_{\bm{\rho}_\text{selected}=\infty}  + \bm{q} \odot (1 - \mathds{1}_{\bm{\rho}_\text{selected}=\infty})$;
    \STATE $\bm{q} \leftarrow \text{Update}(\bm{q})$;
    \STATE $\bm{r}[\bm{q}] \leftarrow l-1$;
    \STATE $n_s \leftarrow n_s + \sum \mathds{1}_{\rho=0}$;
    \STATE $n_\text{dif} \leftarrow n - n_s$;
    \STATE $\bm{r} \leftarrow \text{Update-Rank}(\bm{r}, \bm{q}, n_\text{dif}, l)$; \quad // refer to Algorithm~\ref{alg:update_rank_nsgaiii}
    \STATE $\bm{i}_\text{next} \leftarrow \text{sort}(H(l-\bm{r}) \odot \bm{a} + \infty \cdot (1-H(l-\bm{r})))[:n], \bm{a}=[0,1,\dots, n-1]$;
    \STATE $\bm{X}_\text{next}, \bm{F}_\text{next} \leftarrow \bm{X}[\bm{i}_\text{next}], \bm{F}[\bm{i}_\text{next}]$;
    \end{algorithmic}
\end{algorithm}

\begin{algorithm}
    \caption{Normalization in TensorNSGA-III}
    \label{alg:normalize}
    \begin{algorithmic}[1] % [1] enables line numbers
    \REQUIRE Objective Tensor $\bm{F}$;
    \ENSURE Normalized objective tensor $\bm{F}^\prime$;
    
    \STATE $\bm{z} \leftarrow \min_i(\bm{F});$
    \STATE $\bm{F} \leftarrow \bm{F} - \bm{z};$

    \STATE \(\bm{W} \leftarrow \bm{I}_m\);
    
    \STATE \textbf{Function} \(f(\bm{w})\)
        \STATE \quad \(\bm{e} \leftarrow \text{argmin}(\max_j(\bm{F}/\bm{w}))\);

        \STATE \quad \textbf{return} \(\bm{e}\)

    \STATE $\bm{e} \leftarrow \texttt{vmap}(f)(\bm{W});$
    \STATE $\bm{E} \leftarrow \bm{F}[e];$
    
    \IF {$\text{rank}(\bm{E}) = m$}
        \STATE $\bm{a} \leftarrow \text{compute intercepts based on } \bm{E};$
    \ELSE
        \STATE $\bm{a} \leftarrow \max_i(\bm{F});$
    \ENDIF
    
    \STATE $\bm{F}^\prime \leftarrow \bm{F}/\bm{a};$
    \end{algorithmic}
\end{algorithm}

\begin{algorithm}
    \caption{Update-Rank in TensorNSGA-III}
    \label{alg:update_rank_nsgaiii}
    \begin{algorithmic}[1]
    \REQUIRE Rank tensor $\bm{r}$, selected indices $\bm{s}_\text{sel}$, difference $n_\text{dif}$, last rank $l$.
    \ENSURE Updated rank tensor $\bm{r}$.
    \STATE $\bm{s} \leftarrow \{i \mid \bm{r}_i = l\};$
    \STATE $\bm{s} \leftarrow \text{sort}(\bm{s});$
    \STATE $\bm{s} \leftarrow \bm{s} \odot H(n_\text{dif} - \bm{i}) +  \bm{s}_\text{sel}[0] \cdot H(\bm{i}-n_\text{dif} ), \bm{i}=[1,2,\dots, \text{len}(s)];$
    \STATE $\bm{\mu} \leftarrow \infty \cdot \mathds{1}_{\bm{s}_\text{sel} = \bm{s}_\text{sel}[0]} + \bm{s}_\text{sel}  \odot (1-\mathds{1}_{\bm{s}_\text{sel} = \bm{s}_\text{sel}[0]});$
    \STATE $i_\text{drop} \leftarrow \text{sort}(\bm{\mu})[0];$
    \STATE $\bm{\mu}_\text{adj} \leftarrow \bm{\mu} \odot H(-n_\text{dif} - \bm{i}) +  i_\text{drop}  \cdot H(\bm{i} + n_\text{dif} ), \bm{i}=[1,2,\dots, \text{len}(\bm{\mu})];$
    \STATE $\bm{r}_\text{add}, \bm{r}_\text{cut} \leftarrow \bm{r};$
    \STATE $\bm{r}_\text{add}[\bm{s}] \leftarrow l-1;$
    \STATE $\bm{r}_\text{cut}[\bm{\mu}_\text{adj} ] \leftarrow l;$
    \STATE $\bm{r} \leftarrow H(n_\text{dif}) \odot \bm{r}_\text{add} + H(-n_\text{dif}) \odot \bm{r}_\text{cut};$
    \end{algorithmic}
\end{algorithm}

\section{Tensorization of MOEA/D}
\label{appendix:tensormoead}

This section contrasts the original MOEA/D algorithm with its tensorized version, highlighting how tensorization methodology enhances computational efficiency. The provided pseudocode offers insights into the original and tensorized implementations of environmental selection.

\subsection{Original MOEA/D}
\label{appendix:original_moead}

\begin{algorithm}
    \caption{Original MOEA/D Algorithm}
    \label{alg:modified moead}
    \begin{algorithmic}[1]
    \REQUIRE The maximal number of generations $t_\text{max}$; $n$ weight vectors $\mathbf{\lambda}_1, \dots, \mathbf{\lambda}_n$; the number of the weight vectors in the neighborhood of each weight vector; maximum number of iterations $t_\text{max}$;
    \ENSURE EP (efficient set of solutions);
    
    \STATE EP $\gets \emptyset$;
    \FOR{each weight vector $\mathbf{\lambda}_i$}
      \STATE Compute Euclidean distances to all other weight vectors;
      \STATE Determine the $T$ closest weight vectors $B(i) = \{i_1, \dots, i_T\}$;
    \ENDFOR
    \STATE Generate initial population $\mathbf{x}_1, \dots, \mathbf{x}_n$:
    \FOR{$i = 1$ to $n$}
      \STATE $\mathbf{f}_i \leftarrow \mathbf{f}(\mathbf{x}_i)$;
    \ENDFOR
    \STATE Initialize $\mathbf{z} = (z_1, \dots, z_m)^\top$;

    \FOR{$t = 1$ to $t_\text{max}$}
        \FOR{$i = 1$ to $n$} 
            \STATE Randomly select two indices $p, q$ from $B(i)$;
            \STATE Generate a new solution $\mathbf{y}$ from $\mathbf{x}_p$ and $\mathbf{x}_q$ using genetic operators;
            \STATE Apply problem-specific improvement heuristic on $\mathbf{y}$ to produce $\mathbf{y}^\prime$;
            \FOR{$j = 1$ to $m$}
              \IF{$f_j(\mathbf{y}^\prime) < z_j$}
                \STATE $z_j \leftarrow f_j(\mathbf{y}^\prime)$;
              \ENDIF
            \ENDFOR
            \FOR{each $j \in B(i)$}
              \IF{$g(\mathbf{y}^\prime |\lambda^j, \mathbf{z}) \leq g(\mathbf{x}_j|\lambda^j, \mathbf{z})$}
                \STATE $\mathbf{x}_j \leftarrow \mathbf{y}^\prime$;
                \STATE $\mathbf{f}_j \leftarrow \mathbf{f}(\mathbf{y}^\prime)$;
              \ENDIF
            \ENDFOR
            \STATE Remove from EP all vectors dominated by $\mathbf{f}(\mathbf{y}^\prime)$;
            \IF{no vectors in EP dominate $\mathbf{f}(\mathbf{y}^\prime)$}
              \STATE Add $\mathbf{f}(\mathbf{y}^\prime)$ to EP;
            \ENDIF
        \ENDFOR
    \ENDFOR
    \end{algorithmic}
\end{algorithm}

MOEA/D decomposes a MOP into several single-objective subproblems. Each subproblem is optimized individually, and the solutions are combined to approximate the Pareto front. The original MOEA/D uses weight vectors to guide the search process, and solutions are iteratively updated based on their performance relative to these vectors.

The decomposition process in the original MOEA/D involves calculating the aggregation function for each subproblem and updating the solutions sequentially. This step is computationally intensive, especially as the number of subproblems and the population size increase. The pseudocode for the original MOEA/D is provided in Algorithm~\ref{alg:modified moead}.

\subsection{Tensorized MOEA/D}
\label{appendix:tensorized_moead}

The implementation of TensorMOEA/D retains the core principles of the original algorithm but leverages the parallel processing capabilities of GPUs. The pseudocode for the environmental selection of TensorMOEA/D is shown in Algorithm~\ref{alg:environmental_selection_moead}.

\begin{algorithm}
    \caption{Environmental Selection of TensorMOEA/D}
    \label{alg:environmental_selection_moead}
    \begin{algorithmic}[1]
        \REQUIRE Solution tensor \(\bm{X}\), Objective tensor \(\bm{F}_1\), Offspring tensor \(\bm{O}\), the objective of offspring \(\bm{F}_2\), the population size \(n\), the ideal points \(\bm{z}\), the weights \(\bm{W}\), the neighbors indices \(\bm{I}_\text{nb}\), and the PBI function \(f_{\text{PBI}}\);
        \ENSURE Next solution tensor \(\bm{X}_{\text{next}}\) and next objective tensor \(\bm{F}_{\text{next}}\);

        \STATE \(\bm{z}_{\min} \leftarrow \min_i(\bm{z} \cup \bm{F}_2)\);

        \STATE \(\bm{I}_\text{sub} \leftarrow [i \mid i \in \mathbb{N}, 0 \leq i < n]\);

        \STATE \(\bm{M} \leftarrow \mathbf{0}_{n}\);

        \STATE \textbf{Function} \(f_{\text{op1}}(\bm{i}_{\text{nb}}, \bm{f}_2)\)
        \STATE \quad \(\bm{g}_\text{old} \leftarrow f_{\text{PBI}}(\bm{F}_1[\bm{i}_\text{nb}], \bm{w}[\bm{i}_\text{nb}], \bm{z}_{\text{min}})\);
        \STATE \quad \(\bm{g}_\text{new} \leftarrow f_{\text{PBI}}(\bm{f}_2, \bm{w}[\bm{i}_\text{nb}], \bm{z}_{\text{min}})\);
        \STATE \quad \(\bm{M}[\bm{i}_\text{nb}] \leftarrow H(\bm{g}_{\text{old}} - \bm{g}_{\text{new}})\);

        \STATE \quad \(\bm{I}_\text{sub} \leftarrow \bm{M} \odot (-1) + (1 - \bm{M}) \odot \bm{I}_\text{sub}\);
        \STATE \quad \textbf{return} \(\bm{I}_\text{sub}\)

        \STATE // Comparison and Population Update:
        \STATE \(\bm{I}_\text{new} \leftarrow \texttt{vmap}(f_{\text{op1}})(\bm{I}_\text{nb}, \bm{F}_2)\);

        \STATE \textbf{Function} \(f_{\text{op2}}(\bm{i}_\text{new}, \bm{x}, \bm{f}_1, \bm{w})\)
        \STATE \quad \(\bm{f} \leftarrow \mathds{1}_{\bm{i}_\text{new}=-1} \odot \bm{F}_2 + (1-\mathds{1}_{\bm{i}_\text{new}=-1}) \odot \bm{f}_1\);
        \STATE \quad \(\bm{x} \leftarrow \mathds{1}_{\bm{i}_\text{new}=-1} \odot \bm{O} + (1-\mathds{1}_{\bm{i}_\text{new}=-1}) \odot \bm{x}\);
        \STATE \quad \(i \leftarrow \text{argmin}(f_{\text{PBI}}(\bm{f}, \bm{w}, \bm{z}_{\text{min}}))\);
        \STATE \quad \textbf{return} \(\bm{x}[i], \bm{f}[i]\)

        \STATE // Elite Selection:
        \STATE \(\bm{X}_\text{next}, \bm{F}_\text{next} \leftarrow \texttt{vmap}(f_{\text{op2}})(\bm{I}_\text{new}^\top, \bm{X}, \bm{F}_1, \bm{w})\);

        \STATE // Update the ideal point: 
        \STATE \(\bm{z} \leftarrow \bm{z}_\text{min}\);
    \end{algorithmic}
\end{algorithm}

\section{Tensorization of HypE}
\label{appendix:tensorized_hype}

In this section, we discuss the original Monte Carlo HV estimation method employed in HypE and present its tensorized adaptation, which enhances computational efficiency by utilizing GPU parallelization. The tensorized version, TensorHypE, is specifically designed to handle larger populations and more complex optimization tasks by performing HV contribution calculations in parallel, significantly speeding up the environmental selection process.

\subsection{Original Environmental Selection}
\label{appendix:original_hype}

\begin{algorithm}
    \caption{Original Monte Carlo HV Estimation}
    \label{alg:original_MCHV}
    \begin{algorithmic}[1]
    \REQUIRE Population $\mathbf{P}$, reference set $\mathbf{R}$, fitness parameter $k \in \mathbb{N}$, number of sampling points $M \in \mathbb{N}$
    \ENSURE Estimated fitness values $\mathbf{F}$
    \STATE // determine sampling box $\mathbf{S}$
    \FOR{$i \leftarrow 1$ to $m$}
        \STATE $l_i \leftarrow \min_{\mathbf{a} \in \mathbf{P}} f_i(\mathbf{a})$;
        \STATE $u_i \leftarrow \max_{(r_1, \dots, r_m) \in \mathbf{R}} r_i$;
    \ENDFOR
    \STATE $\mathbf{S} \leftarrow [l_1, u_1] \times \dots \times [l_m, u_m]$;
    \STATE $\mathbf{V} \leftarrow \prod_{i=1}^{m} \max\{0, (u_i - l_i)\}$;
    \STATE // reset fitness assignment
    \STATE $\mathbf{F} \leftarrow \bigcup_{\mathbf{a} \in \mathbf{P}} \{(\mathbf{a}, 0)\}$;
    \STATE // sampling
    \FOR{$j \leftarrow 1$ to $M$}
        \STATE choose $\mathbf{s} \in \mathbf{S}$ uniformly at random;
        \IF{$\exists \mathbf{r} \in \mathbf{R}: \mathbf{s} \leq \mathbf{r}$}
            \STATE $\texttt{UP} \leftarrow \bigcup_{\mathbf{a} \in \mathbf{P}, \mathbf{f}(\mathbf{a}) \leq \mathbf{s}} \{\mathbf{f}(\mathbf{a})\}$;
            \IF{$|\texttt{UP}| \leq k$}
                \STATE // hit in a relevant partition
                \STATE $\mathbf{\alpha} \leftarrow \prod_{l=1}^{|\texttt{UP}|-1} \frac{k-l}{|\mathbf{P}|-l}$;
                \STATE // update HV estimates
                \STATE $\mathbf{F}^\prime \leftarrow \emptyset$;
                \FOR{all $(\mathbf{a}, \mathbf{v}) \in \mathbf{F}$}
                    \IF{$\mathbf{f}(\mathbf{a}) \leq \mathbf{s}$}
                        \STATE $\mathbf{F}^\prime \leftarrow \mathbf{F}^\prime \cup \{(\mathbf{a}, \mathbf{v} + \frac{\mathbf{\alpha}}{|\texttt{UP}|} \cdot \frac{\mathbf{V}}{M})\}$;
                    \ELSE
                        \STATE $\mathbf{F}^\prime \leftarrow \mathbf{F}^\prime \cup \{(\mathbf{a}, \mathbf{v})\}$;
                    \ENDIF
                \ENDFOR
                \STATE $\mathbf{F} \leftarrow \mathbf{F}^\prime$;
            \ENDIF
        \ENDIF
    \ENDFOR
    \end{algorithmic}
\end{algorithm}

The original HypE uses Monte Carlo sampling to estimate the HV contributions of solutions in a population. The algorithm ranks solutions based on their contributions to the overall HV, favoring those that contribute more to the Pareto front's volume. The original implementation of HypE involves sequentially sampling points within a defined HV region and updating fitness values iteratively.

Algorithm~\ref{alg:original_MCHV} illustrates the pseudocode for the original Monte Carlo HV estimation method in HypE. This method can be computationally expensive when applied to large populations and high-dimensional objective spaces, as it relies heavily on sequential processing.

\subsection{Tensorized Environmental Selection}
\label{appendix:tensorhype}

While preserving the core principles of HypE, crucial steps such as nondominated sorting, HV estimation, and selection are accelerated using GPUs. Algorithm~\ref{alg:environmental_selection_hype} illustrates the tensorized environmental selection process, and Algorithm~\ref{alg:hypervolume_estimation} provides details on the GPU-accelerated HV estimation via Monte Carlo sampling.

\begin{algorithm}
    \caption{Environmental Selection of TensorHypE}
    \label{alg:environmental_selection_hype}
    \begin{algorithmic}[1]
        \REQUIRE Merged solution tensor \(\bm{X}\), merged objective tensor \(\bm{F}\), reference point \(\bm{v}_\text{ref}\), the population size \(n\), and the number of sample points \(s\);
        \ENSURE Next solution tensor \(\bm{X}_{\text{next}}\) and next objective tensor \(\bm{F}_{\text{next}}\);
        
        \STATE $\bm{r}, l \leftarrow \text{GPU-based Nondominated Sorting}(\bm{F})$; 
        \STATE $\bm{r}_{\text{mask}} \leftarrow H(l - \bm{r})$;
        \STATE $ k \leftarrow \sum_{i=1}^{\text{len}(\bm{r}_{\text{mask}} )} \bm{r}_{\text{mask},i} - n$;
        \STATE $\bm{v}_\text{hv} \leftarrow \text{Hypervolume Estimation}(\bm{F}, \bm{v}_\text{ref}, k, s)$;
        \STATE $\bm{d} \leftarrow \bm{r}_\text{mask}  \odot \bm{v}_\text{hv} + (1 - \bm{r}_\text{mask} ) \odot (-\infty)$;

        \STATE $\bm{I} \leftarrow \text{lexsort}(\bm{r}, -\bm{d})[:n]$;

        \STATE $\bm{X}_{\text{next}} \leftarrow \bm{X}[\bm{I}]$;
        \STATE $\bm{F}_{\text{next}} \leftarrow \bm{F}[\bm{I}]$;
    \end{algorithmic}
\end{algorithm}

\begin{algorithm}
    \caption{HV Estimation via Monte Carlo Sampling}
    \label{alg:hypervolume_estimation}
    \begin{algorithmic}[1]
        \REQUIRE Objective tensor \(\bm{F}\), reference point \(\bm{v}_\text{ref}\), the parameter \(k\) and the number of sample points \(s\);
        \ENSURE HV values \(\bm{v}_\text{hv}\);
        
        \STATE $n_1, m \leftarrow \text{shape}(\bm{F})$;
        \STATE $\bm{l} \leftarrow [i \mid i \in \mathbb{N}, 1 \leq i < n_1]$;
        \STATE $\bm{\lambda} \leftarrow [1, (k - \bm{l})/(n_1 - \bm{l})]$;
        \STATE $\bm{\alpha}_j \leftarrow \prod_{i=1}^{j} \bm{\lambda}_i/j, j = 1, 2, \dots, k$;

        \STATE $\bm{f}_l \leftarrow \min_i(\bm{F})$;
        \STATE $\bm{f}_u \leftarrow \bm{v}_\text{ref}$;

        \STATE $\bm{S} \gets$ Uniform sampling from $[\bm{f}_l, \bm{f}_u]$ in dimension $s \times m$;

        \STATE $\bm{v}_\text{ds} \leftarrow \bm{0}_{1 \times s}$;

        \STATE \textbf{Function} $f_{\text{pds}}(\bm{f})$
        % \STATE \quad $\bm{T}_\text{pds}[p,:] \leftarrow \mathbb{I} \left(\sum_{j=1}^{m} H(\bm{S}_{ij} - \bm{f}) = m \right)$, where $i=1, 2, \dots, s$;
        \STATE \quad $\bm{t}_\text{pds} \leftarrow \mathds{1}_{\sum_{j=1}^{m} H(\bm{S}_{ij} - \bm{f}) = m}$, where $i=1, \dots, s$;
        \STATE \quad \textbf{return} \(\bm{t}_\text{pds}\);

        \STATE \textbf{Function} $f_{\text{hv}}(\bm{t}_\text{pds})$
        \STATE \quad $\bm{v}_\text{temp} \leftarrow \bm{t}_\text{pds} \odot \bm{v}_\text{ds} - (1-\bm{t}_\text{pds})$;
        \STATE \quad $\bm{v}_\text{hv,i} \leftarrow \sum_{i=1}^{s} \left(\bm{\alpha}[\bm{v}_\text{temp}] \odot \mathds{1}_{\bm{v}_\text{temp} \neq -1}\right)_i$;
        \STATE \quad \textbf{return} \(\bm{v}_\text{hv,i}\);

        \STATE // Compute point dominance scores:
        \STATE $\bm{T}_\text{pds} \leftarrow \texttt{vmap}(f_{\text{pds}})(\bm{F})$;

        \STATE $\bm{T}_\text{temp} \leftarrow \bm{T}_\text{pds} \odot (\mathds{1}_{n \times 1} \cdot \bm{v}_\text{ds} + 1) + (1-\bm{T}_\text{pds}) \odot (\mathds{1}_{n \times 1} \cdot \bm{v}_\text{ds})$;

        \STATE $\bm{v}_\text{ds} \leftarrow \text{maximum}(\sum_{i=1}^{n_1} \bm{T}_{\text{temp},i} - 1, 0)$;

        \STATE // Compute HV values:
        \STATE $\bm{v}_\text{hv} \leftarrow \texttt{vmap}(f_{\text{hv}})(\bm{T}_\text{pds})$;

        \STATE $\bm{v}_\text{hv} \leftarrow \bm{v}_\text{hv} \cdot \prod_{i=1}^{m} (\bm{v}_{\text{ref},i} - \bm{f}_{l,i}) / s$;
    \end{algorithmic}
\end{algorithm}

\section{Introduction to TensorRVEA}
\label{appendix:tensorrvea}

TensorRVEA~\cite{tensorrvea} is a tensorized extension of the original RVEA~\cite{rvea} designed to leverage the computational advantages of modern hardware, such as GPUs. By representing key data structures and operations in tensor form, TensorRVEA efficiently processes large populations and high-dimensional objectives, making it particularly suitable for solving complex MOPs.

The key contribution in TensorRVEA lies in its ability to parallelize the selection process using tensor operations. The selection operation, a crucial component of the algorithm, is responsible for maintaining a diverse set of solutions that are well-distributed along the Pareto front. This operation uses angular distances between objective tensors and reference tensors to guide the search towards unexplored regions of the objective space.

\begin{algorithm}
    \caption{Environmental Selection of TensorRVEA}
    \label{alg:tensorized_rvea_selection}
    \begin{algorithmic}[1]
        \REQUIRE Solution tensor $\bm{X}$ with $n$ individual, objective tensor $\bm{F}$, reference tensors $\bm{V}$ with $r$ vectors, maximum number of generations $t_{\text{max}}$, current generation $t$, and rate of change of penalty $\alpha$;
        \ENSURE Elite solution tensor $\bm{X}_{\text{elite}}$;

        \STATE \(\bm{z}_{\text{min}} \leftarrow \min_i(\bm{F})\);
        \STATE \(\bm{F}^{\prime} \leftarrow \bm{F} - \bm{z}_{\text{min}}\);
        \STATE \( \bm{\Theta} \leftarrow \arccos\left(\bm{F}^{\prime} \cdot \bm{V}^{\top}/(\lVert\bm{F}^{\prime} \rVert \cdot \lVert\bm{V}^{\top}\rVert)\right);\)

        \STATE \(\bm{A} \leftarrow \text{repeat}(\text{argmin}_j{(\bm{\Theta})}, r)\);

        \STATE \(\bm{T}_{\text{part}} \leftarrow \text{repeat}(\begin{bmatrix} 0, 1, \ldots, n-1\end{bmatrix}^{\top}, r)\);
        \STATE \(\bm{I} \leftarrow \text{repeat}(\begin{bmatrix} 0, 1, \ldots, r-1\end{bmatrix}, n)\);

        \STATE \(\bm{T}_{\text{part}} \leftarrow (1 - |\text{sgn}(\bm{A} - \bm{I})|) \odot \bm{T}_{\text{part}} - |\text{sgn}(\bm{A} - \bm{I})|\);
		\STATE \(\bm{\Gamma} \leftarrow \text{argmin}_j{\Big(\arccos\left(\bm{V} \cdot \bm{V}^{\top}/(\lVert\bm{V}\rVert \cdot \lVert\bm{V}^{\top}\rVert)\right)\Big)}\);

        \STATE \textbf{Function} $f_{\text{APD}}(\bm{t}_{\text{part}}, \bm{\gamma}, \bm{\theta})$
        \STATE \quad \(\bm{t}_{\text{APD}} = \Big(1 + m \cdot \left(\frac{t}{t_{\text{max}}}\right)^\alpha \cdot \frac{\bm{\theta}[\bm{t}_{\text{part}}]}{\bm{\gamma}}\Big) \odot \lVert \bm{F}^{\prime}[\bm{t}_{\text{part}}]\rVert\);
        \STATE \quad \textbf{return} \(\bm{t}_{\text{APD}}\);

        \STATE \(\bm{T}_{\text{APD}} \leftarrow \texttt{vmap}(f_\text{APD})(\bm{T}_{\text{part}}, \bm{\Gamma}, \bm{\Theta})\);

        \STATE Replace elements in $\bm{T}_{\text{APD}}$ with $\texttt{inf}$ where $\bm{T}_{\text{part}} = -1$;
        \STATE $\bm{I}_{\text{next}} \leftarrow \text{argmin}_i(\bm{T}_{\text{APD}})$;

        \STATE \(\bm{X}_{\text{elite}} \leftarrow \bm{X}[\bm{I}_{\text{next}}]\);
    \end{algorithmic}
\end{algorithm}

The TensorRVEA environmental selection operation illustrated in Algorithm~\ref{alg:tensorized_rvea_selection} demonstrates the efficiency of tensorization. By performing operations on entire populations simultaneously, TensorRVEA significantly accelerates the selection process compared to its traditional counterpart. This enhancement makes it particularly effective for large-scale multiobjective optimization tasks, where maintaining a diverse set of solutions along the Pareto front is critical.

\section{Multiobjective Robot Control Benchmark}
\label{appendix:benchmark_def}

In this section, we introduce the multiobjective robot control problems (MoRobtrol) designed for this study. These problems are specifically crafted to evaluate the performance of the proposed tensorized EMO algorithms in realistic and challenging scenarios. The MoRobtrol consists of 9 distinct multiobjective problems: MoHalfcheetah, MoHopper, MoSwimmer, MoInvertedDoublePendulum (MoIDP), MoWalker2d, MoPusher, MoReacher, MoHumanoid, and MoHumanoidstandup (MoHumanoid-s). Each problem is a variant of a classic single-objective control task, reformulated to include multiple objectives that often conflict with one another, increasing both complexity and relevance to real-world applications.
The specific mathematical definitions of these problems are detailed below, offering a precise description of the objectives and constraints associated with each task.

\begin{figure}[!htb]
    \centering
     \begin{subfigure}[b]{0.15\textwidth}
        \includegraphics[width=\linewidth]{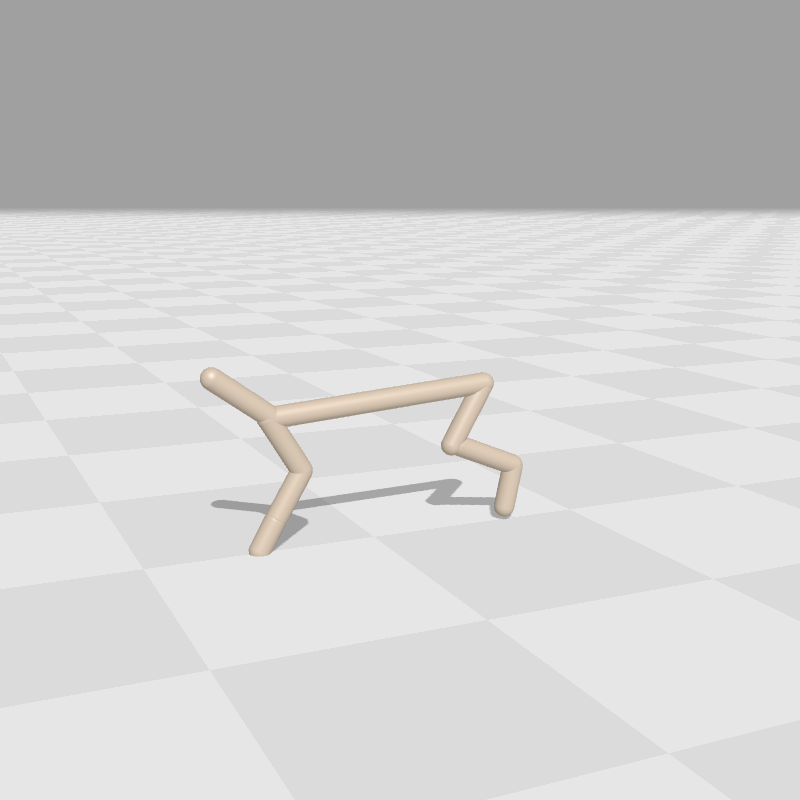}
        \caption{MoHalfcheetah}
        \label{fig:image16}
    \end{subfigure}
    \begin{subfigure}[b]{0.15\textwidth}
        \includegraphics[width=\linewidth]{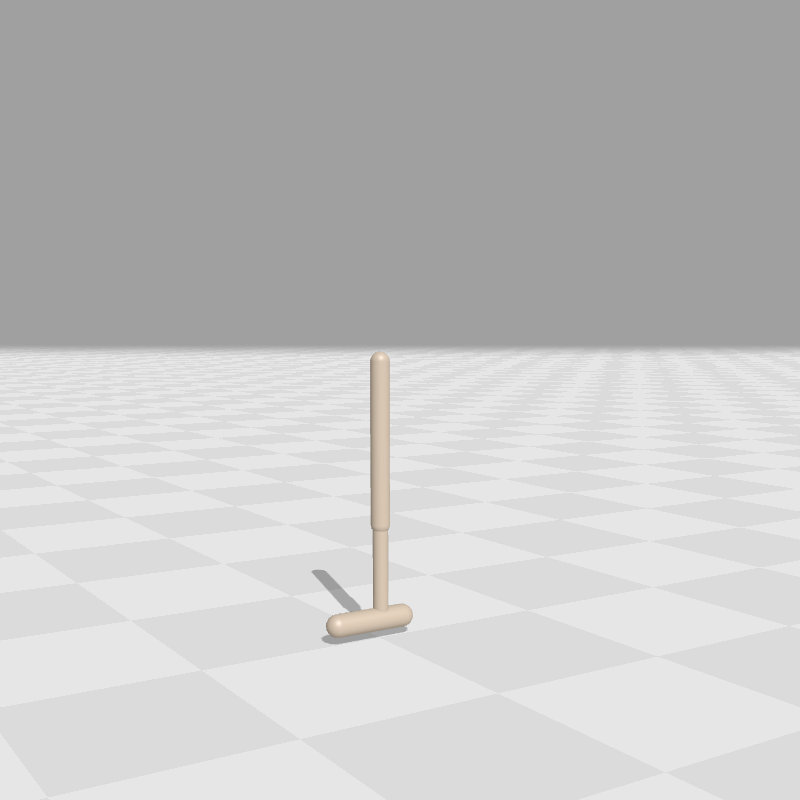}
        \caption{MoHopper}
        \label{fig:image17}
    \end{subfigure}
    \begin{subfigure}[b]{0.15\textwidth}
        \includegraphics[width=\linewidth]{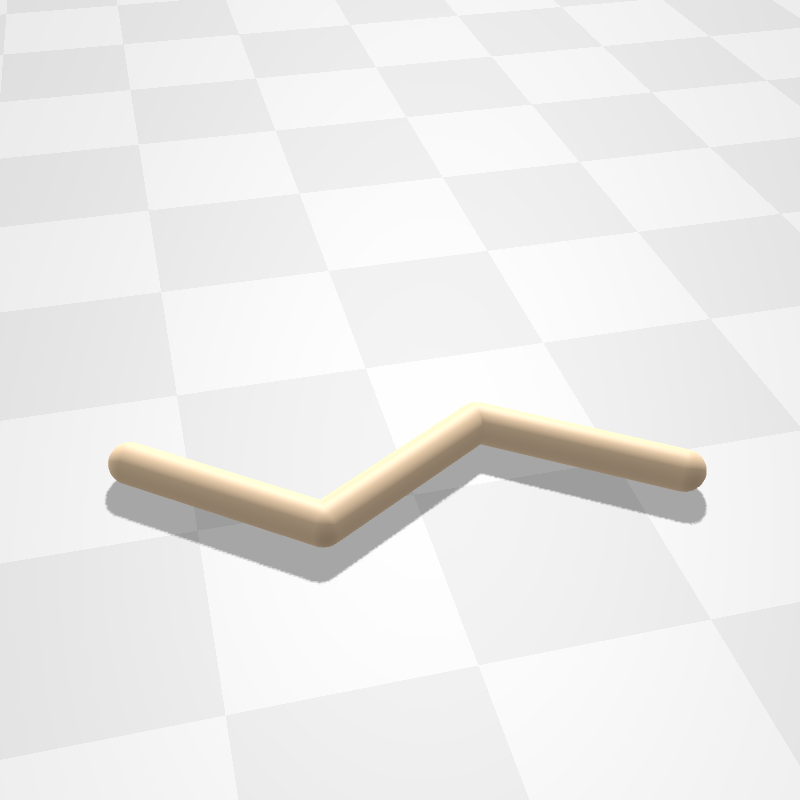}
        \caption{MoSwimmer}
        \label{fig:image18}
    \end{subfigure}

    \begin{subfigure}[b]{0.15\textwidth}
        \includegraphics[width=\linewidth]{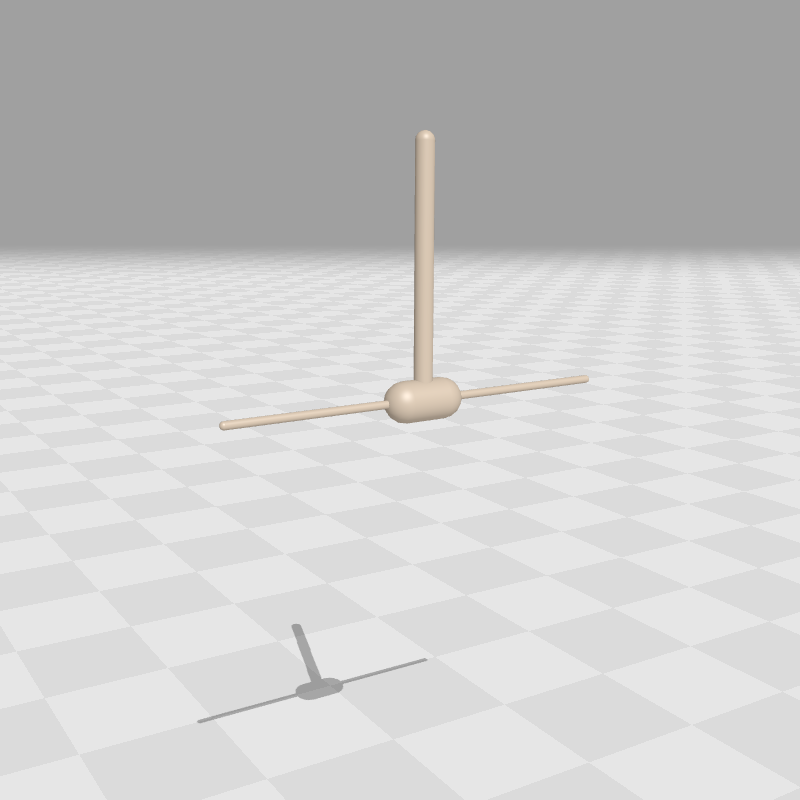}
        \caption{MoIDP}
        \label{fig:image19}
    \end{subfigure}
    \begin{subfigure}[b]{0.15\textwidth}
        \includegraphics[width=\linewidth]{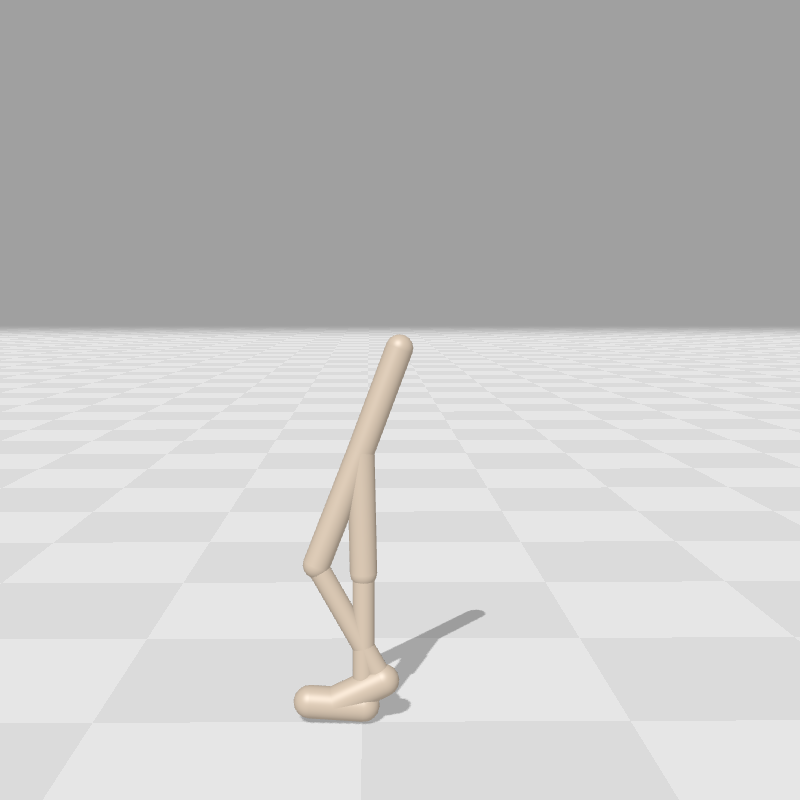}
        \caption{MoWalker2d}
        \label{fig:image20}
    \end{subfigure}
    \begin{subfigure}[b]{0.15\textwidth}
        \includegraphics[width=\linewidth]{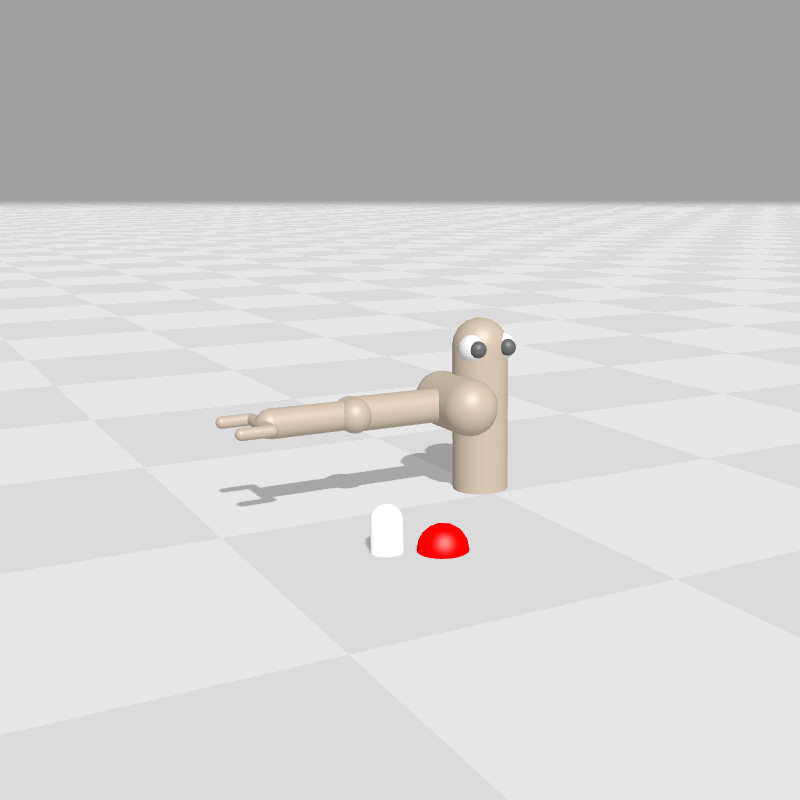}
        \caption{MoPusher}
        \label{fig:image21}
    \end{subfigure} 

    \begin{subfigure}[b]{0.15\textwidth}
        \includegraphics[width=\linewidth]{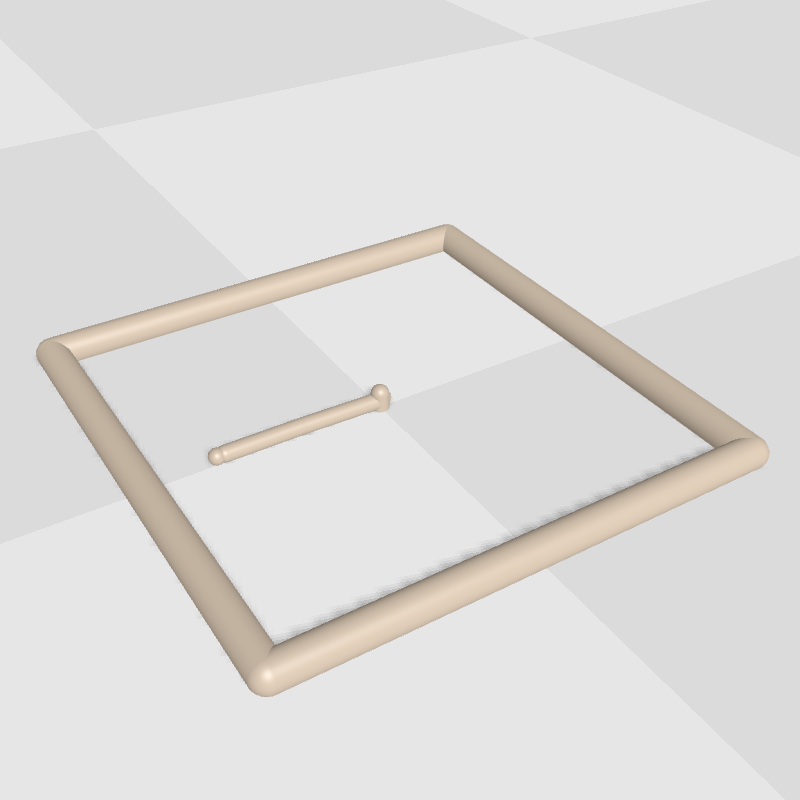}
        \caption{MoReacher}
        \label{fig:image19}
    \end{subfigure}
    \begin{subfigure}[b]{0.15\textwidth}
        \includegraphics[width=\linewidth]{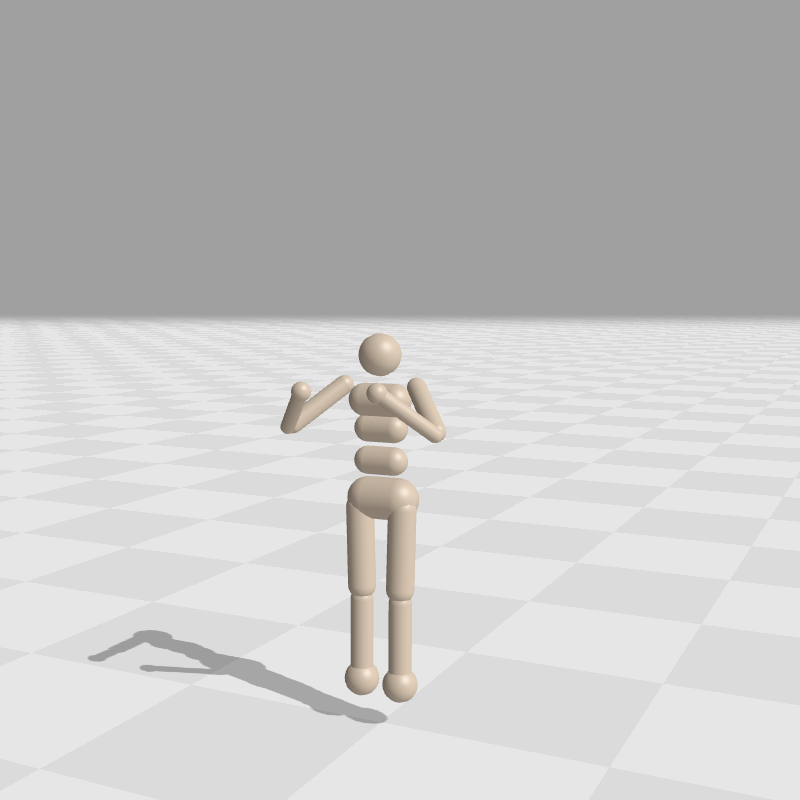}
        \caption{MoHumanoid}
        \label{fig:image20}
    \end{subfigure}
    \begin{subfigure}[b]{0.15\textwidth}
        \includegraphics[width=\linewidth]{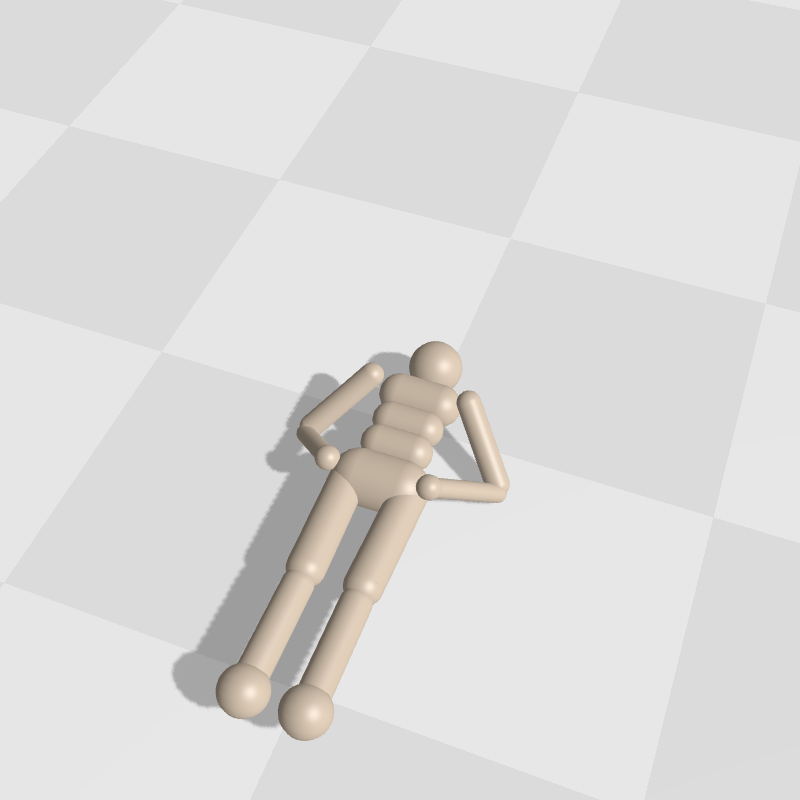}
        \caption{MoHumanoid-s}
        \label{fig:image21}
    \end{subfigure} 

    \caption{Visual illustration of the robot control tasks in the proposed MoRobtrol benchmark.}
    \label{fig:mrcb}
\end{figure}

% \begin{figure}[!htb]
%     \centering

%     \subfloat[MoHalfcheetah]{%
%         \includegraphics[width=0.16\textwidth]{figures/halfcheetah.png}%
%         \label{fig:image16}}
%     \hfill
%     \subfloat[MoHopper]{%
%         \includegraphics[width=0.16\textwidth]{figures/hopper.png}%
%         \label{fig:image17}}
%     \hfill
%     \subfloat[MoSwimmer]{%
%         \includegraphics[width=0.16\textwidth]{figures/swimmer.png}%
%         \label{fig:image18}}

%     \subfloat[MoIDP]{%
%         \includegraphics[width=0.16\textwidth]{figures/MoIDP.png}%
%         \label{fig:image19}}
%     \hfill
%     \subfloat[MoWalker2d]{%
%         \includegraphics[width=0.16\textwidth]{figures/walker2d.png}%
%         \label{fig:image20}}
%     \hfill
%     \subfloat[MoPusher]{%
%         \includegraphics[width=0.16\textwidth]{figures/pusher.png}%
%         \label{fig:image21}}

%     \subfloat[MoReacher]{%
%         \includegraphics[width=0.16\textwidth]{figures/reacher.png}%
%         \label{fig:image22}}
%     \hfill
%     \subfloat[MoHumanoid]{%
%         \includegraphics[width=0.16\textwidth]{figures/humanoid.png}%
%         \label{fig:image23}}
%     \hfill
%     \subfloat[MoHumanoid-s]{%
%         \includegraphics[width=0.16\textwidth]{figures/humannoidstandup.png}%
%         \label{fig:image24}}
%     \caption{\normalsize Visual illustration of the robot control tasks in the proposed MoRobtrol benchmark.}
%     \label{fig:mrcb}
% \end{figure}

\textbf{MoHalfcheetah}:
The observation space is represented as \( \mathbb{R}^{17} \), indicating a 17-dimensional vector for each observation. Similarly, the action space is defined as \( \mathbb{R}^{6} \), indicating a 6-dimensional vector for each action. Each episode is constrained to a maximum of 1000 time steps. The first objective, termed the forward reward, is defined as:
\begin{equation}
    f_\text{v} = w_1 \cdot v_x,
\end{equation}
where \( v_x \) denotes the velocity in the \( x \)-direction, and \( w_1 \) is the weight associated with this velocity component. The second objective, known as the control cost, is expressed as:
\begin{equation}
    f_\text{c} = -w_2 \cdot \sum_{i} a_i^2,
\end{equation}
where \( a_i \) represents the action taken by the \( i \)-th actuator, and \( w_2 \) is the weight applied to the control cost.

\textbf{MoHopper}:
The observation space is represented as \( \mathbb{R}^{11} \), indicating that each observation is an 11-dimensional vector. The action space is defined as \( \mathbb{R}^{3} \), meaning that each action is a 3-dimensional vector. Each episode is constrained to a maximum of 1000 time steps. The first objective, termed the forward reward, is given by:
\begin{equation}
    f_\text{v} = w_1 \cdot v_x + C,
\end{equation}
where \( v_x \) denotes the velocity in the \( x \)-direction, \( w_1 \) is the weight associated with this velocity component, and \( C = 1 \) is a constant survival reward indicating that the agent remains active. The second objective, referred to as height, is expressed as:
\begin{equation}
    f_\text{h} = 10 \cdot (h_\text{curr} - h_\text{init}) + C,
\end{equation}
where \( h_\text{curr} \) represents the current height of the hopper, and \( h_\text{init} \) denotes the initial height. The third objective, known as the control cost, is defined as:
\begin{equation}
    f_\text{c} = - \sum_{i} a_i^2 + C,
\end{equation}
where \( a_i \) represents the action taken by the \( i \)-th actuator.

\textbf{MoSwimmer}:
The observation space is represented as \( \mathbb{R}^{8} \), indicating that each observation is an 8-dimensional vector. The action space is defined as \( \mathbb{R}^{2} \), meaning that each action is a 2-dimensional vector. Each episode is constrained to a maximum of 1000 time steps. The first objective, referred to as the forward reward, is defined as:
\begin{equation}
    f_\text{v} = w_1 \cdot v_x,
\end{equation}
where \( v_x \) denotes the velocity in the \( x \)-direction, and \( w_1 \) is the weight associated with this velocity component. The second objective, known as the control cost, is expressed as:
\begin{equation}
    f_\text{c} = -w_2 \cdot \sum_{i} a_i^2,
\end{equation}
where \( a_i \) represents the action taken by the \( i \)-th actuator, and \( w_2 \) is the weight associated with the control cost.

\textbf{MoInvertedDoublePendulum (MoIDP)}: 
The observation space is represented as \( \mathbb{R}^{11} \), indicating that each observation is an 11-dimensional vector. The action space is defined as \( \mathbb{R}^{1} \), meaning that each action is a scalar. Each episode is constrained to a maximum of 1000 time steps. The first objective, referred to as the distance penalty, is defined as:
\begin{equation}
    f_\text{dp} = - w_1 \cdot (x - x_0)^2 - w_2 \cdot (y - y_0)^2 + C,
\end{equation}
where \( x \) and \( y \) are the current positions of the model, \( x_0 = 0 \) and \( y_0 = 2 \) are the initial positions, \( w_1 \) and \( w_2 \) are the weights for the distance penalty in the \( x \) and \( y \) directions, respectively, and \( C = 10 \) is a constant survival reward indicating that the agent remains operational. The second objective, known as the speed penalty, is expressed as:
\begin{equation}
    f_\text{sp} = -w_3 \cdot v_x^2 - w_4 \cdot v_y^2 + C,
\end{equation}
where \( v_x \) and \( v_y \) are the velocities in the \( x \) and \( y \) directions, respectively, and \( w_3 \) and \( w_4 \) are the weights associated with these velocity components.

\textbf{MoWalker2d}:
The observation space is represented as \( \mathbb{R}^{17} \), indicating that each observation is a 17-dimensional vector. The action space is defined as \( \mathbb{R}^{6} \), meaning that each action is a 6-dimensional vector. Each episode is limited to a maximum of 1000 steps. The first objective, referred to as the forward reward, is defined as:
\begin{equation}
    f_\text{v} = w_1 \cdot v_x + C,
\end{equation}
where \( v_x \) denotes the velocity in the \( x \)-direction, \( w_1 \) is the weight associated with this velocity component, and \( C = 1 \) represents a constant survival reward, indicating that the agent remains active. The second objective, known as the control cost, is expressed as:
\begin{equation}
    f_\text{c} = -w_2 \cdot \sum_{i} a_i^2 + C,
\end{equation}
where \( a_i \) represents the action taken by the \( i \)-th actuator, and \( w_2 \) is the weight associated with the control cost.

\textbf{MoPusher}:
The observation space is represented as \( \mathbb{R}^{23} \), indicating that each observation is a 23-dimensional vector. The action space is defined as \( \mathbb{R}^{7} \), meaning that each action is a 7-dimensional vector. Each episode is constrained to a maximum of 1000 time steps. The first objective, referred to as the near reward, is defined as:
\begin{equation}
    f_\text{n} = -\sqrt{(x_\text{fin} - x_\text{obj})^2 + (y_\text{fin} - y_\text{obj})^2},
\end{equation}
where \( x_\text{fin} \) and \( y_\text{fin} \) denote the coordinates of the fingertip, while \( x_\text{obj} \) and \( y_\text{obj} \) denote the coordinates of the object. The second objective, known as the distance reward, is expressed as:
\begin{equation}
    f_\text{d} = -\sqrt{(x_\text{obj} - x_\text{tar})^2 + (y_\text{obj} - y_\text{tar})^2},
\end{equation}
where \( x_\text{obj} \) and \( y_\text{obj} \) are the coordinates of the object, and \( x_\text{tar} \) and \( y_\text{tar} \) are the coordinates of the target. The third objective, termed the control cost, is defined as:
\begin{equation}
    f_\text{c} = -\sum_{i} a_i^2,
\end{equation}
where \( a_i \) represents the action taken by the \( i \)-th actuator.

\textbf{MoReacher}:
The observation space is represented as \( \mathbb{R}^{11} \), indicating that each observation is an 11-dimensional vector. The action space is defined as \( \mathbb{R}^{2} \), meaning that each action is a 2-dimensional vector. Each episode is constrained to a maximum of 1000 time steps. The first objective, referred to as the distance reward, is defined as:
\begin{equation}
    f_\text{d} = -\sqrt{(x_\text{fin} - x_\text{tar})^2 + (y_\text{fin} - y_\text{tar})^2},
\end{equation}
where \( x_\text{fin} \) and \( y_\text{fin} \) denote the coordinates of the fingertip, and \( x_\text{tar} \) and \( y_\text{tar} \) denote the coordinates of the target. The second objective, known as the control cost, is expressed as:
\begin{equation}
    f_\text{c} = -\sum_{i} a_i^2,
\end{equation}
where \( a_i \) represents the action taken by the \( i \)-th actuator.

\textbf{MoHumanoid}: 
The observation space is represented as \( \mathbb{R}^{244} \), indicating that each observation is a 244-dimensional vector. The action space is defined as \( \mathbb{R}^{17} \), meaning that each action is a 17-dimensional vector. Each episode is constrained to a maximum of 1000 time steps. The first objective, referred to as the forward reward, is defined as:
\begin{equation}
    f_\text{v} = w_1 \cdot v_x + C,
\end{equation}
where \( v_x \) denotes the velocity in the \( x \)-direction, \( w_1 \) is the weight associated with this velocity component, and \( C = 5 \) represents a constant survival reward, indicating that the agent remains operational. The second objective, known as the control cost, is given by:
\begin{equation}
    f_\text{c} = -w_2 \cdot \sum_{i} a_i^2 + C,
\end{equation}
where \( a_i \) represents the action taken by the \( i \)-th actuator, and \( w_2 \) is the weight applied to the control cost.

\textbf{MoHumanoidstandup (MoHumanoid-s)}: 
The observation space is represented as \( \mathbb{R}^{244} \), indicating that each observation is a 244-dimensional vector. The action space is defined as \( \mathbb{R}^{17} \), meaning that each action is a 17-dimensional vector. Each episode is constrained to a maximum of 1000 time steps. The first objective, referred to as the forward reward, is defined as:
\begin{equation}
    f_\text{v} = w_1 \cdot v_y + C,
\end{equation}
where \( v_y \) denotes the velocity in the \( y \)-direction, \( w_1 \) is the weight associated with this velocity component, and \( C = 5 \) represents a constant survival reward, indicating that the agent remains operational. The second objective, known as the control cost, is given by:
\begin{equation}
    f_\text{c} = -w_2 \cdot \sum_{i} a_i^2 + C,
\end{equation}
where \( a_i \) represents the action taken by the \( i \)-th actuator, and \( w_2 \) is the weight applied to the control cost.

\section{Experiments}
\label{appendix:experiments}

This section presents the performance indicators and reference points employed in the MoRobtrol benchmark, along with supplementary experiments covering comparisons with CUDA-based acceleration algorithms, analysis of GPU performance impact, and detailed evaluations on the DTLZ test suite.

\subsection{Performance Indicators}
\label{subsec:performance_indicators}

For a comprehensive analysis of the proposed tensorized EMO algorithms, we employ three key performance indicators: inverted generational distance (IGD), hypervolume (HV), and expected utility (EU) metric.

\subsubsection{IGD}
The IGD measures how closely the set of final solutions \( \bm{F} \) approximates the reference PF \( \bm{F}^* \). It is defined as:
\begin{equation}
    \text{IGD}(\bm{F}, \bm{F}^*) = \frac{\sum_{\bm{f}^* \in \bm{F}^*} \min_{\bm{f} \in \bm{F}} \| \bm{f} - \bm{f}^* \|}{|\bm{F}^*|},
\end{equation}
where \( \| \cdot \| \) denotes the Euclidean distance. A lower IGD value indicates better approximation to the true Pareto Front.

\subsubsection{HV}
The HV metric evaluates the volume of the objective space dominated by the obtained solutions and bounded by a reference point \( \bm{v}_{\text{ref}} \). The HV is calculated as:
\begin{equation}
    \text{HV}(\bm{F}, \bm{v}_{\text{ref}}) = \bigcup_{\bm{f} \in \bm{F}} \text{volume}(\bm{v}_{\text{ref}}, \bm{f} ),
\end{equation}
where \(\text{volume}(v_{\text{ref}}, \bm{f} )\) represents the volume of the hypercube defined by the vector \( \bm{f} \) and the reference point \( \bm{v}_{\text{ref}} \). The reference point \( \bm{v}_{\text{ref}} \) is predetermined as a vector of ones, and \( \bm{f} \) represents the set of normalized objective values. A higher HV value indicates better spread and convergence of the solutions.

\subsubsection{EU}

The EU for a set of objectives \(\bm{F} \in \mathbb{R}^{m \times n}\) is calculated as:

\begin{equation}
    \text{EU}(\bm{F}, \bm{W}) = \frac{1}{n} \sum_{j=1}^{n} \max_{i \in \{1, \dots, m\}} \left( \bm{W}^\top \cdot \bm{f}_i(\bm{F}_j) \right),
\end{equation}
where \(\bm{W} \in \mathbb{R}^{m}\) is the weight vector, \(\bm{F}_j\) is the \(j\)-th solution in the objective space, and \(\bm{f}_i(\bm{F}_j)\) represents the utility function applied to the \(i\)-th objective of the \(j\)-th solution. The expression \(\bm{W}^\top \cdot \bm{f}_i(\bm{F}_j)\) calculates the weighted utility of the objectives, and the \(\max\) operation selects the maximum utility across all objectives for each solution. The mean value across all solutions provides the expected utility.

\subsection{Comparison with CUDA-based Acceleration Algorithms}
\label{appendix:cuda_alg}
In this experiment, we compare the TensorNSGA-II algorithm with the NSGA-II implementation in EvoTorch~\cite{evotorch}. 
EvoTorch is a PyTorch-based evolutionary algorithm library that enables CUDA acceleration. 
Since EvoTorch only provides the NSGA-II algorithm, we conduct a comparison between TensorNSGA-II and the EvoTorch NSGA-II on the DTLZ test suite. 

\subsubsection{Experimental Settings}
The population size is set to 10000, with a dimensionality of 5000 and 3 objectives. 
Both algorithms are independently run 31 times, with 100 iterations per run. 
The comparison is based on the IGD and computation time. 
We employ the Wilcoxon rank-sum test to assess whether there are significant differences between the two algorithms, with the performance metrics of the superior algorithm highlighted.

\subsubsection{Comparison Results}
As shown in Table~\ref{exper_evotorch}, TensorNSGA-II outperforms the NSGA-II implementation in EvoTorch across all DTLZ problems, both in terms of IGD and computation time. 
Furthermore, on the RTX 4090 GPU, the speedup ranged from a maximum of 5.61$\times$ to a minimum of 1.59$\times$, demonstrating the advantages of the tensorized algorithm.

\begin{table}[htb]
    \centering
    \caption{Statistical Results (Mean and Standard Deviation) of the IGD and Runtime (s) Obtained by NSGA-II and TensorNSGA-II in DTLZ1–DTLZ7. The Best Results Are Highlighted.}
    \label{exper_evotorch}
    \resizebox{\columnwidth}{!}{
    \begin{tabular}{c c c c}
    \toprule
    \textbf{Indicator} & \textbf{Problem} & \textbf{NSGA-II} & \textbf{TensorNSGA-II} \\
    \midrule
    \multirow{7}{*}{IGD} 
    & DTLZ1 & 1.4861e$+$05 (4.9045e$+$02) & \textbf{1.3866e$+$05 (5.4954e$+$02)}  \\ 
    & DTLZ2 & 3.7692e$+$02 (2.7958e$+$00) & \textbf{2.3238e$+$02 (8.9165e$-$01)}  \\ 
    & DTLZ3 & 5.0293e$+$05 (6.4328e$+$02) & \textbf{4.6302e$+$05 (1.2591e$+$03)}  \\ 
    & DTLZ4 & 3.6834e$+$02 (6.4508e$+$00) & \textbf{2.3839e$+$02 (1.0780e$+$00)}  \\ 
    & DTLZ5 & 3.7780e$+$02 (2.7668e$+$00) & \textbf{2.4280e$+$02 (1.0162e$+$00)}  \\ 
    & DTLZ6 & 4.4523e$+$03 (2.9406e$+$00) & \textbf{4.3366e$+$03 (3.7426e$+$00)}  \\ 
    & DTLZ7 & 9.2249e$+$00 (3.1569e$-$01) & \textbf{7.1779e$+$00 (4.0208e$-$02)}  \\ 
    \midrule
    \multirow{7}{*}{Time} 
    & DTLZ1 & 9.2198e$+$00 (2.4736e$-$02) & \textbf{1.8127e$+$00 (4.0681e$-$02)}  \\ 
    & DTLZ2 & 9.0389e$+$00 (2.4060e$-$02) & \textbf{2.1148e$+$00 (1.4466e$-$02)}  \\ 
    & DTLZ3 & 9.3291e$+$00 (1.9043e$-$02) & \textbf{1.9310e$+$00 (1.5188e$-$02)}  \\ 
    & DTLZ4 & 1.3494e$+$01 (2.1555e$+$00) & \textbf{8.5085e$+$00 (2.0471e$-$01)}  \\ 
    & DTLZ5 & 9.4792e$+$00 (6.2154e$-$02) & \textbf{3.3992e$+$00 (5.1943e$-$02)}  \\ 
    & DTLZ6 & 8.8182e$+$00 (9.3220e$-$03) & \textbf{1.5714e$+$00 (1.1792e$-$02)}  \\ 
    & DTLZ7 & 9.7167e$+$00 (2.7945e$-$02) & \textbf{2.3176e$+$00 (1.1198e$-$02)}  \\ 
    \bottomrule
    \end{tabular}
    }
    
\end{table}

\subsection{Impact of Different GPUs on Performance}
In this experiment, we compare the performance of three proposed algorithms: TensorNSGA-III, TensorMOEA/D, and TensorHypE. The comparison is conducted across multiple hardware configurations, including a CPU and the GPUs RTX 2080Ti, RTX 3090, and RTX 4090.
The evaluation is conducted using the DTLZ1 problem. 
The goal is to analyze how different hardware configurations affect the performance of the algorithms, with a focus on speedup achieved on GPUs relative to the CPU.

\subsubsection{Experimental Settings}
For each algorithm, the population size is set to 10000, with a dimensionality of 1000 and 3 objectives. 
Each algorithm is independently executed 10 times on each device, with 100 iterations per run. 
The average execution time across the 10 runs is calculated for each configuration, and the speedup achieved on the GPU relative to the CPU is determined, where speedup is defined as the ratio of execution time on the CPU to that on the GPU.

\begin{table}[htb]
    \centering
    \caption{Statistical Results (Mean and Standard Deviation) of the Runtime (s) and Speedup for EMO Algorithms on Different Devices in DTLZ1}
    \label{table:differ_gpu}
    \resizebox{\columnwidth}{!}{
    \begin{tabular}{c c c c c c}
    \toprule
    \textbf{Device} & \textbf{Algorithm} & \textbf{Time} & \textbf{Speedup}\(^\dagger\) \\
    \midrule
    \multirow{3}{*}{CPU} 
    & NSGA-III & 8.7775e$+$02 (8.3293e$+$01) & 1.0  \\
    & MOEA/D & 9.6668e$+$02 (4.9265e$+$01) & 1.0  \\ 
    & HypE & 4.7882e$+$02 (2.2464e$+$01) & 1.0  \\ 
 
    \midrule
    \multirow{3}{*}{RTX 2080Ti} 
    & TensorNSGA-III & 3.8679e$+$00 (2.5282e$-$02) & 226.9 \\
    & TensorMOEA/D & 1.5599e$+$00 (1.2737e$-$02) & 619.7  \\ 
    & TensorHypE & 5.6312e$+$00 (1.2946e$-$01) & 85.0 \\ 

    \midrule
    \multirow{3}{*}{RTX 3090} 
    & TensorNSGA-III & 2.7511e$+$00 (1.7527e$-$02) & 319.1  \\ 
    & TensorMOEA/D & 8.6320e$-$01 (5.6514e$-$03) & 1119.9  \\ 
    & TensorHypE & 2.7511e$+$00 (1.7530e$-$02) & 174.0  \\ 

    \midrule
    \multirow{3}{*}{RTX 4090} 
    & TensorNSGA-III & 1.6648e$+$00 (1.0636e$-$02) & 527.2  \\ 
    & TensorMOEA/D & 6.0490e$-$01 (5.0061e$-$03) & 1598.1  \\ 
    & TensorHypE & 2.4712e$+$00 (6.0123e$-$02) & 193.8  \\ 

    \bottomrule
    \multicolumn{4}{l}{\footnotesize \(^\dagger\) Speedup = CPU Time / GPU Time.}
    \end{tabular}
    }
\end{table}

\subsubsection{Comparison Results}
As shown in Table~\ref{table:differ_gpu}, the type of GPU indeed affects the execution time. 
However, compared to the CPU, even the less powerful RTX 2080Ti GPU achieves a minimum speedup of 85$\times$, illustrating the scalability of tensorized algorithms across different hardware configurations. The maximum speedup reaches 226.9$\times$.
These results highlight the efficiency of tensorized algorithms in utilizing GPU resources, even on relatively low-end consumer hardware.
The acceleration effect is more pronounced with higher-end GPUs, reaching a maximum speedup of 1598.1$\times$ on the RTX 4090 GPU.

\subsection{Detailed Results for the DTLZ Test Suite}

\begin{table*}[t]
    \centering
    \caption{Statistical Results (Mean and Standard Deviation) of the IGD and Runtime (s) for Non-Tensorized and Tensorized EMO Algorithms in DTLZ1--DTLZ7. All Experiments Are Conducted on an RTX 4090 GPU and the Best Results Are Highlighted.}
    \label{table:dtlz_results}

    \begin{tabular}{c c c c c c}
    \toprule
    \textbf{Algorithm} & \textbf{Problem} & \textbf{IGD (Non-Tensorized)} & \textbf{IGD (Tensorized)} & \textbf{Time (Non-Tensorized)} & \textbf{Time (Tensorized)} \\
        \midrule
    \multirow{7}{*}{NSGA-III} 
    & DTLZ1 & \textbf{1.3619e$+$05 (3.6592e$+$02)} & 1.3693e$+$05 (4.3756e$+$02) & 7.9940e$+$01 (2.0346e$+$00) & \textbf{1.8314e$+$00 (1.5300e$-$02)} \\
    & DTLZ2 & 2.3069e$+$02 (1.2189e$+$00) & \textbf{2.2882e$+$02 (9.4040e$-$01)} & 4.2566e$+$01 (1.0087e$+$00) & \textbf{2.3500e$+$00 (1.3600e$-$02)} \\
    & DTLZ3 & \textbf{4.6228e$+$05 (1.1636e$+$03)} & \textbf{4.6225e$+$05 (1.2913e$+$03)} & 4.7783e$+$01 (6.8060e$-$01) & \textbf{2.1542e$+$00 (1.2700e$-$02)} \\
    & DTLZ4 & 2.3785e$+$02 (1.0874e$+$00) & \textbf{2.3572e$+$02 (8.5302e$-$01)} & 5.9497e$+$01 (9.8333e$-$01) & \textbf{8.5898e$+$00 (2.4899e$-$01)} \\
    & DTLZ5 & \textbf{2.3463e$+$02 (1.1257e$+$00)} & 2.3864e$+$02 (1.1237e$+$00) & 5.0183e$+$01 (1.0445e$+$00) & \textbf{4.1632e$+$00 (5.5200e$-$02)} \\
    & DTLZ6 & \textbf{4.2845e$+$03 (3.4493e$+$00)} & 4.3192e$+$03 (2.4402e$+$00) & 8.3887e$+$01 (9.5350e$-$01) & \textbf{1.8032e$+$00 (1.2700e$-$02)} \\
    & DTLZ7 & \textbf{6.8165e$+$00 (4.9090e$-$02)} & 7.0588e$+$00 (4.4870e$-$02) & 6.5831e$+$01 (9.2720e$-$01) & \textbf{2.6303e$+$00 (5.1900e$-$02)} \\
    \midrule
    \multirow{7}{*}{MOEA/D} 
    & DTLZ1 & \textbf{1.2951e$+$05 (1.9198e$+$03)} & 1.3819e$+$05 (5.6393e$+$02) & 8.9395e$+$01 (2.8132e$+$00) & \textbf{1.0660e$+$00 (6.6339e$-$03)} \\ 
    & DTLZ2 & \textbf{1.2359e$+$02 (2.0404e$+$01)} & 2.0785e$+$02 (1.3811e$+$00) & 8.8740e$+$01 (2.6398e$+$00) & \textbf{1.0694e$+$00 (5.3202e$-$03)} \\ 
    & DTLZ3 & \textbf{4.3746e$+$05 (4.3569e$+$03)} & 4.6991e$+$05 (1.7973e$+$03) & 9.0152e$+$01 (3.0706e$+$00) & \textbf{1.0701e$+$00 (3.3552e$-$03)} \\ 
    & DTLZ4 & 3.4160e$+$02 (1.8663e$+$01) & \textbf{2.1645e$+$02 (2.2032e$+$00)} & 7.2891e$+$01 (2.9190e$+$00) & \textbf{1.4081e$+$00 (4.2637e$-$03)} \\ 
    & DTLZ5 & \textbf{1.2483e$+$02 (2.4523e$+$01)} & 2.0857e$+$02 (1.7378e$+$00) & 8.6675e$+$01 (2.9418e$+$00) & \textbf{1.0661e$+$00 (3.9010e$-$03)} \\ 
    & DTLZ6 & \textbf{4.3760e$+$03 (5.5493e$+$00)} & 4.3915e$+$03 (3.0552e$+$00) & 8.7844e$+$01 (3.0132e$+$00) & \textbf{1.0704e$+$00 (2.9017e$-$03)} \\ 
    & DTLZ7 & \textbf{3.0036e$+$00 (1.2811e$-$01)} & 5.0944e$+$00 (1.9928e$-$01) & 8.6198e$+$01 (2.9972e$+$00) & \textbf{1.0743e$+$00 (3.8539e$-$03)} \\ 
    \midrule
    \multirow{7}{*}{HypE} 
    & DTLZ1 & \textbf{1.3705e$+$05 (5.2939e$+$02)} & \textbf{1.3705e$+$05 (5.2939e$+$02)} & 5.7589e$+$01 (1.5513e$+$00) & \textbf{2.3500e$+$00 (3.9000e$-$02)} \\
    & DTLZ2 & \textbf{1.9759e$+$02 (1.0991e$+$00)} & \textbf{1.9759e$+$02 (1.0991e$+$00)} & 5.8689e$+$01 (1.4897e$+$00) & \textbf{3.8541e$+$00 (4.2200e$-$02)} \\
    & DTLZ3 & \textbf{4.5667e$+$05 (8.3459e$+$02)} & \textbf{4.5667e$+$05 (8.3459e$+$02)} & 5.7680e$+$01 (1.5285e$+$00) & \textbf{2.8743e$+$00 (2.8500e$-$02)} \\
    & DTLZ4 & \textbf{2.0196e$+$02 (1.0117e$+$00)} & \textbf{2.0196e$+$02 (1.0712e$+$00)} & 6.7169e$+$01 (1.4286e$+$00) & \textbf{1.1103e$+$01 (3.8963e$-$01)} \\
    & DTLZ5 & \textbf{2.1053e$+$02 (1.1076e$+$00)} & \textbf{2.1053e$+$02 (1.1076e$+$00)} & 6.5312e$+$01 (1.4286e$+$00) & \textbf{1.0254e$+$01 (1.0990e$-$01)} \\
    & DTLZ6 & \textbf{4.3122e$+$03 (3.8700e$+$00)} & \textbf{4.3122e$+$03 (3.8700e$+$00)} & 5.7556e$+$01 (1.5598e$+$00) & \textbf{3.0027e$+$00 (5.2700e$-$02)} \\
    & DTLZ7 & \textbf{6.1372e$+$00 (2.9802e$-$02)} & \textbf{6.1372e$+$00 (2.9802e$-$02)} & 5.6269e$+$01 (1.7775e$+$00) & \textbf{2.9777e$+$00 (1.2100e$-$02)} \\

    \bottomrule
    \end{tabular}
    
\end{table*}

Table~\ref{table:dtlz_results} presents the results for tensorized and non-tensorized algorithms on the DTLZ test suite. Tensorized algorithms achieve faster runtimes while maintaining similar IGD values. The performance differences are small and consistent across most problems. Some variations are due to the batch operations in tensorized algorithms, which simplify computations compared to the original methods.

\subsection{Experiment in Multiobjective Robot Control Benchmark}
\label{subsec:experiment_robot_control}

For the HV calculation in the multiobjective robot control benchmark experiments, reference points were established based on the minimum values of nondominated solutions obtained by all algorithms for each problem. For objectives related to forward movement and height, if the minimum value was less than 0, the reference value was set to 0. In the MoPusher and MoReacher problems, the reference points were derived by taking the minimum values across all algorithms for each objective. Table~\ref{tab:reference_points} provides the reference points used for the 9 multiobjective robot control problems.

\begin{table}[H]
    \centering
    \caption{The Reference Points of 9 MoRobtrol Problems for HV Calculation}
    \label{tab:reference_points}
    \begin{tabular}{cc}
        \toprule
        \textbf{Problem} & \textbf{Reference Point} \\
        \midrule
        MoHalfcheetah & $(0, -530.97)$ \\
        MoHopper & $(283.52, -994.39, -1498.70)$ \\
        MoSwimmer & $(0, -0.20)$ \\
        MoIDP & $(3246.31, 153.32)$ \\
        MoWalker2d & $(0, 11.99)$ \\
        MoPusher & $(-1505.66, -984.00, -13645.28)$ \\
        MoReacher & $(-389.79, -1668.30)$ \\
        MoHumanoid & $(340.74, 321.41)$ \\
        MoHumanoid-s & $(4756.48, -6.06)$ \\
        \bottomrule
    \end{tabular}
\end{table}
\vfill

\end{document}